\documentclass[lettersize,journal]{IEEEtran}
\usepackage{amsmath,amsfonts}
\usepackage{algorithmic}
\usepackage{algorithm}
\usepackage{array}
\usepackage[caption=false,font=normalsize,labelfont=sf,textfont=sf]{subfig}
\usepackage{textcomp}
\usepackage{stfloats}
\usepackage{url}
\usepackage{verbatim}
\usepackage{graphicx}
\usepackage{cite}
\usepackage{algorithmic}
\usepackage{algorithm}
\usepackage{tabularx}
\usepackage{booktabs}
\usepackage{multirow}
\usepackage{comment}
\usepackage{graphicx}
\usepackage{float}
\usepackage{pifont}
\usepackage{subcaption}
\usepackage{CJKutf8}
\usepackage{mathrsfs}
\usepackage{makecell}

\usepackage{color}

\usepackage{colortbl}
\usepackage{xcolor}
\colorlet{light}{green!50}
\colorlet{lightlight}{green!25}

\colorlet{colorSep}{blue!5}

\newcommand{\FrameworkNM}[1]{Hi-LOAM#1}

\usepackage[colorlinks,
            linkcolor=blue,
            anchorcolor=blue,
            citecolor=blue]{hyperref}

\newcommand{\xmark}{\ding{55}}%

\begin{document}

\title{Hi-LOAM: Hierarchical Implicit Neural Fields for LiDAR Odometry and Mapping}

\author{
Zhiliu Yang,~\IEEEmembership{Member,~IEEE},
Jianyuan Zhang,~\IEEEmembership{Student Member,~IEEE},
Lianhui Zhao,~\IEEEmembership{Student Member,~IEEE},
Jinyu Dai,~\IEEEmembership{Student Member,~IEEE},
Zhu Yang,~\IEEEmembership{Member,~IEEE}


\thanks{
This manuscript is the accepted version of IEEE Transactions on Multimedia. ~\copyright~2026 IEEE. Personal use of this material is permitted. Permission from IEEE must be obtained for all other uses, in any current or future media, including reprinting/republishing this material for advertising or promotional purposes, creating new collective works, for resale or redistribution to servers or lists, or reuse of any copyrighted component of this work in other works.}


\thanks{
Zhiliu Yang, Jianyuan Zhang, Lianhui Zhao, and Jinyu Dai are with the School of Information Science and Engineering, Yunnan University, Kunming, Yunnan 650500, China.
}

\thanks{
Zhu Yang is with the School of Information and Electronics, Beijing Institute of Technology, Beijing 100081, China.}

\thanks{
Zhiliu Yang is also with the Yunnan Key Laboratory of Intelligent Systems and Computing, Yunnan University, Kunming, Yunnan 650500, China.
}

\thanks{
(Corresponding authors: Zhiliu Yang; Zhu Yang.)}

}

\markboth{Journal of \LaTeX\ Class Files,~Vol.~14, No.~8, August~2021}%
{Shell \MakeLowercase{\textit{et al.}}: A Sample Article Using IEEEtran.cls for IEEE Journals}

\IEEEpubid{0000--0000/00\$00.00~\copyright~2021 IEEE}

\maketitle

\begin{abstract}
LiDAR Odometry and Mapping (LOAM) is a pivotal technique for embodied-AI applications such as autonomous driving and robot navigation. Most existing LOAM frameworks are either contingent on the supervision signal, or lack of the reconstruction fidelity, which are deficient in depicting details of large-scale complex scenes. To overcome these limitations, we propose a multi-scale implicit neural localization and mapping framework using LiDAR sensor, called \FrameworkNM{}. \FrameworkNM{} receives LiDAR point cloud as the input data modality, learns and stores hierarchical latent features in multiple levels of hash tables based on an octree structure, then these multi-scale latent features are decoded into signed distance value through shallow Multilayer Perceptrons (MLPs) in the mapping procedure. For pose estimation procedure, we rely on a correspondence-free, scan-to-implicit matching paradigm to estimate optimal pose and register current scan into the submap. The entire training process is conducted in a self-supervised manner, which waives the model pre-training and manifests its generalizability when applied to diverse environments. Extensive experiments on multiple real-world and synthetic datasets demonstrate the superior performance, in terms of the effectiveness and generalization capabilities, of our \FrameworkNM{} compared to existing state-of-the-art methods.
\end{abstract}


\section{Introduction}

\IEEEPARstart{L}{ocalization} and mapping is indispensable to the deployment of intelligent embodied agents, such as mobile robots and self-driving vehicles \cite{chen2023sgsr,raychaudhuri2025semantic}. It requires to estimate the 6-degree-of-freedom (6-DoF) poses of moving agents and build the surrounding map of the environment, which lays a foundation for down-stream tasks such as collision avoidance and path planning \cite{wang2022d}. Although the Global Navigation Satellite System (GNSS) is a solution for tackling localization challenges, it may still fail to provide sufficiently accurate positional information in certain scenarios where the GNSS signal is denied, such as tunnels, densely built areas, and parking structures \cite{zhou2021t}. Vision-based SLAM system is another feasible solution, but it is suffered to the depth estimation issue and heavily dependent on the surface textures and illumination \cite{yuan2023cr}. In this paper, we leverage a multi-scale neural feature embedding and a scan-to-implicit-map matching strategy to achieve the LiDAR odometry and mapping (LOAM).
\begin{figure}[t!] 
    \centering 
    \includegraphics[width=0.48\textwidth]{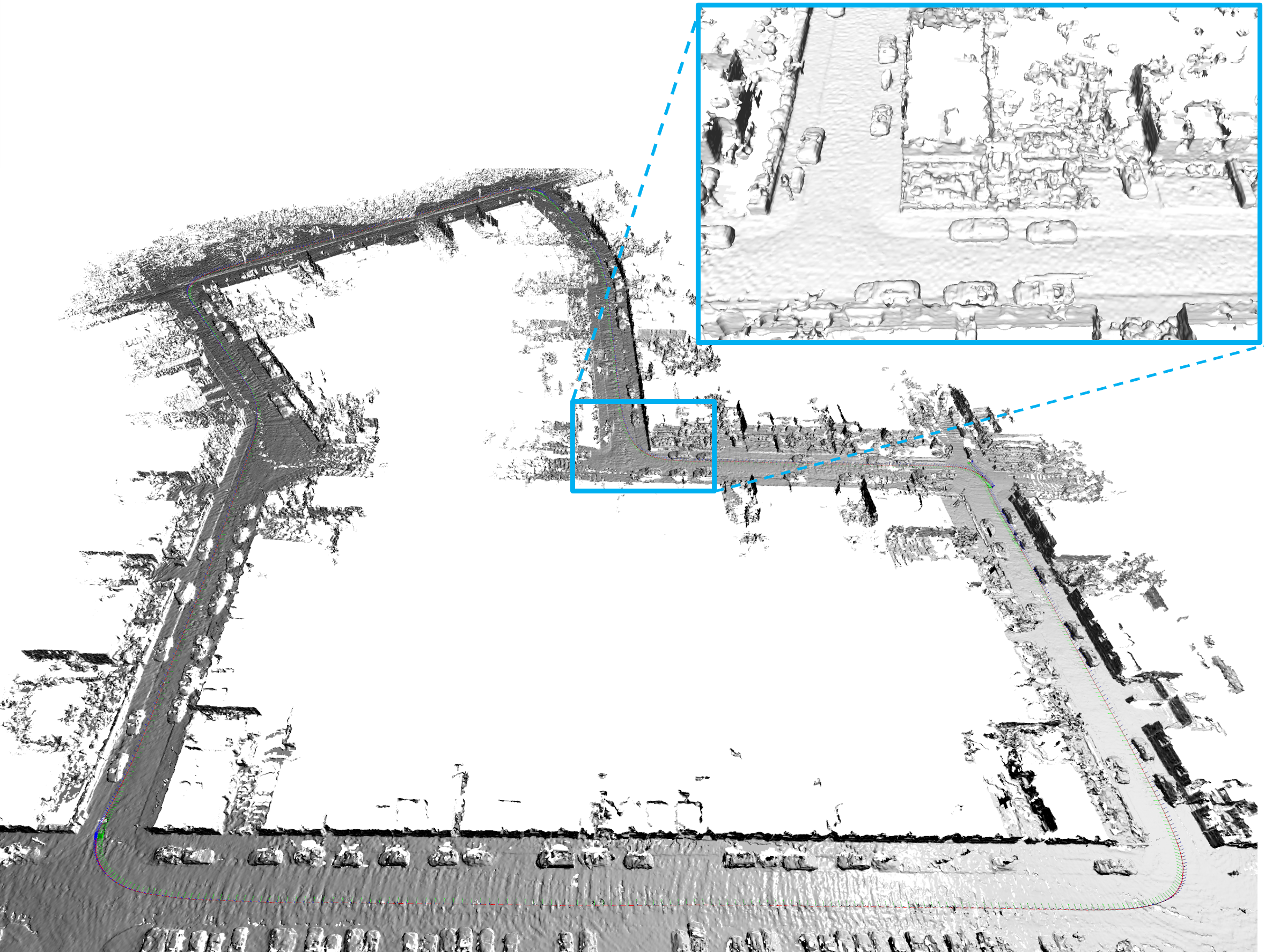} 
    \caption{\textbf{Reconstructed map and estimated trajectory generated via our implicit \FrameworkNM{}.} The above case is evaluated on the sequence 07 of \textit{KITTI} dataset, superior mapping quality and accurate poses are obtained via deep utilization of  hierarchical neural features.} 
    \label{fig:kitti_07} 
\end{figure}


\IEEEpubidadjcol
The traditional LiDAR odometry estimates the transformation between consecutive LiDAR frames via employing Iterative Closest Point (ICP) algorithm \cite{besl1992method}, which aligns consecutive scans by minimizing point cloud distance. Some variants of ICP method improve its precision and efficiency by introducing a point-to-plane \cite{grant2012point} technique which takes advantage of surface normal information. Then, IMLS-SLAM \cite{deschaud2018imls} tackles the LiDAR odometry problem using a scan-to-model matching framework. However, due to the non-uniformity and sparsity of the LiDAR point clouds, these methods fail in certain corner-case datasets. Hence, lots of learning-based methods are proposed to achieve significant performance boost, like LO-Net \cite{li2019net}, LodoNet \cite{zheng2020lodonet}, and PWCLO \cite{wang2021pwclo}. While, those learning-based methods require ground truth poses in the training stage, impeding their applications in the real-world scenarios. The self-supervised learning-based odometry is worth to be explored.

Recently, the technique called neural implicit representation \cite{vizzo2022vdbfusion} manifests the promising potential in surface modeling and pose tracking, but most of those works, such as iMAP \cite{sucar2021imap}, Nice-slam \cite{zhu2022nice}, Vox-fusion \cite{yang2022vox},  focus only on the indoor reconstruction and utilize cameras as sensors. A pivotal work named NeRF-LOAM \cite{deng2023nerf} extends mapping to the large scale environment via employing sparse octree-based voxels embedding. Different from the grid-based representation of NeRF-LOAM \cite{deng2023nerf}, PIN-SLAM \cite{pan2024pin} proposes a point-based implicit neural representation. Similar to PIN-SLAM \cite{pan2024pin}, we leverage a scan-to-implicit map for pose estimation, but for achieving higher localization precision, we adopt multi-scale latent features based on the octree rather than single-scale latent feature. The single-scale feature is lightweight for speeding-up the computation, but it is usually insufficient to capture and adapt to the nuances of the environment, resulting in the inaccurate pose estimation.

For the mapping fidelity, traditional map representations, such as point cloud \cite{zhang2014loam, vizzo2023kiss}, voxel grid \cite{zhou2023lidar}, and surfels \cite{chen2019suma++}, lack the connectivity among their mapping primitives, which is a sparse representation. 
The implicit representation \cite{mao2024ngel} widely attracts researchers' attention due to its high-fidelity rendering ability. iSDF \cite{Ortiz:etal:iSDF2022} utilizes a neural network to map input 3D coordinates to approximate signed distances. SHINE-Mapping \cite{zhong2023shine} also approximates signed distances via storing latent features in octree-based sparse voxel grids. However, these implicit methods require ground truth poses for map reconstruction, lacking the pose estimation ability also prohibits their further applications.

In this paper, we leverage a hierarchical latent feature to improve the localization accuracy and mapping quality. We first employ an octree-based feature volume to adaptively fit geometry shape with multiple discrete levels of detail. This hierarchical latent feature improves the reconstruction quality. Then, since the map quality is refined, a scan-to-implicit-map localization module is considered to enhance the localization accuracy. For the registration, we adopt the scan-to-submap strategy instead of global map to avoid the interference from global feature embeddings. For the mapping, local feature embeddings are optimized and directly integrated into the global map. To visualize the implicit map, we utilize the Marching Cubes algorithm to generate the mesh. The entire training
process of \FrameworkNM{} is conducted in a self-supervised manner, which waives
the model pretraining and manifests its advantage when
applied to diverse environments. In conclusion, we propose a hierarchical neural implicit framework for large scale mapping and pose estimation in a self-supervised fashion. The contributions of our paper are listed as follows:
\begin{itemize}
    \item We propose a novel localization and mapping framework for LiDAR-only data by adopting a hierarchical neural implicit representation. Our source code is publicly available\footnote{https://github.com/Zhangjyhhh/Hi-LOAM}.
    \item Our hierarchical feature embedding method achieves better mapping quality than state-of-the-art (SOTA) LiDAR SLAM approaches.
    \item Our method achieves localization accuracy better than SOTA learning-based LiDAR Odometry and on par with the SOTA ICP-based LiDAR Odometry.
    \item We demonstrate the superiority of our method for both mapping and localization on more than seven datasets. 
\end{itemize}


\section{Related Work}
\subsection{LiDAR Odometry}
Precise localization and reconstruction of the surrounding environment are widely explored by previous researchers. In this paper, we specifically focus on the LiDAR-based localization and reconstruction. LiDAR odometry can be roughly divided into three categories \cite{jonnavithula2021lidar}: point correspondence, distribution correspondence, and network correspondence.

\subsubsection{\textbf{Point Correspondence}} Besl et al. \cite{besl1992method} is the earliest work leveraging ICP to find the transformation matrix between consecutive LiDAR frames, which belongs to the point-to-point paradigm \cite{shan2018lego}, \cite{dellenbach2022ct}. Then a point-to-plane ICP variant \cite{grant2012point} is proposed to improve the accuracy and efficiency of the odometry. Generalized ICP \cite{segal2009generalized} combines the point-to-point ICP with point-to-plane ICP to yield better results than either method alone. Later, SemanticICP \cite{parkison2018semantic} incorporates semantic information and simultaneously optimizes it with the geometric information, further improves the odometry accuracy. SuMa++ \cite{chen2019suma++} also incorporates semantic information to improve precision on top of the SuMa \cite{behley2018efficient} and leverages dynamic objects removal to facilitate the localization. DCP \cite{wang2019deep} proposes the transformer-based point cloud matching to avoid falling into local optima, which helps to mitigate the sensitivity issue lies in the initialization of the ICP methods.

\subsubsection{\textbf{Distribution Correspondence}} In order to reduce the impact of subtle environment changes, the Normal Distribution Transform (NDT) \cite{biber2003normal} algorithm is introduced. The NDT transforms a single scan into a continuous and differentiable probability density defined on the 2D plane, then it is matched with the following scan, which eliminates finding correspondences between the features or points as previous methods. 3DNDT \cite{magnusson20083d} extends this approach into three-dimensional space. PNDT \cite{hong2017probabilistic} improves the accuracy of NDT-based registration by generating distributions in all occupied cells regardless of the cell resolution. AugmentedNDT \cite{akai2017robust} leverages offline estimated uncertainty information, particle filtering algorithms, and road marker matching to enhance the localization accuracy, especially when there are changes in the environment's appearance. WeightedNDT \cite{lee2020robust} removes dynamic objects to improve the localization accuracy by assigning each point a weight that reflects the static probability.

\subsubsection{\textbf{Network Correspondence}} This category of method employs the neural network to register two continuous LiDAR frames \cite{liu2023translo, adis2021d3dlo, zheng2020lodonet}. LO-Net \cite{li2019net} utilizes a deep convolutional neural network to realize real-time LiDAR-based pose estimation in an end-to-end manner. DeepPCO \cite{wang2019deeppco} proposes a novel parallel neural network to estimate translation and orientation separately, achieving more accurate results. PWCLO \cite{wang2021pwclo} employs an hierarchical network structure to obtain multi-scale feature for the registration. However, those methods are supervised and rely on expensive ground truth values which are difficult to obtain in practical scenarios. 
Currently, unsupervised localization method \cite{fu2022self} attracts more and more attention. Cho et al. \cite{cho2020unsupervised} propose first unsupervised learning-based LiDAR odometry whose loss function is similar to the point-to-plane error.
HPPLO-net \cite{zhou2023hpplo} is considered as the most precise deep LiDAR odometry using hierarchical multi-scale feature for the registration. 
Due to the incorporation of more local or global environment information, multi-scale approaches lead to more accurate localization compared to single-scale methods.

\subsection{LiDAR Mapping}
Traditional explicit map representations, such as point cloud \cite{vizzo2023kiss}, \cite{pan2021mulls}, voxel grid \cite{zhou2023lidar}, surfels \cite{chen2019suma++}, and triangular mesh \cite{vizzo2021poisson}, are widely used in tasks like path planning, navigation, three-dimensional reconstruction. Currently, more and more attention is paid to the implicit map representation due to its regression ability can cover more surface details. NGLOD \cite{takikawa2021neural} implicitly represents 3D object surfaces with an octree-based feature volume, achieving the real-time rendering. Di-fusion \cite{huang2021di} balances between the storage efficiency and surface quality via leveraging a Probabilistic Local Implicit Voxels model to encode the uncertainty of the scene geometry. SLIM \cite{yu2025slim} proposes a scalable and lightweight LiDAR mapping system, mainly focuses on the large-scale map merging. $\mathrm{N}^3$-Mapping \cite{song2024n} constructs normal guided neural non-projective signed distance fields to build a large-scale map, while it is relied on the ground truth pose.

NeRF-LOAM \cite{deng2023nerf} is the first neural implicit LiDAR odometry and mapping (LOAM) framework for the large-scale environment, which map the 3D coordinate into the signed distance field (SDF) value with the sparse voxel embedding. LONER \cite{isaacson2023loner} integrates an ICP-based online LiDAR odometry into the incremental implicit neural mapping, achieving real-time performance. PIN-SLAM \cite{pan2024pin} is a point-based implicit neural representation, which solves the loop closure correction and globally-consistent mapping using implicit neural map representation, obtaining the competitive reconstruction quality and localization precision. SHINE\_Mapping \cite{zhong2023shine} also maps the 3D coordinate into the SDF value via Multilayer Perceptrons (MLPs), but it is not a complete SLAM system. Takikawa et al. \cite{takikawa2021neural} adopt different level of features to fit objects' shapes, using multiple discrete levels of detail (LODs) to improve reconstruction quality, but it is not scaled up to the large-scale scenario. On the other hand, CURL-SLAM \cite{zhang2025curl} employs spherical harmonics functions to implicitly encode 3-D LiDAR points, generating an ultra-compact and globally consistent 3-D map, it trades localization precision for memory efficiency.

With the advent of the 3D Gaussian Splatting (3DGS) \cite{kerbl20233d}, which can accelerate rendering and improve reconstruction quality, an increasing number of works leverage 3DGS to boost the mapping performance. Zou et al. \cite{zou2024triplane} employ a hybrid Triplane-Gaussian to obtain high quality reconstruction and fast rendering speed. However, these works are only valid for small-scale visual scenes, not applicable to outdoor LiDAR data. Recently, GS-LiDAR \cite{jiang2025gs} proposes a method of LiDAR-based novel view synthesis that employs Periodic Vibrating 2D Gaussian primitives. Splat-LOAM \cite{giacomini2025splat} leverages 2D Gaussian primitives as the sole scene representation that achieve high quality map and low memory consumption, but its localization ability is not fully verified and its mapping result is restricted to the ground truth pose. Besides, more works, such as PINGS \cite{pan2025pings} and Gaussian-LIC2 \cite{lang2025gaussian}, further explore the role of Gaussian splatting in the LiDAR mapping task, but their system inputs require extra sensors, such as cameras and IMUs, to fuse with LiDAR, increasing the complexity of the system.

\subsection{Multi-scale Features for 3-D Scene Understanding}
Multi-scale, or hierarchical, features of 3-D scene understanding is firstly utilized in vision-based works \cite{takikawa2021neural,Ortiz:etal:iSDF2022,wang2022go,zhu2022nice,johari2023eslam}. NGLOD \cite{takikawa2021neural} adaptively fits shapes with multiple discrete levels of detail (LODs) to improve reconstruction quality for complex shapes. iSDF \cite{Ortiz:etal:iSDF2022} utilizes a neural network to map input 3D coordinates to approximate signed distances, provides adaptive levels of details with plausible mapping quality. Go-SURF \cite{wang2022go} models the scene with the learned hierarchical feature, NICE-SLAM \cite{zhu2022nice} introduces a hierarchical scene representation, ESLAM \cite{johari2023eslam} represents scene as multi-scale axis-aligned perpendicular feature.
All these methods improve the geometric and appearance modeling efficiency by introducing multi-level features. Later, hierarchical features of 3-D scene understanding is adopted in LiDAR-based SLAM systems. SHINE Mapping \cite{zhong2023shine} 
maps the 3D coordinate into the SDF value via multi-level features and MLPs. HPPLO-net \cite{zhou2023hpplo} leverages the hierarchical multi-scale feature for registration stage to improve the overall LiDAR odometry.

\subsection{Mesh-based LOAM}
Mesh-based LOAM leverages triangular mesh to represent the surrounding environment rather than the sparse representation, such as point cloud and surfels. LONER \cite{isaacson2023loner} is the first real-time LiDAR SLAM that employs a neural implicit scene representation to generate mesh. Slamesh \cite{ruan2023slamesh} proposes the first CPU-only real-time LiDAR SLAM system that utilizes the Gaussian process reconstruction to realize the fast building, registration, and updating of mesh maps. Although Slamesh achieves excellent results, it becomes significantly complicated when modeling complex geometries. To overcome this issue, Mesh-LOAM \cite{zhu2024mesh} employs an incremental voxel meshing strategy and a parallel spatial-hashing scheme to recover trajectories and mesh maps. Although the aforementioned works have made significant progress in terms of the real-time performance, the quality of the generated maps is still prone to artifacts and holes. 

 \begin{figure*}[ht!]
    \centering
    
    \includegraphics[width=0.99\linewidth]{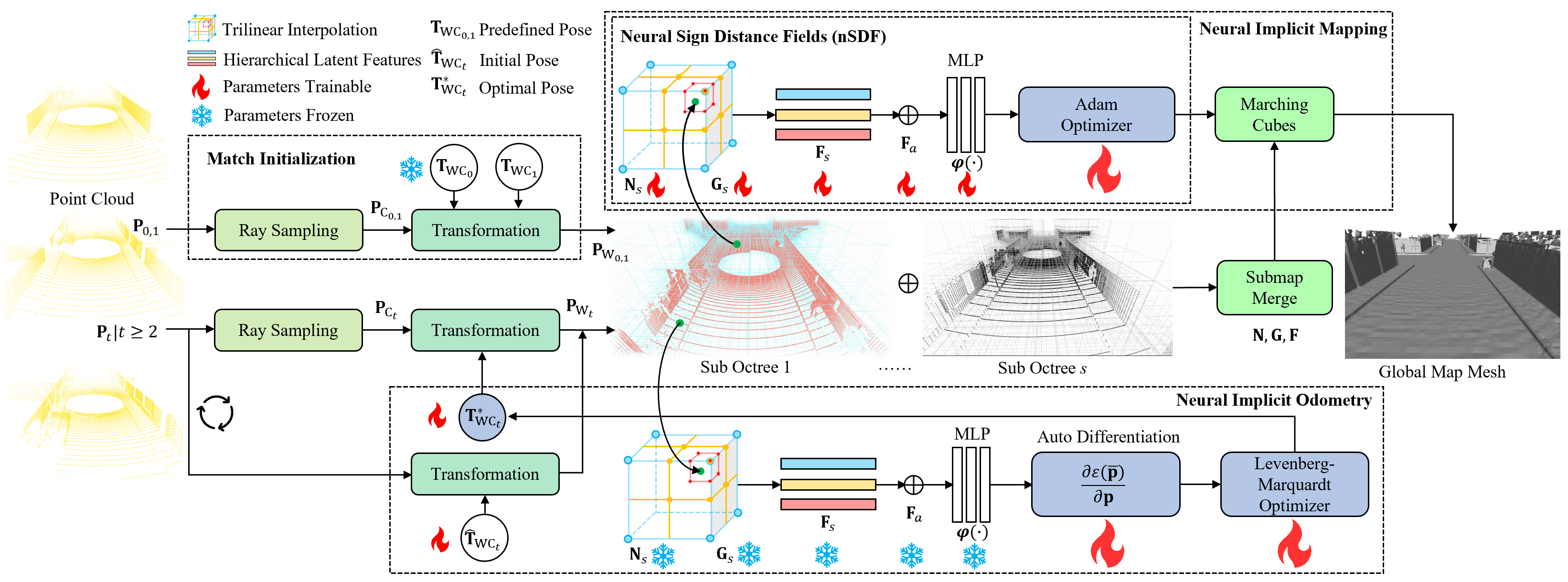}
    
    \caption{\textbf{The Overview of Our \FrameworkNM{} Framework}. The main data stream is depicted as following. Sampled points $\mathbf{P}_{\mathrm{C}_t}$ along the ray is used to construct neural signed distance fields (nSDF) based on the octree. For the first two scans, poses $\mathbf{P}_{0}$ and $\mathbf{P}_{1}$ are predefined and nSDF are created. For following scans, original points $\mathbf{P}_{t}$ are firstly matched with octree of sub map to estimate the pose $\mathbf{T}_{\mathrm{WC}_t}^*$. Then, $\mathbf{P}_{t}$ is transformed into the global coordinate system via $\mathbf{T}_{\mathrm{WC}_t}^*$, $\mathbf{P}_{\mathrm{W}_t}$ is registered to expand the nSDF, this process is repeated in turn. The separate optimization strategies are designed for odometry and mapping, the trainable parameters and frozen parameters are marked correspondingly.
    }
    \vspace{-0.2cm}
    \label{fig:sysoverview}
\end{figure*}

\section{Methodology}

In this paper, we present an incremental LiDAR odometry and mapping (LOAM) framework via hierarchical neural implicit representation, called \FrameworkNM{}, for large-scale environment, the system overview of \FrameworkNM{} is shown in Fig. \ref{fig:sysoverview}. Firstly, we exploit the sampled LiDAR data to construct hierarchical latent features based on the octree, which implicitly constructed the neural signed distance fields (nSDF). Then the scan-to-implicit matching utilizes the hierarchical latent features of the current scan to execute the pose estimation and scan registration. Finally, for explicitly displaying the implicit scene, the Marching Cubes algorithm is employed to generate the triangular mesh.

\subsection{Problem Statement} 
Our framework is targeted at the odometry and the mapping. The input point cloud of a 3D LiDAR scan is denoted as $\mathbf{P}_t$, where the $ t \in \mathbb{N}$ is the scan index. An individual point within the scan is denoted as $\mathbf{p} \in \mathbb{R}^3$. The output goal is to estimate \textbf{(i)} the six-degree-of-freedom (6-DoF) LiDAR pose $\mathbf{T}_{\mathrm{WC}_{t}}$, and \textbf{(ii)} the 3D map $\mathcal{M}$ of the surrounding environment, which is represented as the mesh. The map $\mathcal{M}$ is represented by a hierarchical sequence of multi-scale features based on an octree structure, which is detailed in \ref{sec: map}. For rest of the paper, we define $\mathbf{T}_\mathrm{BA} \in \mathrm{SE}(3)$ as the transformation that converts the point cloud $\mathbf{P}_\mathrm{A}$ in the coordinate frame $\textrm{A}$ to the coordinate frame $\textrm{B}$, where $\mathrm{SE}(3)$ denotes the Special Euclidean Group in three dimensions. As seen in the equation (\ref{equ:trsfm}), $\mathbf{T}_\mathrm{BA}$ always can be decomposed to a rotation $\mathbf{R}_\mathrm{BA} \in \mathrm{SO}(3)$ and a translation $\mathbf{t}_\mathrm{BA}\in \mathbb{R}^3$ respectively, where $\mathrm{SO}(3)$ denotes the Special Orthogonal Group in three dimensions.

\begin{equation}
\mathbf{T}_\textrm{BA} = 
\begin{bmatrix} \mathbf{R}_\mathrm{BA} & \mathbf{t}_\mathrm{BA} \\ 0 & 1 \end{bmatrix}
\label{equ:trsfm}
\end{equation}
The $t_\mathrm{th}$ frame in the sensor coordinate employs pose $\mathbf{T}_{\mathrm{WC}_{t}} \in \mathrm{SE}(3)$ to convert scan $\mathbf{P}_t$ to the world coordinate frame. We utilize the first frame $\mathbf{P}_0$ as reference frame, so $\mathbf{T}_{\mathrm{WC}_{0}}$ is defined as an identity matrix or a constant matrix.

\subsection{Mapping via Implicit Neural Representation}
\label{sec: map}
\subsubsection{\textbf{Ray Sampling}}
\label{sec:map:raysamp} The first step in mapping framework is the ray sampling along the ray. We perform sampling at two zones: \textbf{(i)} One sampling zone is close to the surface of the objects, and \textbf{(ii)} The other zone is the random-selected free space. Assuming the original input point cloud is $\mathbf{P}_t$, and its sampled point cloud is $\mathbf{P}_{\mathrm{C}_t}$. Then it is converted from sensor coordinate frame to world coordinate frame via pose $\mathbf{T}_{\mathrm{WC}_t}$, which is defined as 
\begin{equation}
\mathbf{T}_{\mathrm{WC}_t} = \mathbf{T}_{\mathrm{WC}_0}\mathbf{T}_{\mathrm{C}_0\mathrm{C}_1}\mathbf{T}_{\mathrm{C}_1\mathrm{C}_2}\mathbf{T}_{\mathrm{C}_2\mathrm{C}_3}...\mathbf{T}_{\mathrm{C}_{t-1}\mathrm{C}_t}
\end{equation}
\begin{equation}
\mathbf{P}_{\mathrm{W}_t} = \mathbf{T}_{\mathrm{WC}_t}\mathbf{P}_{\mathrm{C}_t}
\end{equation}
$\mathbf{P}_{\mathrm{W}_t}$ is corresponding point of $\mathbf{P}_{\mathrm{C}_t}$ in the world coordinate frame. Same to PIN-SLAM \cite{pan2024pin}, our affine transformation needs the homogeneous coordinate conversion of points, but this operation is not explicitly derived. All sampled points $\mathbf{P}_{\mathrm{W}_t}$ in world coordinate frame are fed to construct hierarchical latent features, and distances from each sampled point to the endpoint of the ray are deemed as the ground truth SDF value to self-supervise the training of these latent features. For the rest of the paper, SDF value is denoted as $\varepsilon$, and the function to obtain this SDF value is abbreviated as $\varepsilon(\cdot)$ for the simplification.   

\subsubsection{\textbf{Hierarchical Implicit Feature Embedding}} 
\begin{figure}[t!]
    \centering
    \includegraphics[width=1\linewidth]{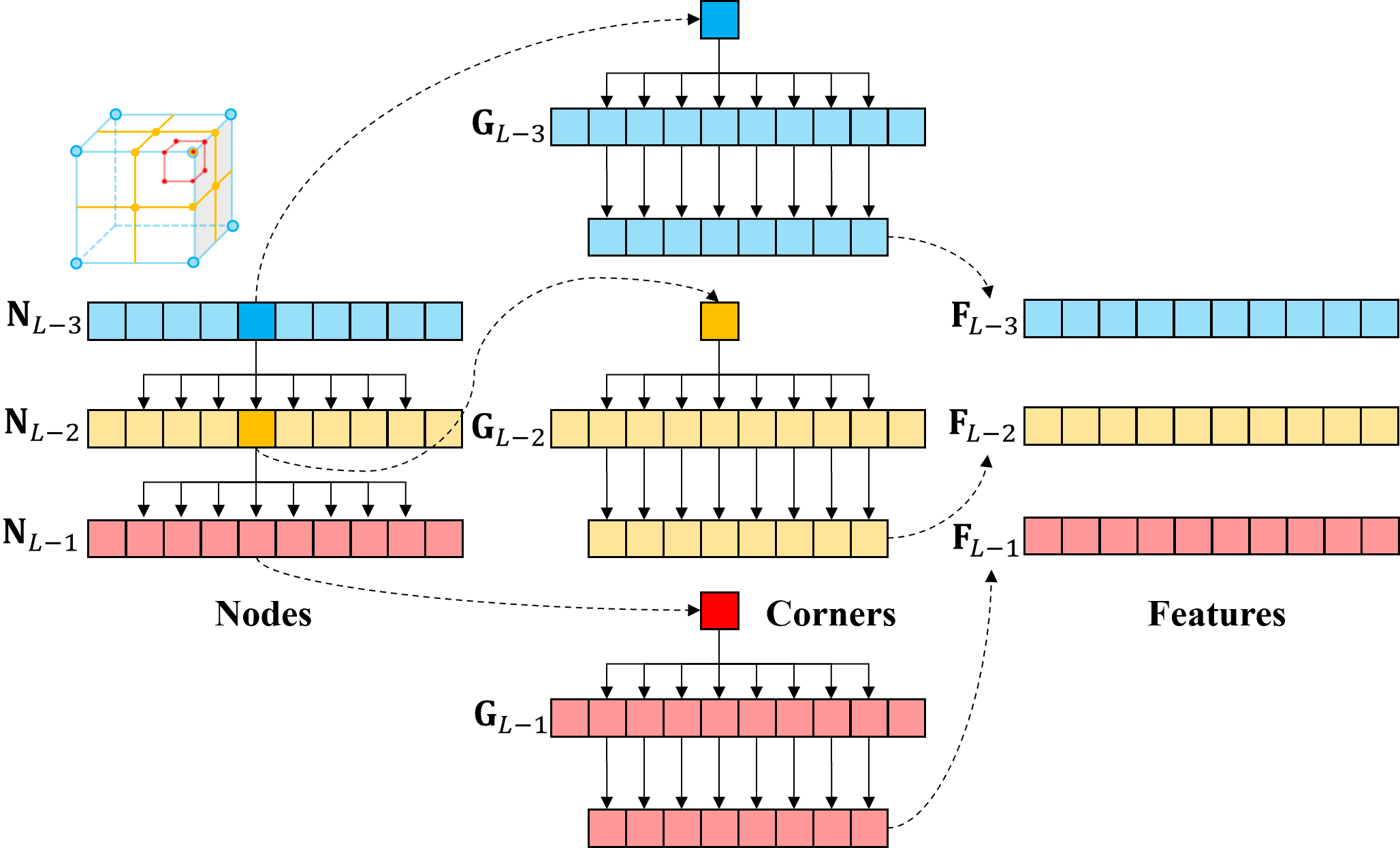}
    \caption{The correspondence among the three tables in certain level of octree. The eight corners of a node correspond to eight features stored in table $\mathbf{F}$. The hierarchical nodes and their corresponding corners are stored in Morton code form in tables $\mathbf{N}$ and $\mathbf{G}$.  
    }
    \label{fig:lookup_table}
\end{figure}

After the sampling along the ray, coordinates of sampled points $\mathbf{P}_{\mathrm{W}_t}$ are scaled to the range of $[-1,1]$ according to the scaling factor $\sigma$ which is related to the number of octree levels $L$. The reason to choose $L$ and $L_f$ is according to the scene size, it will be further discussed in sensitivity analysis in Section \ref{sec:exp:subsec:AblandSensit}. The scaling factor is defined as 
\begin{equation}
    \sigma =\frac{1}{L_f* (2^{L-1})} 
    \label{eq:sigma}
\end{equation}
where $L_f$ denotes the size of leaf nodes. Inspired by the LOD \cite{takikawa2021neural}, we also leverage different levels of feature to improve localization accuracy. The hierarchical feature embedding is generated via an auto-encoder similar to DeepSDF \cite{park2019deepsdf}, but we utilize different number of layers within an octree data structure to produce multi-scale feature embeddings $\mathbf{F}$, which is assembled as $\mathbf{F}=\{\{\mathbf{F}_0\},\{\mathbf{F}_1\}...\{\mathbf{F}_{L-2}\},\{\mathbf{F}_{L-1}\}\}$. As illustrated in the Fig. \ref{fig:lookup_table}, the other two 
tables are produced simultaneously, and their correspondence with feature embeddings are established. Table $\mathbf{N}=\{\{\mathbf{N}_0\},\{\mathbf{N}_1\}...\{\mathbf{N}_{L-2}\},\{\mathbf{N}_{L-1}\}\}$ establishes the relationship between hierarchical nodes and corners, where the Morton code of a node corresponds to the indices of eight corners in the table $\mathbf{G}$. Table $\mathbf{G}=\{\{\mathbf{G}_0\},\{\mathbf{G}_1\}...\{\mathbf{G}_{L-2}\},\{\mathbf{G}_{L-1}\}\}$ establishes the relationship between the corner points and hierarchical features embedding, where the Morton code of a corner point is associated to the index of a single feature embedding in the table $\mathbf{F}$. 

In general, concatenating feature embeddings from more levels of the hierarchy can improve the quality of the map $\mathcal{M}$. However, it involves a trade-off between the mapping quality and the memory consumption. To lower the memory usage and increase the reference speed, we only utilize features from the last three levels. Besides, we employ Morton code as a hash function to correspond a 3D coordinate to a 1D data encoding, which allows the efficient retrieval of features for a given point in an arbitrary position. $\mathbf{F}$ is randomly initialized with Gaussian distribution and is optimized in the training stage, as detailed in section \ref{subsec:TrainOpti}.

The overall procedures to implicitly construct signed distance functions (SDF) value $\varepsilon$ is depicted in the Algorithm \ref{alg:query1}. Given an arbitrary point $\mathbf{p}$ in the 3D space, we only leverage the last three levels of the octree based on the querying of table $\mathbf{N}$, $\mathbf{G}$, and $\mathbf{F}$, trilinear interpolation is adopted to generate multi-scale features. The hierarchical feature embeddings are concatenated to $\mathbf{F}_a$, which is then fed into a multi-layer perceptron (MLP) $\varphi$ to further decode as SDF value $\varepsilon$.

\begin{algorithm}[!t]
    \caption{Inferring SDF Value of a Given Point}
    \label{alg:query1}
    \renewcommand{\algorithmicrequire}{\textbf{Input:}}
    \renewcommand{\algorithmicensure}{\textbf{Output:}}
    
    \begin{algorithmic}[1]
        \REQUIRE Point $\mathbf{p} \in \mathbb{R}^3$ in the 3D space
        \ENSURE SDF value $\varepsilon$ of $\mathbf{p}$, abbreviated as $\varepsilon(\mathbf{p})$
        
        \STATE  Scale $\mathbf{p}$ to [-1,1], transform to a Morton value $E$
        
        \FOR{ nodes $n_k$ in  $\{N_{L-k}\}, k=1, 2, 3 $}
            \IF {$E == n_k$}
                \STATE $\mathbf{G}_{L-k} \leftarrow $ Eight corner points corresponding to $n_k$
            \ELSE
                \STATE $\mathbf{p}$ is outside the map's boundary.
            \ENDIF
        \ENDFOR
        
        \FOR{each $i \in {1, 2, 3}$}

        \STATE $\{\mathbf{f}_{L-i}\} \leftarrow \{\mathbf{F}[\mathbf{G}_{L-i}]\} $ 
        
        \STATE $\mathbf{F}_{L-i} \leftarrow $ Trilinear interpolation for $\{\mathbf{f}_{L-i}\}$

        \ENDFOR
        \STATE $\mathbf{F}_a \leftarrow \mathbf{F}_{L-3}+\mathbf{F}_{L-2}+\mathbf{F}_{L-1}$
        \STATE SDF value of $\mathbf{p}$ : $ \varepsilon \leftarrow \varepsilon(\mathbf{p}) = \varphi(\mathbf{F}_a)$ 
        \RETURN
    \end{algorithmic}
\end{algorithm}
\subsubsection{\textbf{Surface Reconstruction of Implicit Map}} These feature embeddings implicitly represent the SDF values $\varepsilon$ of points $\mathbf{p}$ in the environment. The map visualization is inspired by \cite{zhong2023shine}, the Marching Cubes algorithm is utilized to extract iso-surfaces and generate the mesh. First, we compute the map size by examining the maximum and minimum nodes at the corresponding level. Based on the map size, we create a volume and subdivide it into more fine-grained grids according to the resolution of the Marching Cubes. Then, we apply the Algorithm \ref{alg:query1} to infer the SDF values for newly created grid coordinates and generate a mesh-based map. 

\subsection{Neural Implicit Odometry}
\subsubsection{\textbf{Match Initialization}} For the first scan, we utilize an identity matrix or a constant matrix $\mathbf{T}_{\mathrm{WC}_0}$ to transform the input sampled point cloud $\mathbf{P}_{\mathrm{C}_0}$ from sensor coordinate frame to world coordinate frame $\mathbf{P}_{\mathrm{W}_0}$, and employ $\mathbf{P}_{\mathrm{W}_0}$ to initialize the map. For the second scan, we initialize the pose with a given translation value $\mathbf{t}_{\mathrm{WC}_1}$, which is based on the robot's movement direction. This translation is typically set as 0.01 times the bounding box range. The bounding box is adopted to trim off point clouds that lie outside a certain range, as the range accuracy of point cloud data tends to be ambiguous with the distance from the sensor increasing. For following poses $\hat{\mathbf{T}}_{\mathrm{WC}_t}$, the Constant Move Model is adopted to initialize them. In this way, the pose of the current frame, $\hat{\mathbf{T}}_{\mathrm{WC}_t}$, can be simply estimated by poses from previous two frames $\mathbf{T}_{\mathrm{WC}_{t-1}}$ and $\mathbf{T}_{\mathrm{WC}_{t-2}}$, which is shown in the following equation and illustrated in Fig. \ref{fig:sysoverview}.
\begin{equation}
     \hat{\mathbf{T}}_{\mathrm{WC}_t} =\mathbf{T}_{\mathrm{WC}_{t-1}}  (\mathbf{T}_{\mathrm{WC}_{t-2}} ^{-1} \mathbf{T}_{\mathrm{WC}_{t-1}})
     \label{eq:pre}
\end{equation}

\subsubsection{\textbf{Scan-to-implicit-map Matching}} Overall, we employ multi-scale feature embedding to execute the registration to achieve an efficient and precise pose estimation. Inspired by \cite{wiesmann2023locndf}, we perform scan registration within the neural signed distance field (nSDF) based on the second-order optimization.

As shown in Fig. \ref{fig:sysoverview}, our goal is to search the optimal pose $\mathbf{T}_{\mathrm{WC}_t}^*$, made up by rotation $\mathbf{R}_{\mathrm{WC}_t}^*$ and translation $\mathbf{t}_{WC_t}^*$, that aligns the point cloud $\mathbf{P}_t$ to the nSDF of the submap ${\mathcal{M}_s} \subset \mathcal{M}$, which can minimize the least square error of the SDF prediction $\hat{\varepsilon}$. Note that $\mathbf{P}_t$ is the original point cloud, ray sampling is not executed for the odometry stage. $\mathcal{M}_s$ is the submap which is generated every other $s$ scans, acting as the local map for matching. $\mathcal{M}$ is the implicit global map, its feature embeddings implicitly represent the SDF value distribution of the environment. And the above procedures are depicted as the following 
\begin{equation}
    \mathbf{R}_{\mathrm{WC}_t}^*,\mathbf{t}_{\mathrm{WC}_t}^*=\mathop{\min}\limits_{\mathbf{R}_{\mathrm{WC}_t},\mathbf{t}_{\mathrm{WC}_t}} \sum_{\mathbf{p} \in \mathbf{P}_{t}} ||\varepsilon(\mathbf{R}_{\mathrm{WC}_t}\mathbf{p}+\mathbf{t}_{\mathrm{WC}_t})||^2
    \label{eq:LM}
\end{equation}
To solve the Equation (\ref{eq:LM}), we implement the Levenberg-Marquardt (LM) algorithm \cite{ranganathan2004levenberg} as the optimization method, and use the estimated poses $\hat{\mathbf{T}}_{\mathrm{WC}_t}$ from Equation (\ref{eq:pre}) as the initial pose for the optimization. As shown in Fig. \ref{fig:sysoverview}, the feature embeddings and the MLP parameters are frozen during the iterative optimization of the pose, only the pose is fine-tuned. The specific optimization process is detailed in \ref{subsubsec:Odometry}.

From the Equation (\ref{eq:LM}), our method prevents the use of the explicit point-to-point or point-to-plane correspondence as routines of typical ICP-based methods. 
In the process of obtaining the SDF values $\varepsilon$, we adopt multi-scale feature embeddings rather than a singular scale feature embedding. This strategy is anticipated to improve the localization accuracy by following rationales. \textbf{(i)}. Hierarchical features can represent environmental characteristics at different scales, which are able to adaptively describe the geometric properties of the environment. \textbf{(ii)}. At certain hierarchical scales, the features tend to ignore or reduce the influence of dynamic objects. 

Based on the precise implicit representation of environment, the scan is designed to match with the implicit submap $\mathcal{M}_s$  rather than the global map $\mathcal{M}$. This is attributed to the fact that there are similarities between different places of the environment, resulting in implicit features in the global map resemble with each other, which potentially disrupt the matching accuracy, hence we leverage submaps to mitigate this kind of noise.

\subsubsection{\textbf{Submap to Global Map}} For the map building, we only maintain a global map. For the pose estimation, we keep creating submaps $\mathcal{M}_s$, and only features in the submaps are optimized during the training phase of odometry. As described in Algorithm \ref{alg:query}, three tables, $\mathbf{F}_s \subset \mathbf{F}$, $\mathbf{N}_s \subset \mathbf{N}$, $\mathbf{G}_s \subset \mathbf{G}$, are established along the submap $\mathcal{M}_s$. Next, to merge the submaps into the global map, we separately check whether $\mathbf{N}_s$, $\mathbf{G}_s$ are existed in $\mathbf{N}$, $\mathbf{G}$. If not, we add corresponding $\mathbf{N}_s$, $\mathbf{G}_s$ into $\mathbf{N}$, $\mathbf{G}$ and add corresponding feature embedding $\mathbf{F}_s$ into $\mathbf{F}$. After this process is completed, we clear these three tables $\mathbf{N}_s$, $\mathbf{G}_s$, $\mathbf{F}_s$ to store the following generated submaps.

\begin{algorithm}[!t]
    \caption{Merging Submaps to Global Map}
    \label{alg:query}
    \renewcommand{\algorithmicrequire}{\textbf{Input:}}
    \renewcommand{\algorithmicensure}{\textbf{Output:}}
    
    \begin{algorithmic}[1]
        \REQUIRE $\mathbf{N}$, $\mathbf{G}$, $\mathbf{F}$, $\mathbf{N}_s$, $\mathbf{G}_s$, $\mathbf{F}_s$
        \ENSURE   $\mathbf{N}$, $\mathbf{G}$, $\mathbf{F}$, $\mathbf{N}_s'$, $\mathbf{G}_s'$, $\mathbf{F}_s'$

            \FOR{node $n$ in $\mathbf{N}_{s_{L-k}}$, $ k = 1,2,3$}

                      \IF {$n \notin  \mathbf{N}_{L-k}$}

                         \STATE Insert new node $n$ into Table $\mathbf{N}_{L-k}$
                     \ENDIF
                
            \ENDFOR   
                  
            \FOR{corner $g$ in $\mathbf{G}_{sL-k}$, $ k = 1,2,3$}
            
            \IF{$g \notin \mathbf{G}_{L-k}$}
            \STATE Put new corners $g$ into table $\mathbf{G}_{L-k}$
            \STATE i $\leftarrow$ the index of $g$
            \STATE Put new feature embedding $\mathbf{F}_{s_{L-k}}$[i] into table $\mathbf{F}_{L-k}$
            \ENDIF
            
            \ENDFOR
            \STATE $\mathbf{N}_s', \mathbf{G}_s', \mathbf{F}_s'$ $\leftarrow$ Release contents of Tables $\mathbf{N}_s, \mathbf{G}_s, \mathbf{F}_s$
        \RETURN
    \end{algorithmic}
\end{algorithm}

\subsection{Training and Optimization} 
\label{subsec:TrainOpti}
\subsubsection{\textbf{Optimization for the Odometry}}
\label{subsubsec:Odometry}
As mentioned above, we leverage LM method \cite{ranganathan2004levenberg} to solve equation (\ref{eq:LM}). The optimization procedure is further explained here. Firstly, we calculate a Jacobian matrix $\mathbf{J}$ as below:
\begin{equation}
    \mathbf{J}=[\frac{ \partial \varepsilon(\overline{\mathbf{p}}) }{ \partial \theta }, \frac{\partial \varepsilon(\overline{\mathbf{p}}) }{ \partial \mathbf{t}_{\mathrm{WC}_t} }]=[(\frac{\overline{\mathbf{p}}}{\sigma}) \times (d*\sigma),d*\sigma]
    \label{eq:J}
\end{equation}
where $\overline{\mathbf{p}}= \mathbf{R}_{\mathrm{WC}_t}\mathbf{p}+\mathbf{t}_{\mathrm{WC}_t}$, $\sigma$ is a scaling factor as defined in the equation (\ref{eq:sigma}), $\theta$ is the axis-angle parameter of $\mathbf{R}_{\mathrm{WC}_t}$, $d=\nabla \varepsilon (\overline{\mathbf{p}})$ is the distance gradient at $\overline{\mathbf{p}}$, which can be inferred from our nSDF by automatic differentiation. For the Jacobian matrix in equation (\ref{eq:J}), we rescale the points' coordinate back to their original scale. Since the input points are scaled, gradients are also rescaled correspondingly to ensure that the pose possesses the physical significance in the real-world environment. After obtaining the Jacobian matrix, the Hessian matrix can be directly obtained as $\mathbf{H}=\mathbf{J}^\mathrm{T}\mathbf{J}$. The step increment, $\Delta \mathbf{h}$, of transformation parameters is calculated as
\begin{equation}
    \Delta \mathbf{h} = (\mathbf{H} + \lambda_e \mathrm{diag}(\mathbf{H}))^{-1} \mathbf{\gamma}
\end{equation}
where $\mathbf{\gamma}$ is the residual vector, $\lambda_e$ is the damping parameter for Levenberg-Marquardt optimization. By fine-tuning the value of step increment, the pose can be computed iteratively.

\subsubsection{\textbf{Optimization for the Mapping}}
\label{subsubsec:optMap}
A self-supervised approach is adopted for the training within the mapping, as the LiDAR provides precise range information. Therefore, the distance from the sampled points to the endpoints is defined as the ground truth SDF, $\varepsilon^{gt}$ , for the supervision. Totally $b$ points along each laser beam are randomly sampled as described in Section \ref{sec:map:raysamp}. Binary cross-entropy is exploited as the loss function for SDF values. To be noticed, a replay strategy same to PIN-SLAM \cite{pan2024pin} is implemented to solve the ‘catastrophic
forgetting’ problem in the incremental learning. A local data pool $DP_t$ is constructed via accumulating history LiDAR scans $\{\mathbf{P}_{{\mathrm{C}_{t-j}}} 
| j=1,2,3,...,w\}$, $w$ is the size of sliding window. Then, training points $\mathbf{P}_{{\mathrm{W}_t}}$ is made up by two parts, one part is randomly sampled from points of current scan $\mathbf{P}_{\mathrm{C}_t}$, another part is from points of data pool $DP_t$. Given a sampled point $\mathbf{p}_i \in \mathbf{P}_{{\mathrm{W}_t}}$ and signed distance to the surface $\varepsilon^{gt}_i$, cross-entropy loss $\mathcal{L}_1$ is denoted as
\begin{equation}
\begin{aligned}
\mathcal{L}_1 
&= \Psi(\varepsilon^{gt}_i)log(\Psi(\varepsilon(\mathbf{p}_i)))\\ 
& + (1-\Psi(\varepsilon^{gt}_i))log(1-\Psi(\varepsilon(\mathbf{p}_i)))
\end{aligned}
\end{equation}
where $\Psi(x)=1/(1+e^{x / \alpha})$ is the activation function and the $\alpha$ is a hyper-parameter. $\varepsilon(\mathbf{p}_i)$, which is equal to $\varphi(F_a)$, stands for the SDF value of the MLP output at point $\mathbf{p}_i$. Eikonal regularization \cite{zhong2023shine} is applied to add another term, $\mathcal{L}_2$, into the loss function
\begin{equation}
    \mathcal{L}_2=| \|\bigtriangledown_{\mathbf{p}_i} \varepsilon(\mathbf{p}_i) \|-1|
\end{equation}
The loss $\mathcal{L}_3$ is adopted to enforce the gradient direction, at the point $\mathbf{p}_i$, to be aligned with the normal direction 
\begin{equation}
    \mathcal{L}_3= \angle (\nabla \varepsilon(\mathbf{p}_i), \phi)
\end{equation}
where $\phi$ is the normal of point $\mathbf{p}_i$. It is important to note that we apply loss function $\mathcal{L}_3$ only to the sampled points close to the objects' surface.

The final loss is formed as a weighted sum of binary cross-entropy term, Eikonal term , and direction term. 
\begin{equation}
    \mathcal{L}=  \mathcal{L}_1 +  \lambda_1*\mathcal{L}_2 + \lambda_2*\mathcal{L}_3
\end{equation}
where $\lambda_1$, $\lambda_2$ are hyper-parameters used to control the weights of different loss functions.

\begin{table*}[b!]
\centering 
\caption{\textbf{Comparison of Average Relative Translational Drifting Error (\%) on \textit{KITTI} Odommetry Benchmark.} The best results are highlighted in bold, and the second-best results are underlined. ``-" indicates the result is not reported in the related paper, `-LO' in the name of methods indicates that only the odometry component of the method is used for evaluation. And the definition of those symbols are applicable to other tables in the rest of the paper. Consistent with other methods, we use the average relative translational error in \% as the metric for evaluating odometry drift.}
\resizebox{0.985\textwidth}{!}{
  \begin{tabular}{cc|ccccccccccc|c}
    \toprule
    Method & Type & \text{00} & \text{01} & \text{02} & \text{03}&  \text{04} & \text{05} & \text{06} & \text{07} & \text{08} & \text{09} & \text{10} & \textbf{Avg.} \\ 
    \midrule
    LeGO-LOAM \cite{shan2018lego} & feature points & 2.17 & 13.4 & 2.17 & 2.34 & 1.27 & 1.28 & 1.06 & 1.12 & 1.99 & 1.97 & 2.21 & 2.49   \\
    F-LOAM~\cite{wang2021f} & feature points   & 0.78 & 1.43 &	0.92 & 0.86 &	0.71 & 0.57 &	0.65 & 0.63 &	1.12 & 0.77	& 0.79 & 0.84  \\
    MULLS~\cite{pan2021mulls}  & feature points   & 0.56 & \underline{0.64}	& 0.55 & 0.71 &	0.41 & \underline{0.30} &	0.30 & 0.38 &	\bf{0.78} & \bf{0.48} &	0.59 & 0.52  \\
    VG-ICP~\cite{koide2021voxelized} & dense points   & 2.16 & 2.38	& 0.99 & 0.67	& 0.55 & 0.45	& \bf{0.24} & 0.99 &	1.74 & 0.95	& 0.95 & 1.10  \\
    CT-ICP~\cite{dellenbach2022ct} & dense points  & \bf{0.49} & 0.76 & \underline{0.52} & 0.72 & 0.39 & \bf{0.25} & 0.27 & \bf{0.31} & 0.81 & \underline{0.49} & \underline{0.48} & \underline{0.50}  \\
    KISS-ICP~\cite{vizzo2023kiss} & dense points  & 0.52 & \bf{0.63} & \bf{0.51} & 0.66 &	0.36 & 0.31 &	\underline{0.26} & \underline{0.33} &	0.82 & 0.51 &	0.56 & \underline{0.50}  \\
    SuMa-LO~\cite{behley2018efficient} & surfels & 0.72	& 1.71 & 1.06 &	0.66 & 0.38 &	0.50 & 0.41 &	0.55 & 1.02 &	0.48 & 0.71	& 0.75  \\
    SuMa++~\cite{chen2019suma++} & surfels & 0.64	& 1.60 & 1.00 &	0.67 & 0.37 &	0.40 & 0.46 &	0.34 & 1.10 &	0.47 & 0.66	& 0.70  \\
    Litamin-LO~\cite{yokozuka2021litamin2} & normal distribution &  0.78 & 2.10 & 0.95 & 0.96 & 1.05 & 0.55 & 0.55 & 0.48 & 1.01 & 0.69 & 0.80 & 0.88 \\
    IMLS-SLAM~\cite{deschaud2018imls} & implicit model & \underline{0.50} & 0.82 & 0.53 & 0.68 & 0.33 & 0.32 & 0.33 & \underline{0.33} & \underline{0.80} & 0.55 & 0.53 & 0.52 \\
    Puma~\cite{vizzo2021poisson}  & mesh   & 1.46 & 3.38 &	1.86 & 1.60 &	1.63 & 1.20 &	0.88 & 0.72 &	1.44 & 1.51 &	1.38 & 1.55  \\
    SLAMesh~\cite{ruan2023slamesh} & mesh  & 0.77 & 1.25 &	0.77 & 0.64 &	0.50 & 0.52 &	0.53 & 0.36 &	0.87 & 0.57 &	0.65 & 0.68  \\
    Mesh-LOAM \cite{zhu2024mesh}& mesh  & 0.53 & \underline{0.64} &	0.52 & \underline{0.63} &	0.41 & 0.34 &	0.35 & \bf{0.31} &	0.82 & 0.47 &	0.56 & 0.51  \\
    \midrule
    LONet~\cite{li2019net} & supervised & 1.47 & 1.36 & 1.52 & 1.03 & 0.51 & 1.04 & 0.71 & 1.70 & 2.12 & 1.37 & 1.80 & 1.33 \\
    TransLO~\cite{liu2023translo} & supervised & 0.85 & 1.16 & 0.88 & 1.00 & 0.34 & 0.63 & 0.73 & 0.55 & 1.29 & 0.95 & 1.18 & 0.87 \\
    PWCLONet~\cite{wang2021pwclo} & supervised & 0.78 & 0.67	& 0.86 & 0.76	& 0.37 & 0.45	& 0.27 & 0.60 &	1.26 & 0.79	& 1.69 & 0.77  \\
    NeRF-LOAM~\cite{deng2023nerf} & neural implicit & 1.34 & 2.07 & -    & 2.22 &	1.74 & 1.40 &  -   & 1.00 &  -	   & 1.63 &	2.08 & 1.69 \\
    PIN-LO \cite{pan2024pin} &  neural implicit & 0.55 &	0.68	& 0.54 &	0.76 &	\bf{0.22} & \underline{0.30} &	0.35 &	0.34 &	\underline{0.80}	& 0.54	& 0.50	& 0.51  \\
    \midrule
    Hi-LOAM &  neural implicit & 0.51 &	\bf{0.63}	& 0.53 &	\bf{0.61} &	\underline{0.31} &	0.31 &	0.31 &	0.35 &	0.81	& \underline{0.49}	& \bf{0.45}	& \bf{0.48}  \\

    \bottomrule

  \end{tabular}
  } 
  \vspace{-2pt}
  \label{tab:kitti_odom}
\end{table*}


\section{Experiment}
In this section, the experimental result and analysis are presented based on the comparison between our \FrameworkNM{} and state-of-the-art LOAM approaches on various public datasets, both the quantitative and qualitative results are demonstrated, which shows the superior effectiveness and robustness of our framework.

\subsection{Experiment Setup}
\subsubsection{\textbf{Implementation Details}} Our project is built on top of the PyTorch \cite{paszke2019pytorch}. The size of leaf node is set as $L_f = 0.2 $ m, the number of octree level is set based on the scene scale, the typical number are $L = 10$, $L = 13$, and $L = 15$. The rationale to set the $L$ is given in Section \ref{sec:exp:subsec:AblandSensit}. The number of randomly-sampling points along the ray is set as $b = 6$, specifically, 2 points in free-space area and 4 points near the objects' surface area. The weight coefficients are set as $\lambda_1$ = 0.1, $\lambda_2 = 0.1 $. For odometry of \textit{KITTI} dataset, the size of the submap is set as $s = 200$. For odometry of other datasets, the size of the submap is set between 50 to 100. The rationale to set the $s$ is given in Section \ref{sec:exp:subsec:AblandSensit}. For both of the mapping and the localization phase, we iterate $100$ rounds for the optimization of each frame with a learning rate of 0.0001, and adopt Adam optimizer for the mapping phase and implement LM algorithm for odometry phase. We utilize voxel down-sampling strategy to extract the input point cloud. For mapping, we use the voxel size of 0.1 m for down-sampling, and for the odometry part, we adopt a voxel size of 0.2 m. The number of hidden layers for MLP is set as 2, and the size of each hidden layer is 32. The length of a single feature vector $\mathbf{F}$ is chosen to be 12. The hyper-parameter $\alpha$ is set as 0.08.

\subsubsection{\textbf{Datasets}} We validate our method on various public datasets, including \textit{KITTI} \cite{geiger2012we}, \textit{SemanticPOSS} \cite{pan2020semanticposs}, \textit{Newer College} \cite{ramezani2020newer}, \textit{Hilti-21} \cite{2109.11316}, \textit{Hilti-23} \cite{nair2024hiltislamchallenge2023}, \textit{MulRAN} \cite{gskim-2020-mulran}, and \textit{Mai City} \cite{vizzo2021poisson}. The \textit{KITTI} is a large-scale outdoor dataset collected by Velodyne HDL-64E S2 LiDAR, which comprises 22 sequences of LiDAR scans, we evaluate our odometry accuracy on sequences 00 to 10 because these 11 sequences provide ground truth trajectories. The \textit{SemanticPOSS} comprises 6 sequences of LiDAR scans, and totally contains 2988 complicated LiDAR scans with large quantity of dynamic object instances. To further examine our approach, we utilize the \textit{Newer College} and the \textit{Hilti} datasets, which are recorded through a hand-held LiDAR platform, resulting in less constant motion patterns. The \textit{Newer College} dataset is collected using two different sensors, the Ouster OS-1 64 LiDAR and the Ouster OS-0 128 LiDAR, and collected with hand-held devices at typical walking speeds. The \textit{Hilti-21} dataset stands for Hilti SLAM Dataset 2021, includes both indoor and outdoor scenes, collected using a handheld Ouster OS0-64 LiDAR, and \textit{Hilti-23} is an extended version collected from multiple active construction site environments. In addition to using real-world datasets, we also try a synthetic dataset to evaluate our odometry and mapping performance. \textit{Mai city} is a synthetic dataset consisting of 3 sequences, using a virtual Velodyne HDL-64 LiDAR as the sensor.



\subsection{Odometry Evaluation}
\subsubsection{\textbf{Odometry Evaluation with SOTA Methods}} 
\begin{figure}[h!] 
    \centering 
    
    \includegraphics[width=0.48\textwidth]{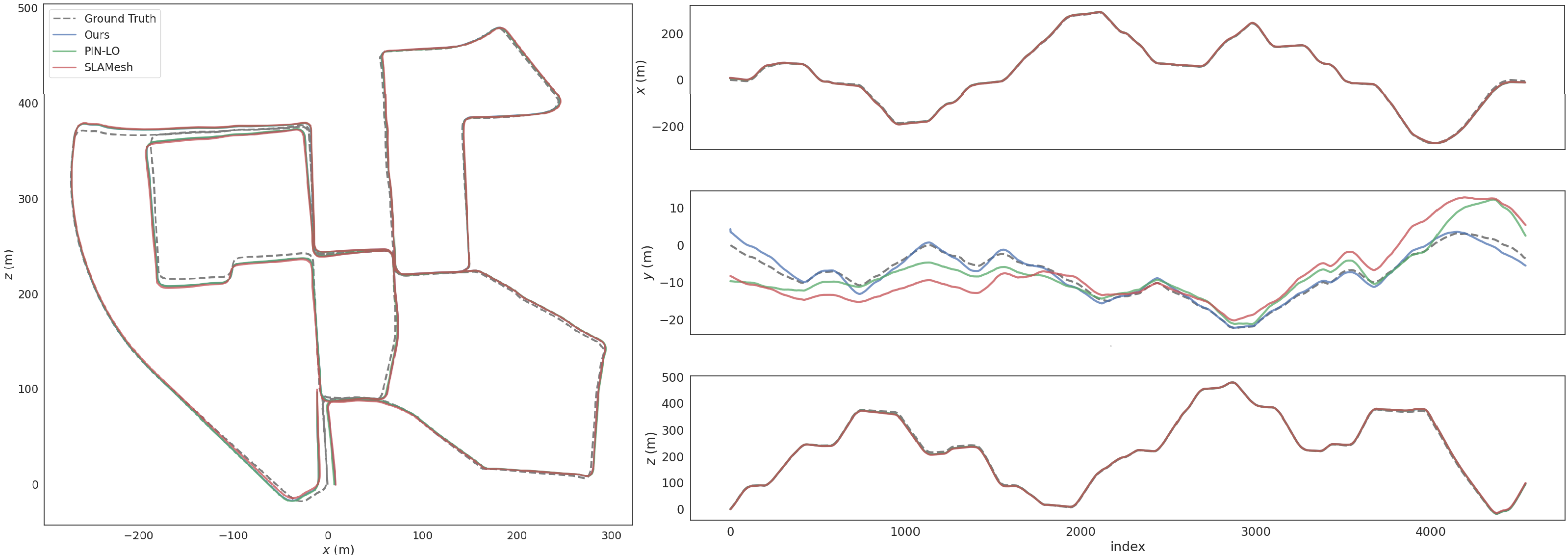} 
    \caption{
    \textbf{The Qualitative Results of Odometry Comparison on Sequence 00 of \textit{KITTI} Dataset.} The left side displays trajectories comparison of different methods. The right side compares errors along the X, Y, and Z axes among our method and other methods. The black dashed line represents the ground truth trajectory. (\textit{Best viewed with zoom-in.}) 
    }
    \label{fig:kitti00_traj} 
\end{figure}

To evaluate the performance of our localization task, we conduct comparative experiments on the \textit{KITTI} dataset with various state-of-the-art SLAM or odometry methods which utilize different map representation techniques, such as point cloud, neural points, and mesh. The quantitative experiment results on the \textit{KITTI} dataset are reported in Table \ref{tab:kitti_odom}. Distinguished from other supervised learning-based methods, such as TransLO \cite{liu2023translo}, PWCLONet \cite{wang2021pwclo} etc., which are built on top of the pre-training,
\begin{table*}[b!]
\centering 
\caption{\textbf{Comparison of Absolute Trajectory Error (m) on \textit{KITTI} Odometry Benchmark.}}
\resizebox{0.985\textwidth}{!}{
  \begin{tabular}{cc|ccccccccccc|c}
    \toprule
    Method & Type & \text{00} & \text{01} & \text{02} & \text{03} & \text{04} & \text{05} & \text{06} & \text{07} & \text{08} & \text{09} & \text{10} & 
    \textbf{Avg.} \\  
    \midrule

    LeGO-LOAM \cite{shan2018lego} & feature points & \bf{2.3} & 19.7 & 5.3 & 1.6 & 0.4 & \bf{1.2} & 1.0 & 1.5 & 5.9 & 2.0 & 1.7 & 5.3   \\    

    F-LOAM~\cite{wang2021f} & feature points   & 5.0 & 3.2 &	8.6 & \underline{0.7} &	\underline{0.3} & 3.4 &	0.5 & 0.6 &	3.1 & 1.6	& 1.2 & 2.6 \\

    KISS-ICP~\cite{vizzo2023kiss} & dense points  & 5.7 & 30.7 & 17.8 & 3.4 &	1.1 & 1.9 &	0.9 &0.8 &	4.8 & 3.7 &	2.3 & 6.6  \\
    SuMa~\cite{behley2018efficient} & surfels &\underline{2.9}	& 13.8 & \underline{8.4} &	0.9 & 0.4 &	\bf{1.2} & \underline{0.4} &	\underline{0.5} & \underline{2.8} &	2.9 & 1.3	& 3.2  \\

    Litamin~\cite{yokozuka2021litamin2} & normal distribution &  5.8 & 15.9 & 10.7 & 0.8 & 0.7 & 2.4 & 0.9 & 0.6 &\bf{2.5} & 2.1 & 1.0 & 3.9 \\

    Puma~\cite{vizzo2021poisson}  & mesh   & 6.6 & 32.6 &	18.5 & 2.2 &	0.9 & 3.3 &	2.4 & 0.9 &	6.3 & 3.9 &	4.4 & 7.5  \\
    SLAMesh~\cite{ruan2023slamesh} & mesh  & 5.5 & 10.9 &	13.2 & 0.8 &	\underline{0.3} & 3.7 &	0.7 & 0.8 &	5.1 & \bf{1.0} &	1.1 & 3.9  \\
    Mesh-LOAM \cite{zhu2024mesh}& mesh  & 5.3 & 3.0 &	\bf{7.4} & \bf{0.5} &	\underline{0.3} & 1.7 &	\bf{0.3} & \bf{0.4} &	3.3 & \underline{1.7} &	\underline{0.9} & \underline{2.3}  \\

     PIN-LO \cite{pan2024pin} &  neural implicit & 5.6 &	\underline{4.3}	& 9.3 &	\underline{0.7} &	\bf{0.1} & 1.7 &	0.5 &	\underline{0.5} &	3.0	& 1.8	& \bf{0.8}	& 3.2\\     
    \midrule
    Hi-LOAM &  neural implicit & 5.0 &	\bf{2.0}	& 8.6 &	\bf{0.5} & \bf{0.1} & \underline{1.5} &	0.4 &	\underline{0.5} &	\underline{2.8}	& 1.8	& \bf{0.8}	& \bf{2.1}  \\

    \bottomrule

  \end{tabular}
  } 
  \vspace{-2pt}
  \label{tab:kitti_ATE}
\end{table*}
\begin{table*}[b!]
  \centering
  \caption{\textbf{Localization Performance Comparison (ATE RMSE [m]) on \textit{MulRAN} LiDAR Dataset.}} 
  \label{tab:ate_mulran}
  \setlength{\tabcolsep}{12pt} 

      \begin{tabular}{c|ccccccccc|c}
          \toprule
          Method   & \text{KA1}     & \text{KA2}   & \text{KA3} & \text{DC1} & \text{DC2} & \text{DC3}  & \text{RS1}   & \text{RS2}     & \text{RS3}  & \textbf{Avg.}   \\
          \midrule
          F-LOAM~\cite{wang2021f} & 49.16	& 35.93 &	38.36 &	36.89 &	26.72 &	31.97 &	91.85 &	104.25 &	119.79 & 59.44 \\ 
          KISS-ICP~\cite{vizzo2023kiss} & 20.81 & \underline{13.52} & 14.85 & 20.26 & \underline{12.79} & 13.17 &  \bf{14.35} & \bf{32.74} & 75.14 & \underline{24.18} \\
          \midrule
          SuMa~\cite{behley2018efficient} & \bf{10.45} & \bf{9.33} & \underline{10.33} & \bf{13.00} & 86.04 & \underline{12.03} & 114.89 & 818.74 & 367.41 & 160.25 \\ 
          MULLS~\cite{pan2021mulls} & \underline{19.72} & 15.26 & \bf{6.93} & \underline{14.95} & \bf{11.31} & \bf{6.00} & \underline{41.25}  & \underline{45.05} & \underline{44.18} &  \bf{22.74} \\

          \midrule
          PIN-LO~\cite{pan2024pin} & 29.07 & 24.74 & 23.05 & 22.07 & 12.94 & 23.42 & 45.81 & 54.41 & 44.41 &	31.10  \\

         Hi-LOAM & 18.73 & 20.80 & 25.98 & 20.20 & 16.38 & 15.42 & 42.36 & 52.14 & \bf{38.95} &	27.88  \\
          \bottomrule
      \end{tabular}
\end{table*}
NeRF-LOAM \cite{deng2023nerf}, PIN-LO \cite{pan2024pin} and our approach are learning-based implicit methods which are waived from the pre-training. As can be seen, our results outperform all other learning-based and non-learning-based methods for the average value across 11 sequences, this is due to the strong modeling ability of the hierarchical feature embeddings and careful tuning of the optimization method. For better reflecting the global consistency of the estimated trajectory, we also calculate the Root Mean Square Error (RMSE) of the Absolute Trajectory Error (ATE) \cite{zhang2018tutorial} as the metric, to evaluate the performance of the odometry, the result is illustrated in Table \ref{tab:kitti_ATE}. The listed results of other algorithms are either reported in their original paper or re-executed by us if they are open-sourced, we achieve slightly better results than implicit localization method like PIN-LO, and our average results are better than ICP-based methods as well. Additionally, the qualitative results for the localization, from the KITTI dataset, is shown in Fig. \ref{fig:kitti00_traj}.
\begin{table}[h!]
  %
  \centering
  \caption{\textbf{Localization Performance Comparison (ATE RMSE [m]) on \textit{SemanticPOSS} LiDAR Dataset.}}
  \label{tab:ate_poss}
  \resizebox{0.48\textwidth}{!}{
      \begin{tabular}{r|cccccc|c}
        \toprule
          Method  & \text{00} & \text{01}  & \text{02}  & \text{03} & \text{04} & \text{05} & \textbf{Avg.}  \\
          \midrule
          SLAMesh~\cite{ruan2023slamesh} & 0.28 & 0.56  & 0.28 & 0.29 & 0.38 & 0.33 & 0.35 \\
          MULLS~\cite{pan2021mulls} & 0.38 & 0.55 & 0.26 & 0.35 & 0.38 & 0.43 & 0.39 \\
          PIN-SLAM~\cite{pan2024pin} & \bf{0.25} & \underline{0.39} & \bf{0.19} & \underline{0.26} & \underline{0.24} & \underline{0.27} & \underline{0.27}\\
                    \midrule
          Hi-LOAM & \underline{0.27} & \bf{0.29} & \underline{0.22} & \bf{0.25} & \bf{0.23} & \bf{0.23} & \bf{0.25}
          \\
          \bottomrule
      \end{tabular}
  }
\end{table}

\subsubsection{\textbf{Adaptability Test for Scenes Variety}} To evaluate our localization performance in highly dynamic environment and demonstrate the adaptability of our method, another set of experiment is conducted via utilizing \textit{SemanticPOSS} dataset, which contains various and complicated LiDAR scans with large quantity of dynamic instances. As shown in Table \ref{tab:ate_poss}, compared with other state-of-the-art LiDAR SLAM methods, we achieve favorable results. It demonstrates the robustness of our method in dynamic environments, and one possible explanation for performance boost of our method is that the fusion of multi-scale features may help to mitigate the impact of dynamic objects. 

\begin{figure*}[t!]
\begin{scriptsize}  
   \begin{tabular}{cccccc}
      \makecell{\includegraphics[width=.166\linewidth]{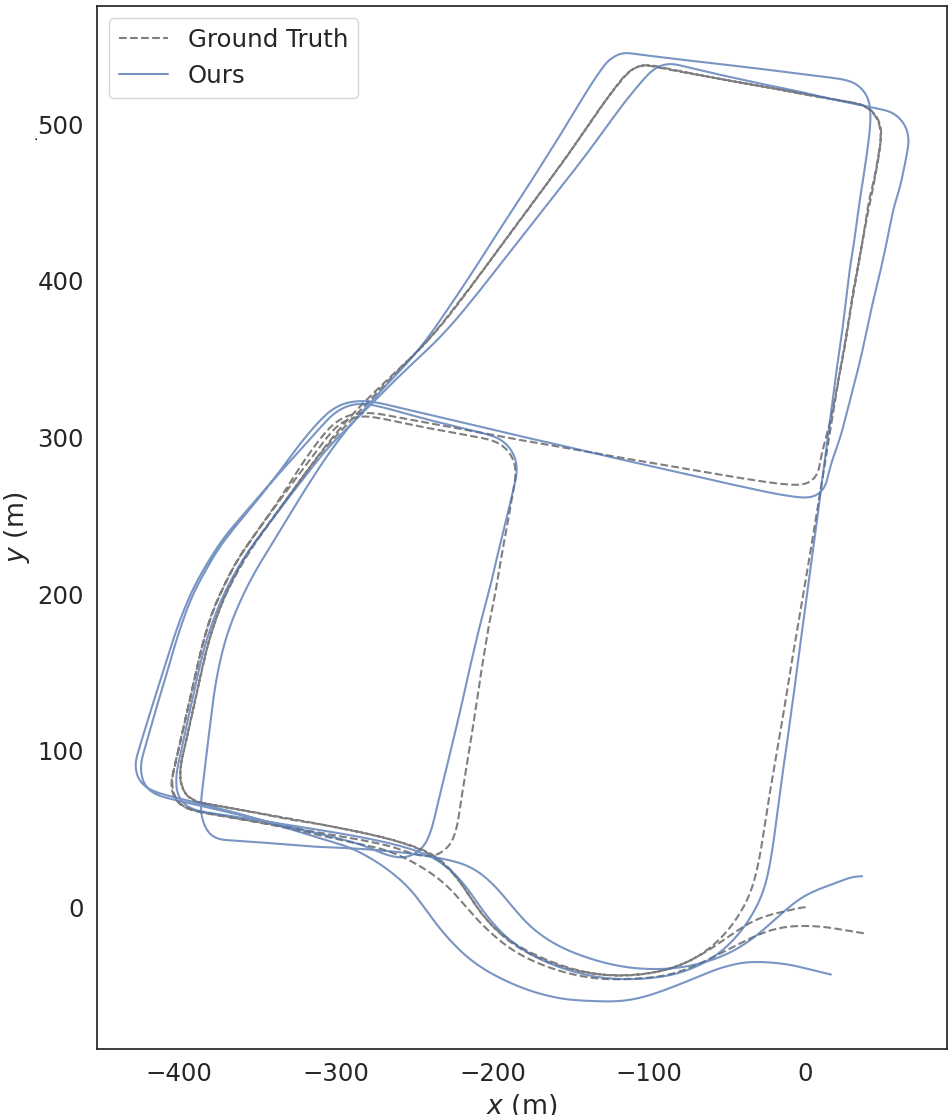} \\ DC1} & 
      \makecell{\includegraphics[width=.176\linewidth]{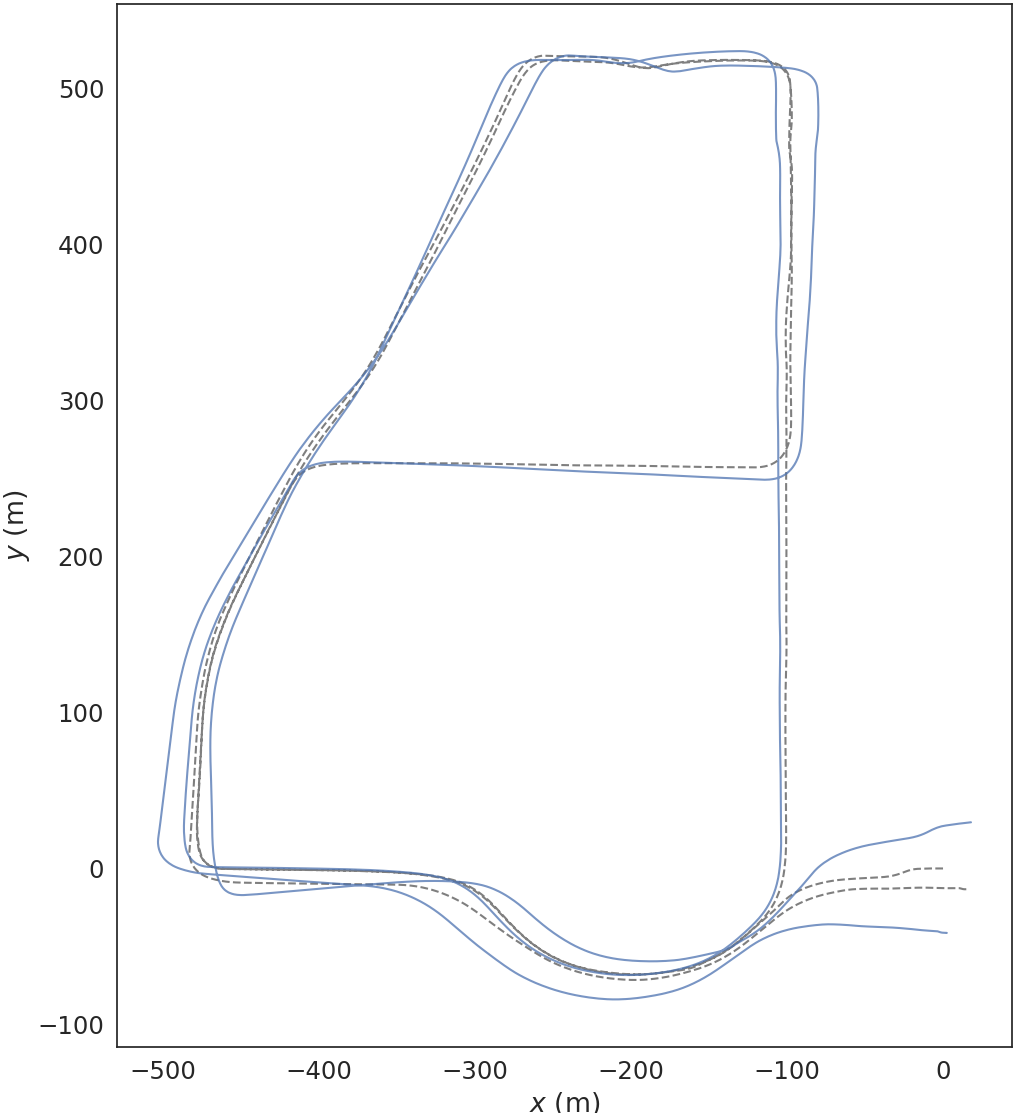} \\ DC2} & 
      \makecell{\includegraphics[width=.176\linewidth]{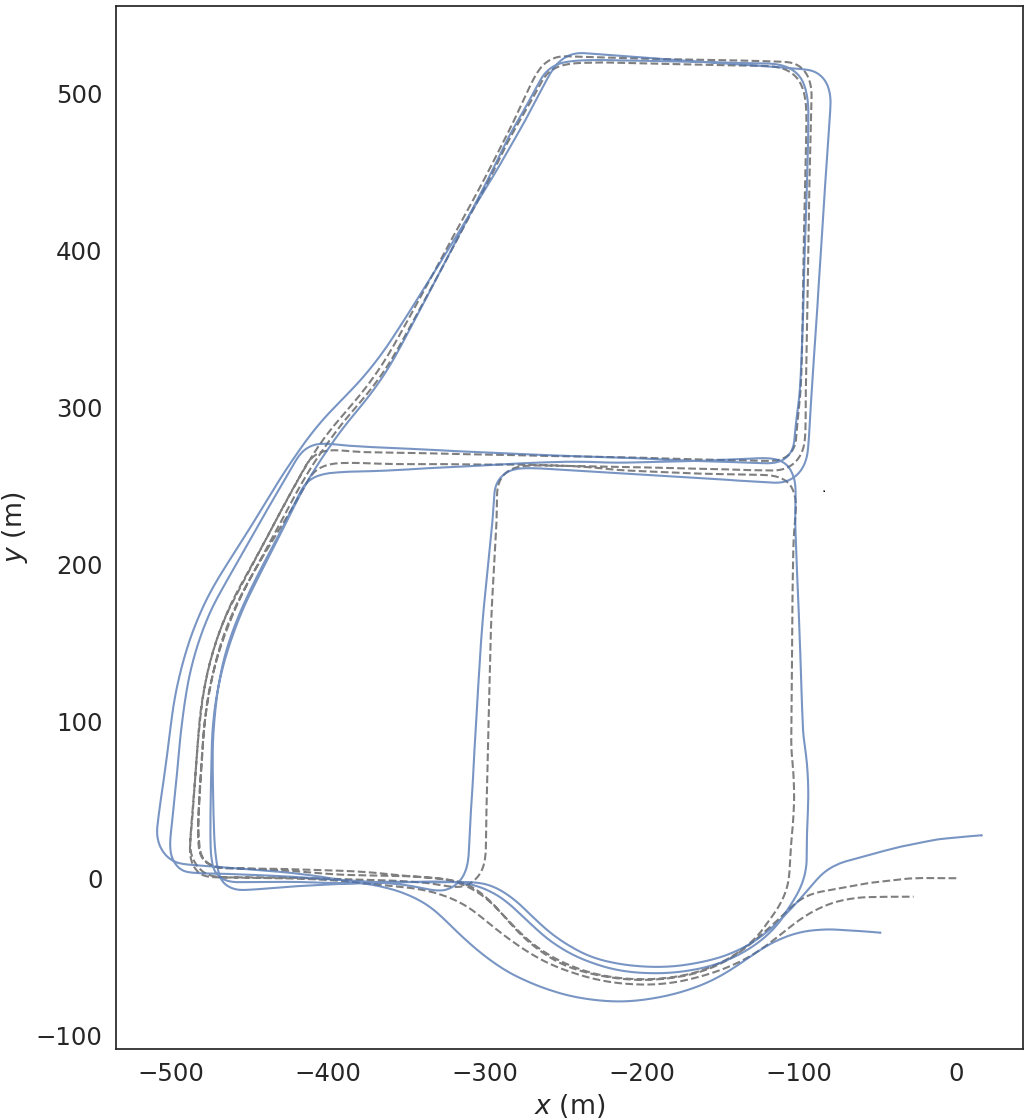} \\ DC3} &

      \multirow{2}{*}[10.6ex]{\makecell{\rotatebox{90}{\includegraphics[width=.353\linewidth]{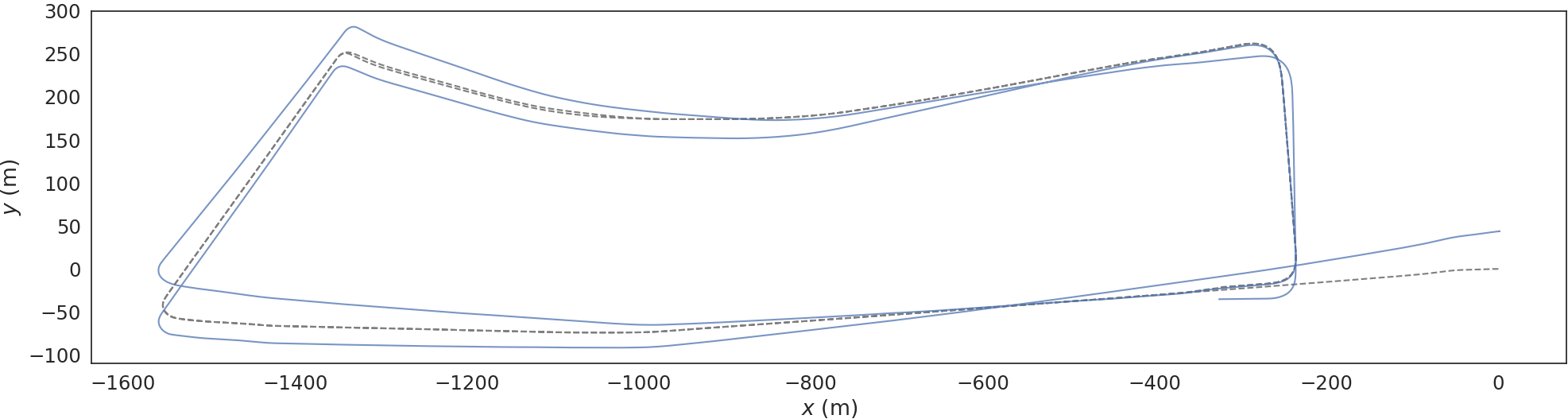}} \\ RS1}}   & 
      \multirow{2}{*}[10.6ex]{\makecell{\rotatebox{90}{\includegraphics[width=.353\linewidth]{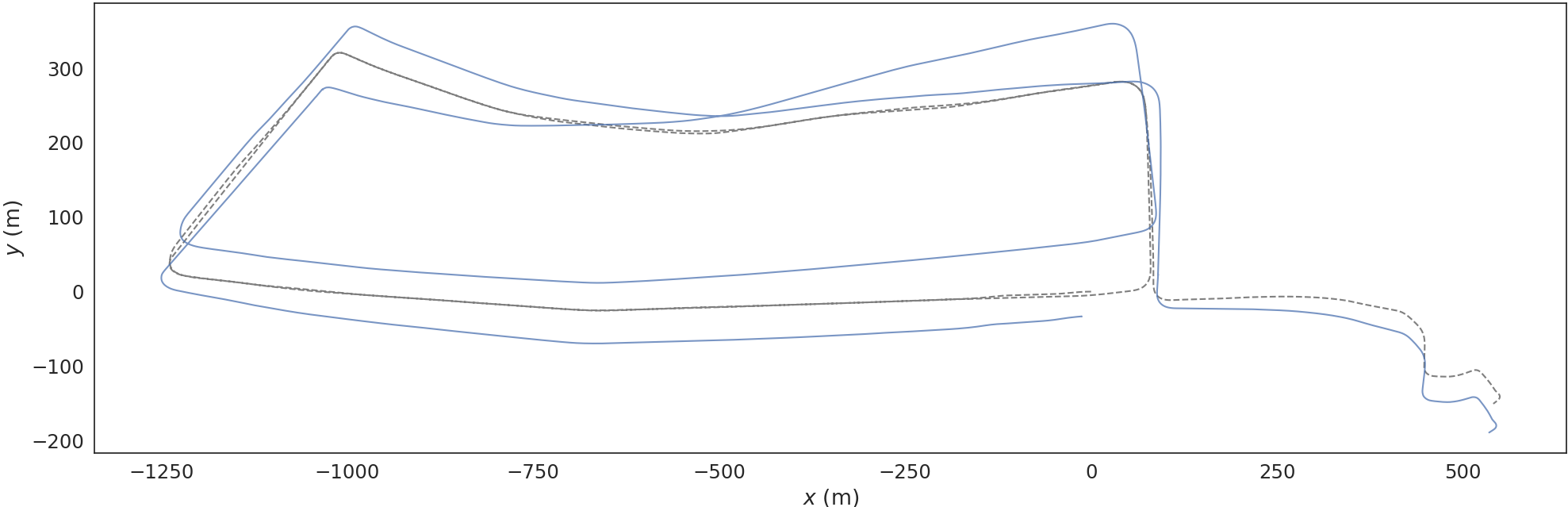}} \\ RS2}}  & 
      \multirow{2}{*}[10.6ex]{\makecell{\raisebox{0pt}{\rotatebox{90}{\includegraphics[width=.353\linewidth]{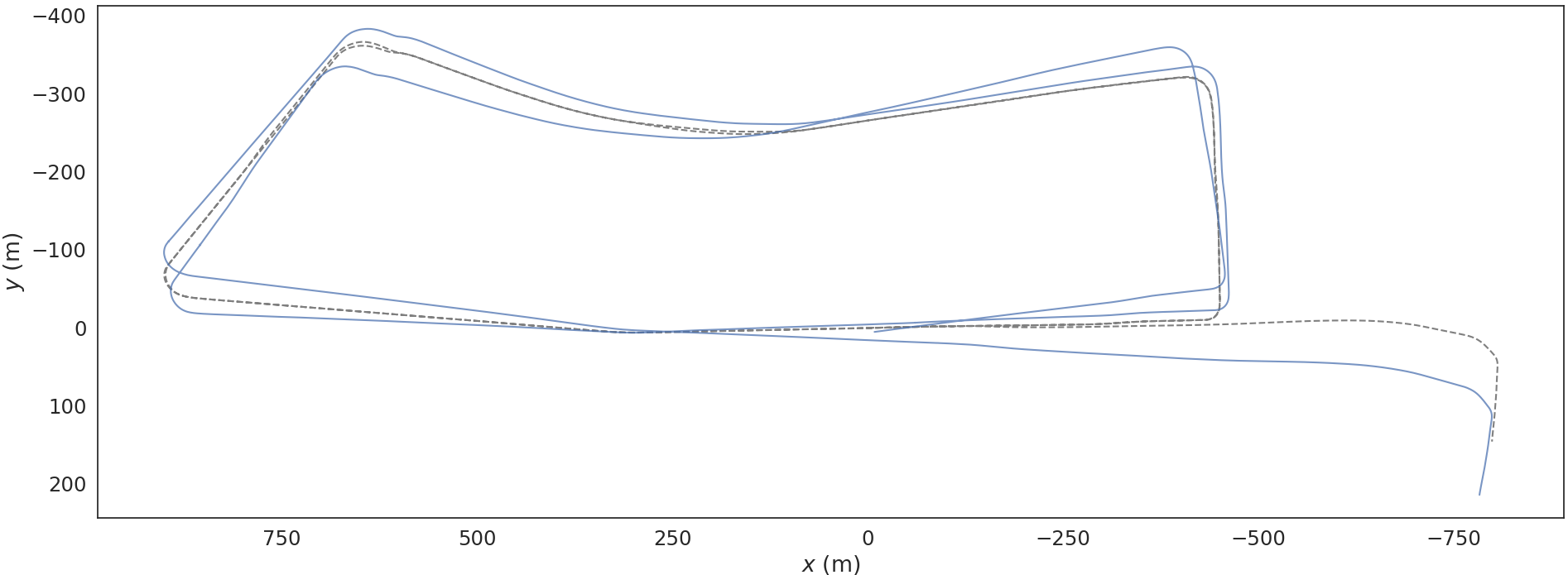}}} \\ RS3}}  \\  
      \makecell{\includegraphics[width=.171\linewidth]{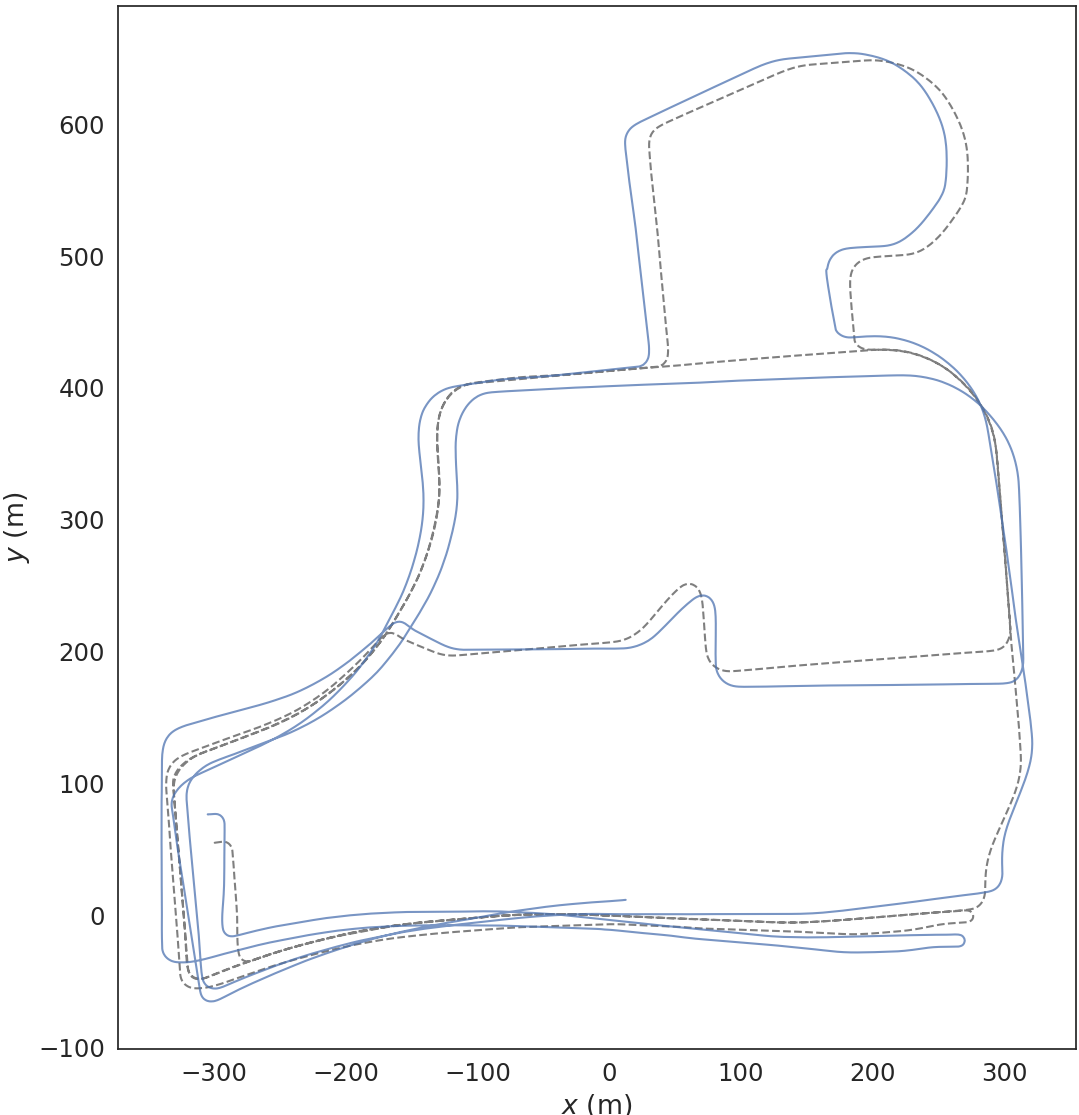} \\ KA1} & 
      \makecell{\includegraphics[width=.176\linewidth]{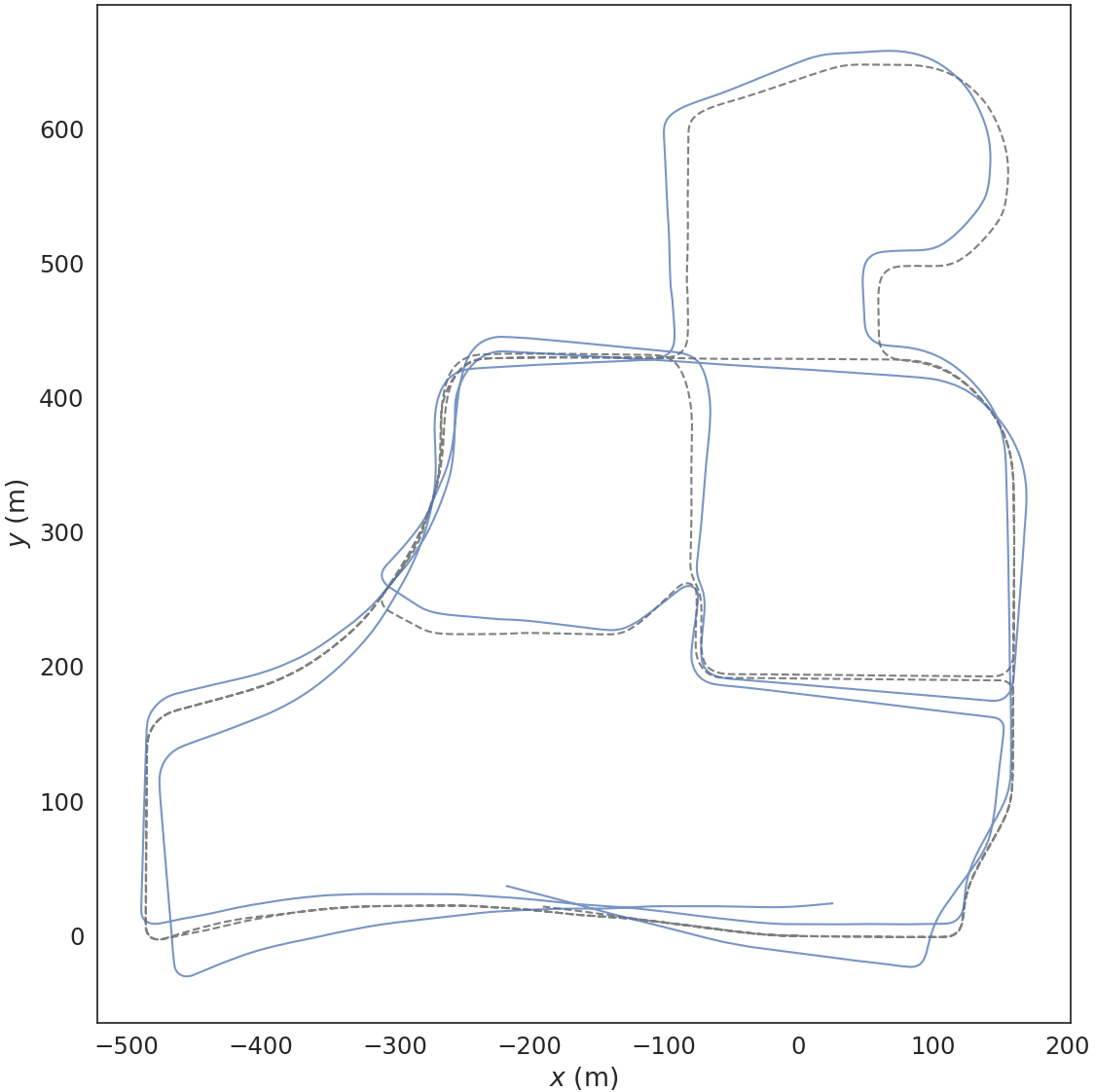} 
      \\ KA2} & 
      \makecell{\includegraphics[width=.176\linewidth]{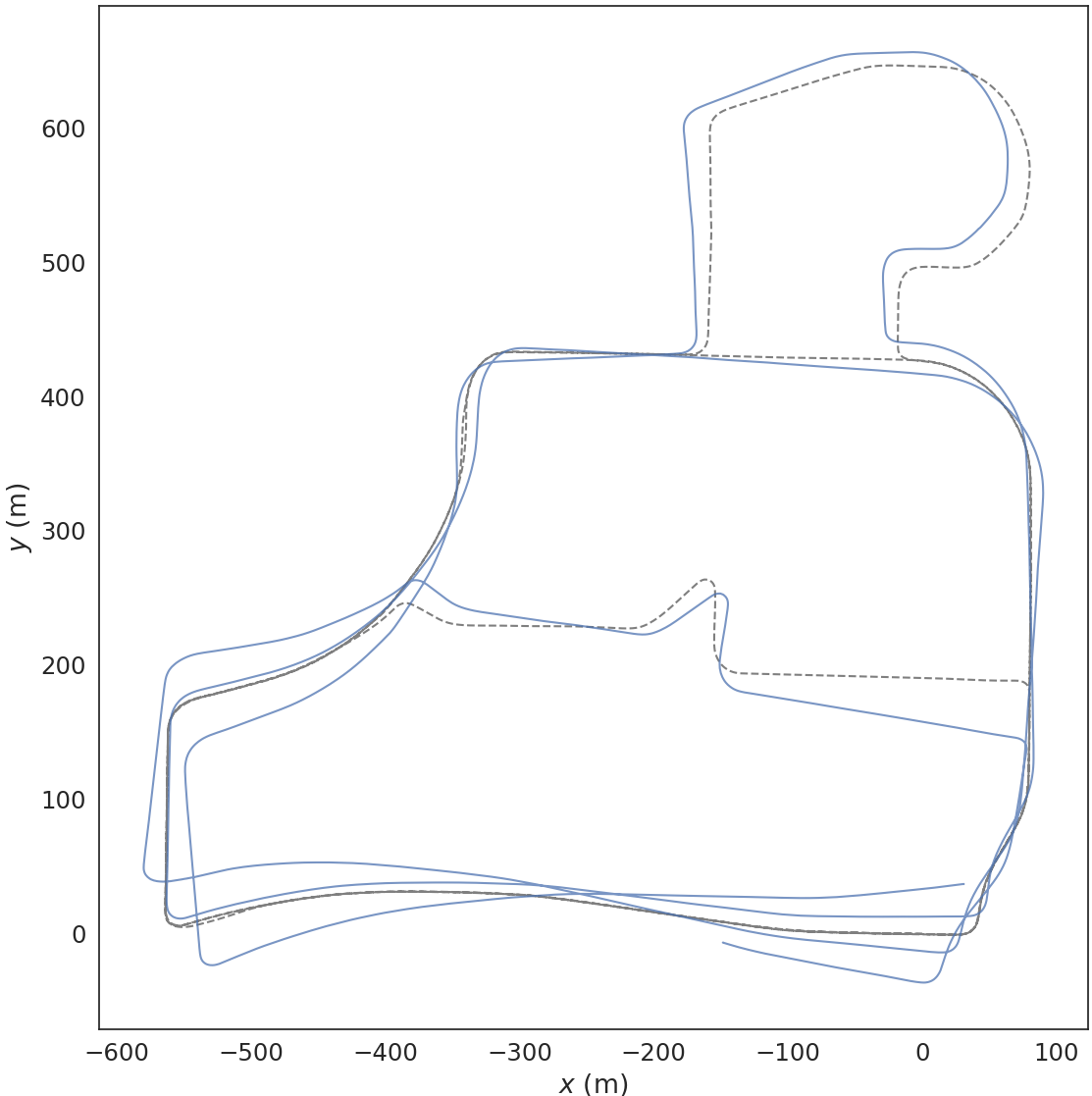}\\ KA3}
      
  \end{tabular} 
  \caption{\textbf{The Qualitative Results of Our Odometry on \textit{MulRAN} Dataset.} The black dashed line represents the ground truth, while the blue solid line represents our trajectory. (\textit{Best viewed with zoom-in.})} 
    \label{fig:mulran_traj} 
\end{scriptsize}
\end{figure*}
To evaluate our localization performance for the long-distance roaming and further demonstrate the adaptability of our method, the experiment is conducted on \textit{MulRAN} dataset that has lots of more challenging driving scenes with a longer trajectory. As shown in Table \ref{tab:ate_mulran}, our method achieves a moderate level of localization accuracy compared to other state-of-the-art LiDAR odometry or SLAM approaches. This is because of the long roaming distance and the existence of loops in the \textit{MulRAN} dataset. For DC sequences and KA sequences, 
\begin{table}[b]
  %
  \centering
  \caption{\textbf{Localization Performance Comparison (ATE RMSE [m]) on \textit{Hilti-21}  LiDAR Dataset.}}
  \label{tab:ate_hilti}
  \resizebox{0.48\textwidth}{!}{
      \begin{tabular}{r|cccccc|c}
        \toprule
          Method  & \text{rpg} & \text{lab}  & \text{base1}  & \text{base4} & \text{cons2} & \text{camp2} & \textbf{Avg.}  \\
          \midrule
          F-LOAM~\cite{wang2021f} & 2.78 & 0.18  & 0.91 & 0.29 & 11.52 & 8.95 & 4.10 \\
          KISS-ICP~\cite{vizzo2023kiss} & 0.22 & 0.07 & 0.32 & \underline{0.11} & 0.84 & 1.98 & 0.58 \\
          HDLGraph-SLAM~\cite{koide2019portable} & 0.35 & \underline{0.05} & \underline{0.28} & 0.37 & 0.74 & \underline{0.35} & 0.36 \\
         SLAMesh~\cite{ruan2023slamesh} & \bf{0.17} & \underline{0.05} & \bf{0.18} & 0.33 & \underline{0.34} & 0.65 & 0.29 \\
        SuMa~\cite{behley2018efficient} & 0.26 & \underline{0.05} & 2.24 & 0.29 & 1.64 & - & 0.90 \\        
          PIN-SLAM~\cite{pan2024pin} & 0.21 & \bf{0.04} & 0.30 & \bf{0.08} & 0.41 & \bf{0.11} & \underline{0.19}\\
          \midrule
          Hi-LOAM & \underline{0.19} & \underline{0.05} & 0.30 & 0.12 & \bf{0.24} & \bf{0.11} & \bf{0.17}
          \\
          \bottomrule
      \end{tabular}
  }
\end{table}
the localization accuracy of our method is not prominent as SuMa \cite{behley2018efficient} and MULLS \cite{pan2021mulls}, they  incorporate function of the loop-closure detection, which eliminates accumulated drifts and thereby improves accuracy. Trajectory loops are shown in different sequences of Fig. \ref{fig:mulran_traj}, which is the qualitative results of our odometry visualization on \textit{MulRAN} dataset. However, compared to similar type of learning-based implicit odometry method like PIN-SLAM \cite{pan2024pin}, we achieve more accurate results. And the performance of ICP-based method like KISS-ICP\cite{vizzo2023kiss} is not robust and consistent across different scenarios. To be noticed, although our method lacks loop closure, we still produce better results than the loop-closure-based SLAM method called SuMa \cite{behley2018efficient} for RS sequences.

\begin{table}[b!]
  %
  \centering
  \caption{\textbf{Localization Performance Comparison (ATE RMSE [m]) on \textit{Newer College} LiDAR Dataset.} \xmark \hspace{0.03cm} denotes the execution failure.}

  \resizebox{0.48\textwidth}{!}{
      \begin{tabular}{r|ccccccc|c}
          \toprule
          Method   & \text{01}     & \text{02}      & \text{quad\_e}         &  \text{math\_e}    & \text{ug\_e}  &  \text{cloister}     & \text{stairs}  & \textbf{Avg.} \\
        
          \midrule
          F-LOAM~\cite{wang2021f} & 6.74 & \xmark & 0.40 &  0.26 &  \underline{0.09} & 7.69 & \xmark & 3.04  \\
          KISS-ICP~\cite{vizzo2023kiss} & \bf{0.62} & 1.88 & \underline{0.10} & \bf{0.07} & 0.33 & 0.30 & \xmark & \bf{0.55} \\ 
          \midrule
          SuMa~\cite{behley2018efficient} & 2.03 & 3.65 & 0.28 & 0.16 & \underline{0.09} & 0.20 & 1.85 & 1.18 \\
          MULLS~\cite{pan2021mulls} &  2.51 & 8.39 & 0.12 & 0.35 & 0.86 & 0.41 & \xmark & 2.11 \\ 
          MD-SLAM~\cite{di2022md} &  - & \underline{1.74} & 0.25 & - & - & 0.36 & 0.34 &  -\\
          SC-LeGO-LOAM~\cite{shan2018lego} & - & \bf{1.30} & \bf{0.09} & - & - & 0.20 & 3.20 & - \\ 
          
          PIN-LO~\cite{pan2024pin}& 2.08 & 5.32 & \bf{0.09} & \underline{0.09} & \bf{0.07} &  \underline{0.19} &  \bf{0.07} & 1.13 \\

          \midrule
          Hi-LOAM &  \underline{1.61} & 5.41& \bf{0.09} & 0.11 & \bf{0.07} & \bf{0.16}  & \underline{0.08} & \underline{1.08} \\
          \bottomrule
      \end{tabular}
      }
    \label{tab:ate_ncd}
  \vspace{-6pt}
\end{table}

\subsubsection{\textbf{Evaluation on Datasets from Hand-held Devices}} To evaluate the effectiveness and robustness of our method, we conducted further experiments on the \textit{Newer College} \cite{ramezani2020newer}, \textit{Hilti-21}  \cite{2109.11316}, and \textit{Hilti-23} \cite{nair2024hiltislamchallenge2023} datasets.
\begin{table}[b!]
\centering
\caption{\textbf{Localization Performance Comparison (ATE RMSE [cm]) on the Sequence 01 of \textit{MaiCity} Dataset.} \xmark \hspace{0.03cm} denotes the execution failure.}
\renewcommand{\arraystretch}{0.9} 
\setlength{\tabcolsep}{16pt} 
\begin{tabular}{rcc}
\toprule
Approach     & Sensor              & \textit{MaiCity 01}  \\ \midrule
SuMa\cite{behley2018efficient}     & \multirow{8}{*}{Velodyne HDL-64} & 6.1  \\
F-LOAM \cite{wang2021f}    &                      & 6.2  \\
KISS-ICP \cite{vizzo2023kiss} &                      & \xmark \\
Puma \cite{vizzo2021poisson}    &                      & 5.8  \\
SLAMesh \cite{ruan2023slamesh} &                      & 1.7  \\ 
Mesh-LOAM \cite{zhu2024mesh} &                      & 1.6  \\
NeRF-LOAM \cite{deng2023nerf} &                      & 13 \\
PIN-SLAM \cite{pan2024pin}&                      & \underline{0.9} \\
\midrule
Hi-LOAM         &                      & \bf{0.7} \\ 
\bottomrule
\end{tabular}
 \label{tab:ate_mai}
\end{table}
These two datasets are collected using hand-held LiDAR devices, therefore their motion variations are likely to be more pronounced, and there are more vibrations on the vertical direction. As shown in Table \ref{tab:ate_hilti}, 
\begin{table*}[b!]
  \caption{\textbf{The 3D Reconstruction Accuracy of Different Methods on \textit{Newer College} and \textit{Mai City} Datasets.} }
  \setlength{\belowcaptionskip}{-6pt}
  \centering
     \resizebox{0.94\textwidth}{!}{
     \begin{tabular}{lc|cccc|cccc}
        \toprule 
        \multirow{2}{*}{Method}& \multirow{2}{*}{Pose} & \multicolumn{4}{c|}{\textit{Quad} from \textit{Newer College}}& \multicolumn{4}{c}{\textit{Mai City}}\\
        &&Map. Acc. $\downarrow$&Map. Comp. $\downarrow$&C-l1. $\downarrow$& F-score (20cm)  $\uparrow $&Map. Acc. $\downarrow$&Map. Comp. $\downarrow$&C-l1. $\downarrow$& F-score (10cm) $\uparrow $\\
        \toprule
        VDB-Fusion~\cite{vizzo2022vdbfusion}& \multirow{2}{*}{KISS-ICP~\cite{vizzo2023kiss}} & 14.03 &25.46&19.75& 69.50 & 15.21 & 28.66 & 21.94 & 63.35\\
        SHINE~\cite{zhong2023shine}& & 14.87
        & 20.02 & 17.45 & 68.85 & 14.46 & 34.03 & 24.24 & 64.38\\
        \midrule     Puma~\cite{vizzo2021poisson}&\multirow{5}{*}{Own Odometry} &  15.30&71.91& 43.60& 57.27 & 7.89 & 9.14 & 8.51 & 68.04 \\
        SLAMesh~\cite{ruan2023slamesh} & & 19.21 & 48.83 & 34.02 & 45.24 & \underline{5.67} & 12.99 & 9.33 & 75.10 \\
        NeRF-LOAM~\cite{deng2023nerf} & & 12.89 & 22.21 & 17.55 & 74.37 & 5.69 & 11.23 & 8.46 & 77.23\\
        PIN-SLAM~\cite{pan2024pin} && \underline{11.55} & \underline{15.25} & \underline{13.40} & \underline{82.08} & 5.83 & \underline{4.98} & \underline{5.40} & \underline{88.40} \\
        Hi-LOAM && \bf{11.37} & \bf{14.82} & \bf{13.09} & \bf{88.32} & \bf{4.48} & \bf{4.15} & \bf{4.32} & \bf{92.76} \\
        \bottomrule
     \end{tabular}
     }
  \label{tab:recon_experiments}
\end{table*}
\begin{figure*}[b!]
    \centering
\captionsetup{position=top}
\captionsetup[subfigure]{labelformat=empty}
\subfloat[\scriptsize SLAMesh \cite{ruan2023slamesh} ]{\includegraphics[width=.24\linewidth]
{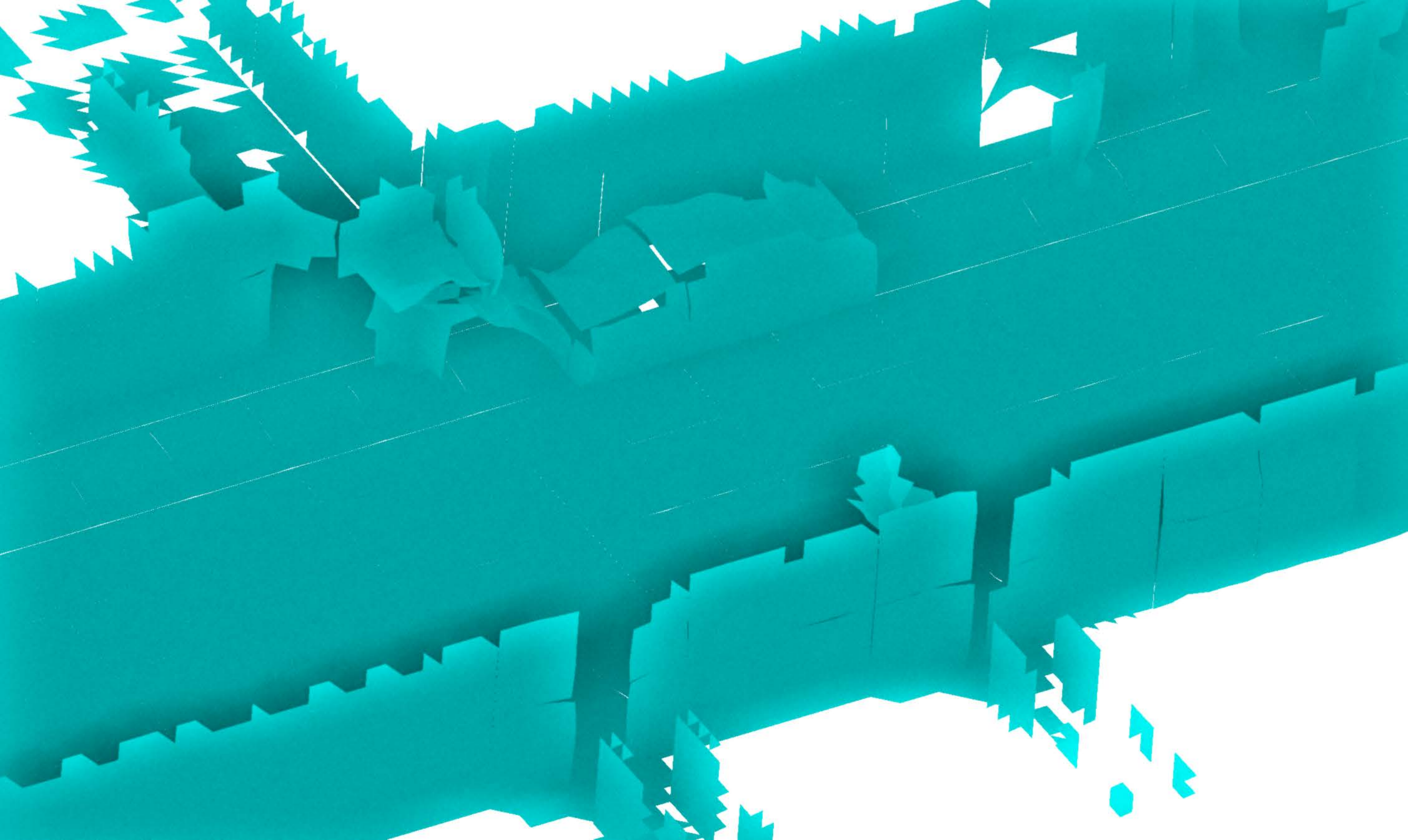}}
\hspace{0.05em} 
\vspace{-0.15em}
\subfloat[\scriptsize NeRF-LOAM \cite{deng2023nerf}  ]{\includegraphics[width=.24\linewidth]
{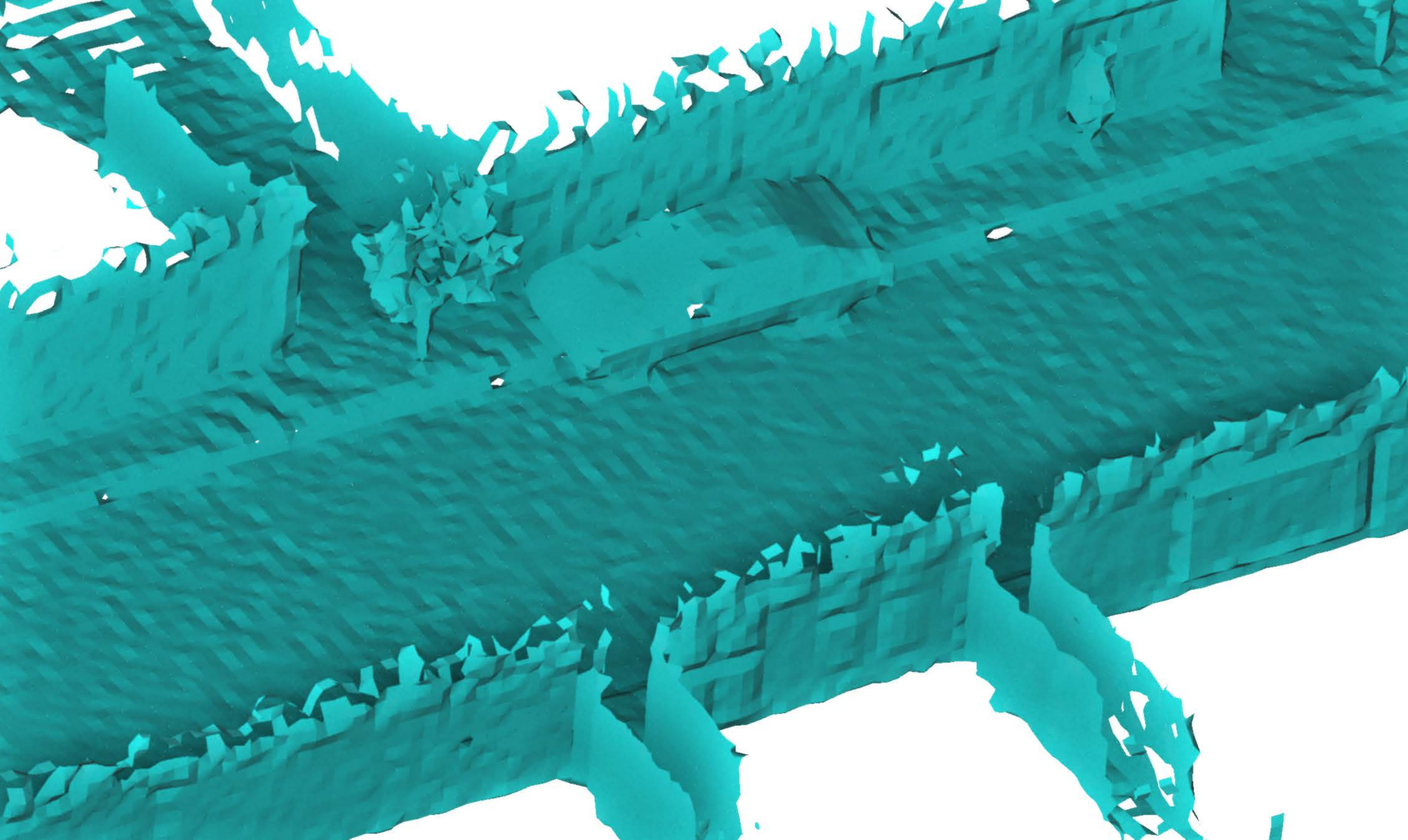}}
\hspace{0.05em} 
\vspace{-0.15em}
\subfloat[\scriptsize PIN-SLAM \cite{pan2024pin} ]{\includegraphics[width=.24\linewidth]{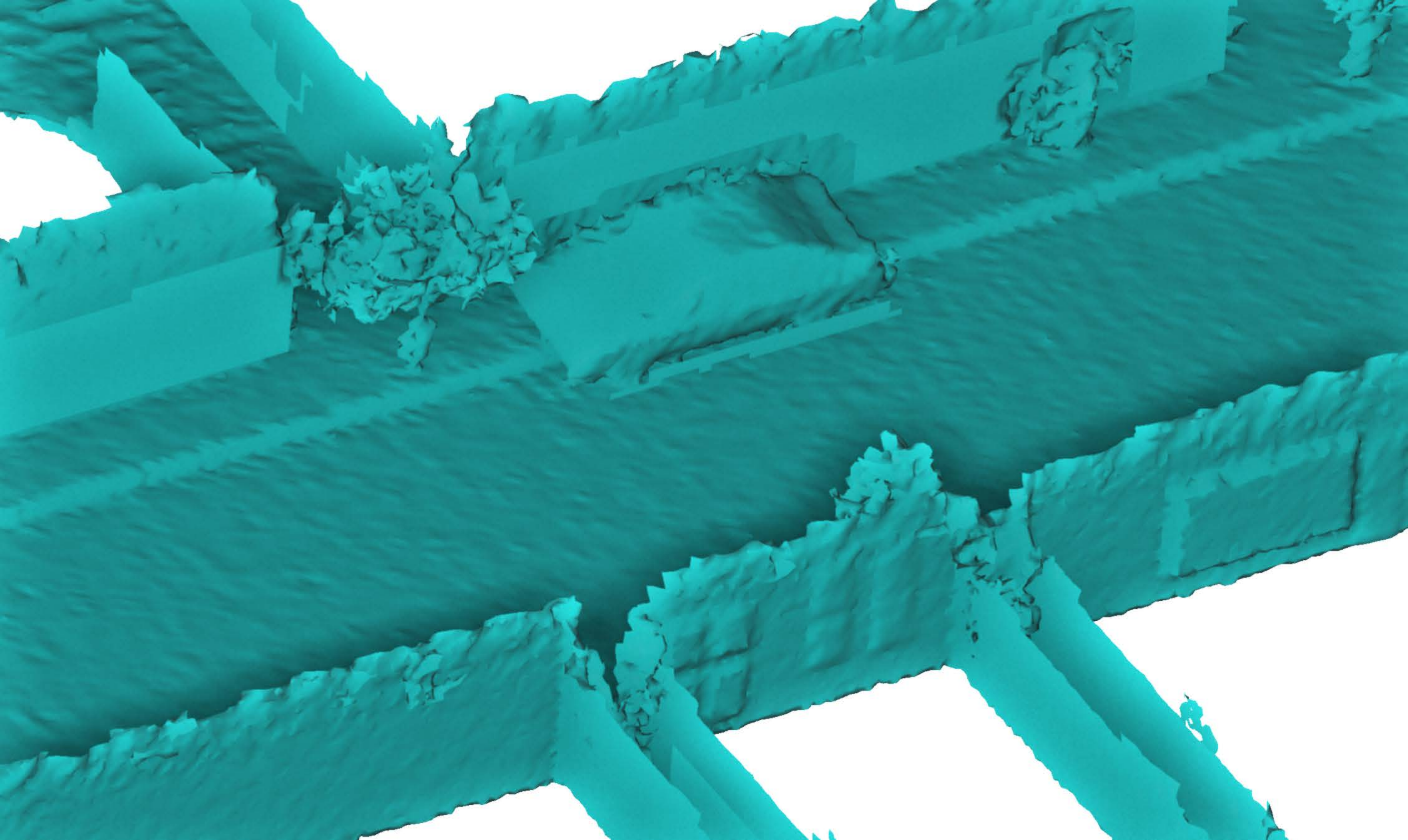}}
\hspace{0.05em} 
\vspace{-0.15em}
\subfloat[\scriptsize Hi-LOAM]{\includegraphics[width=.24\linewidth]{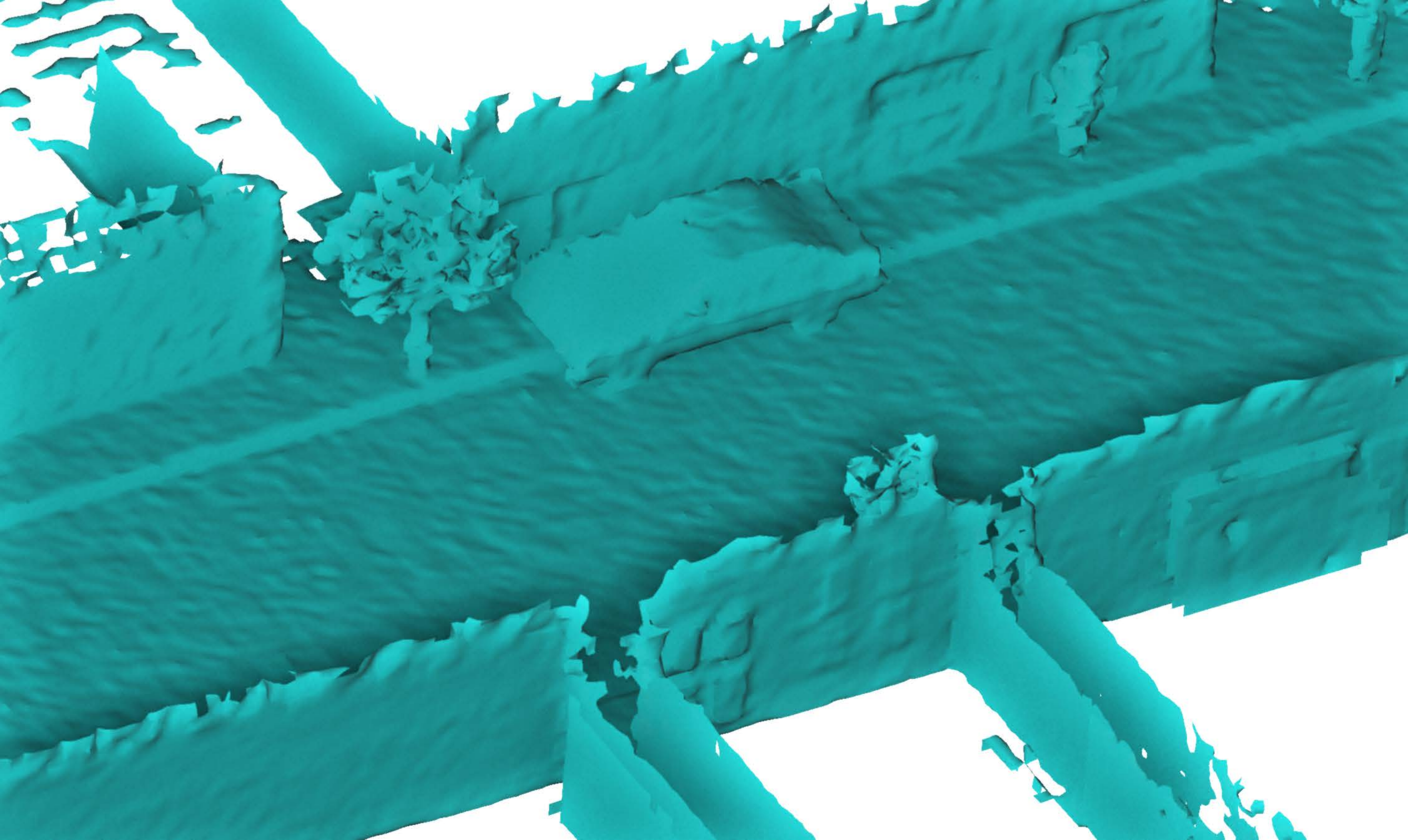}}
\vspace{-0.15em}

\subfloat{\includegraphics[width=.24\linewidth]{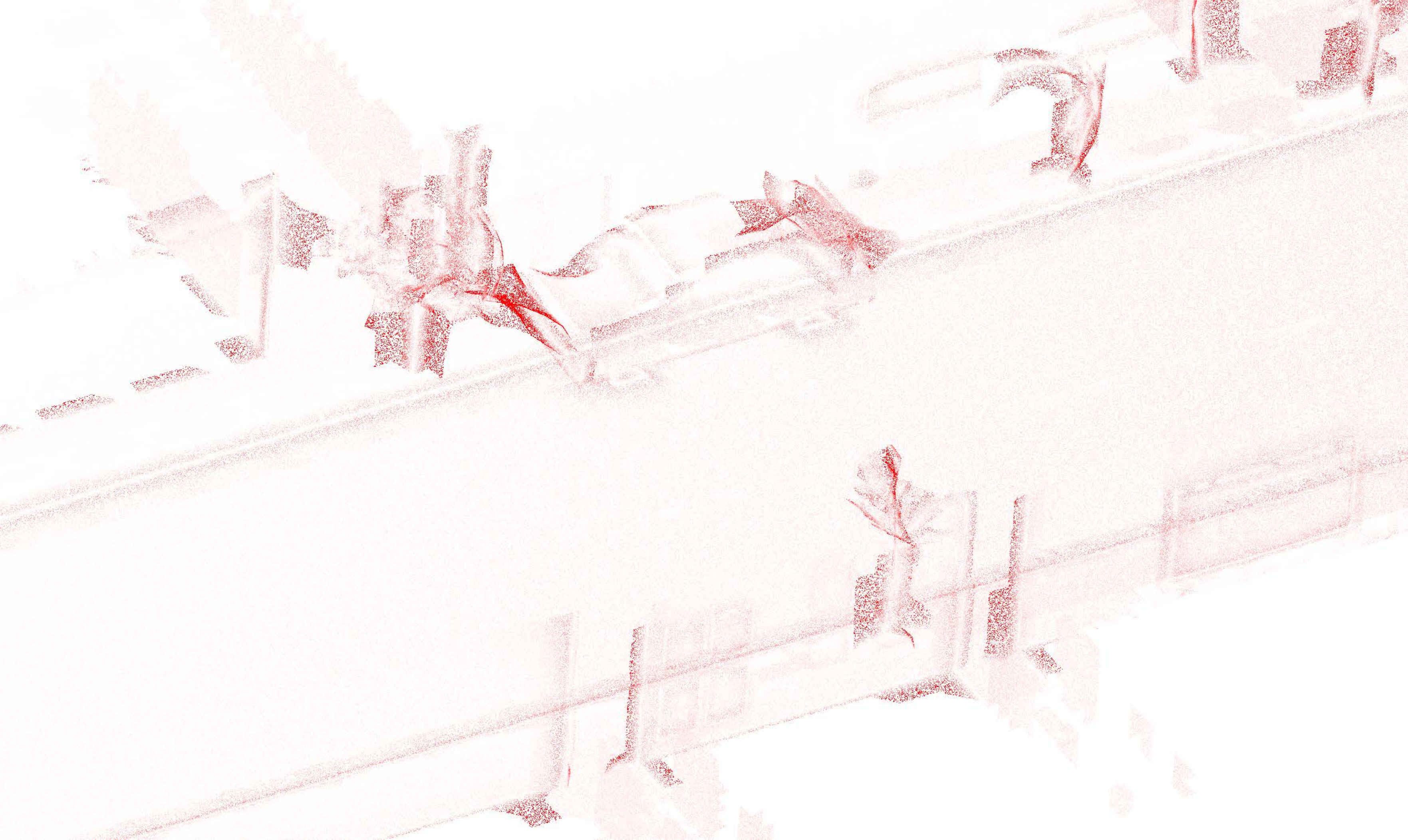}}
\hspace{0.05em} 
\vspace{-0.3em}
\subfloat{\includegraphics[width=.24\linewidth]{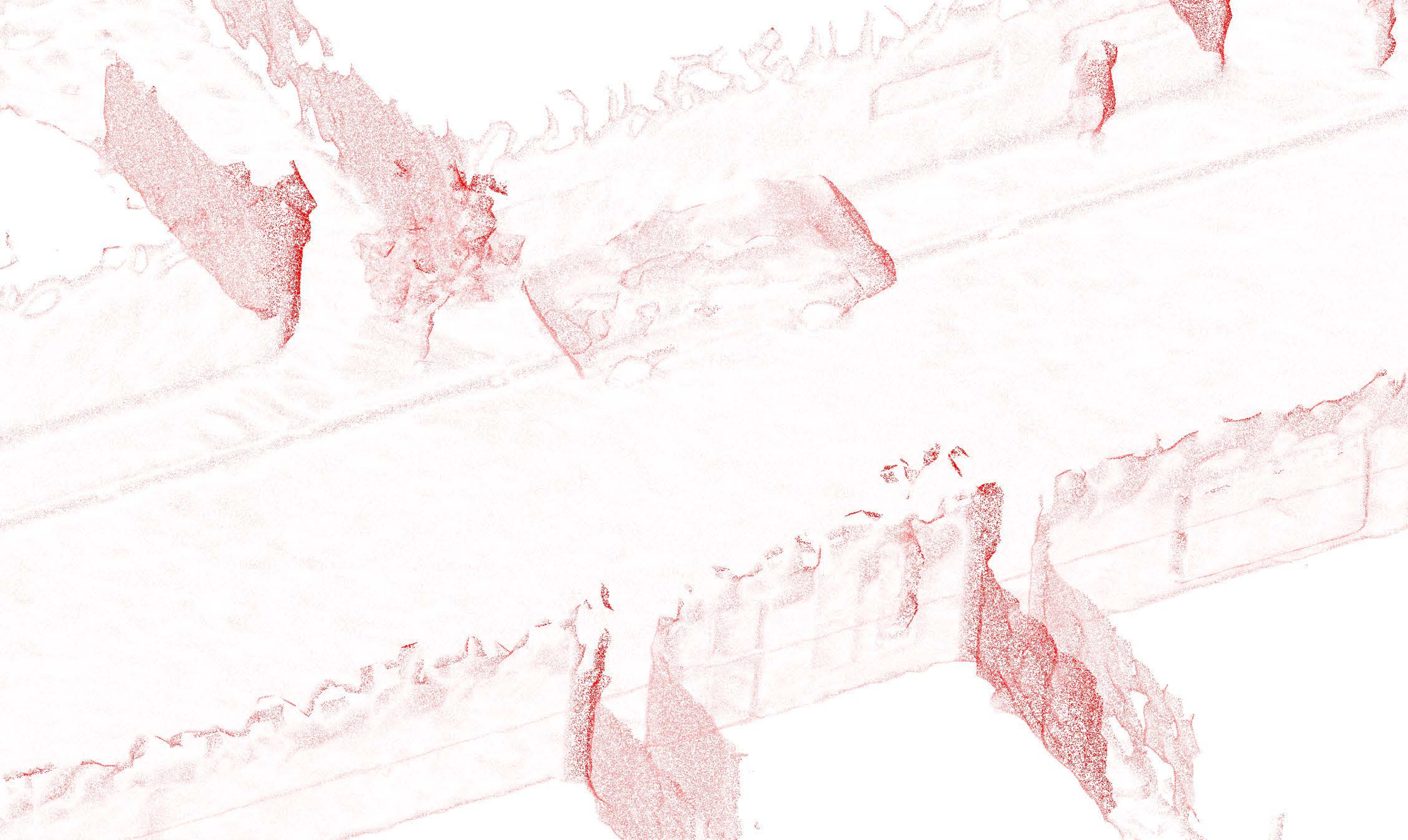}} 
\hspace{0.05em} 
\vspace{-0.3em}
\subfloat{\includegraphics[width=.24\linewidth]{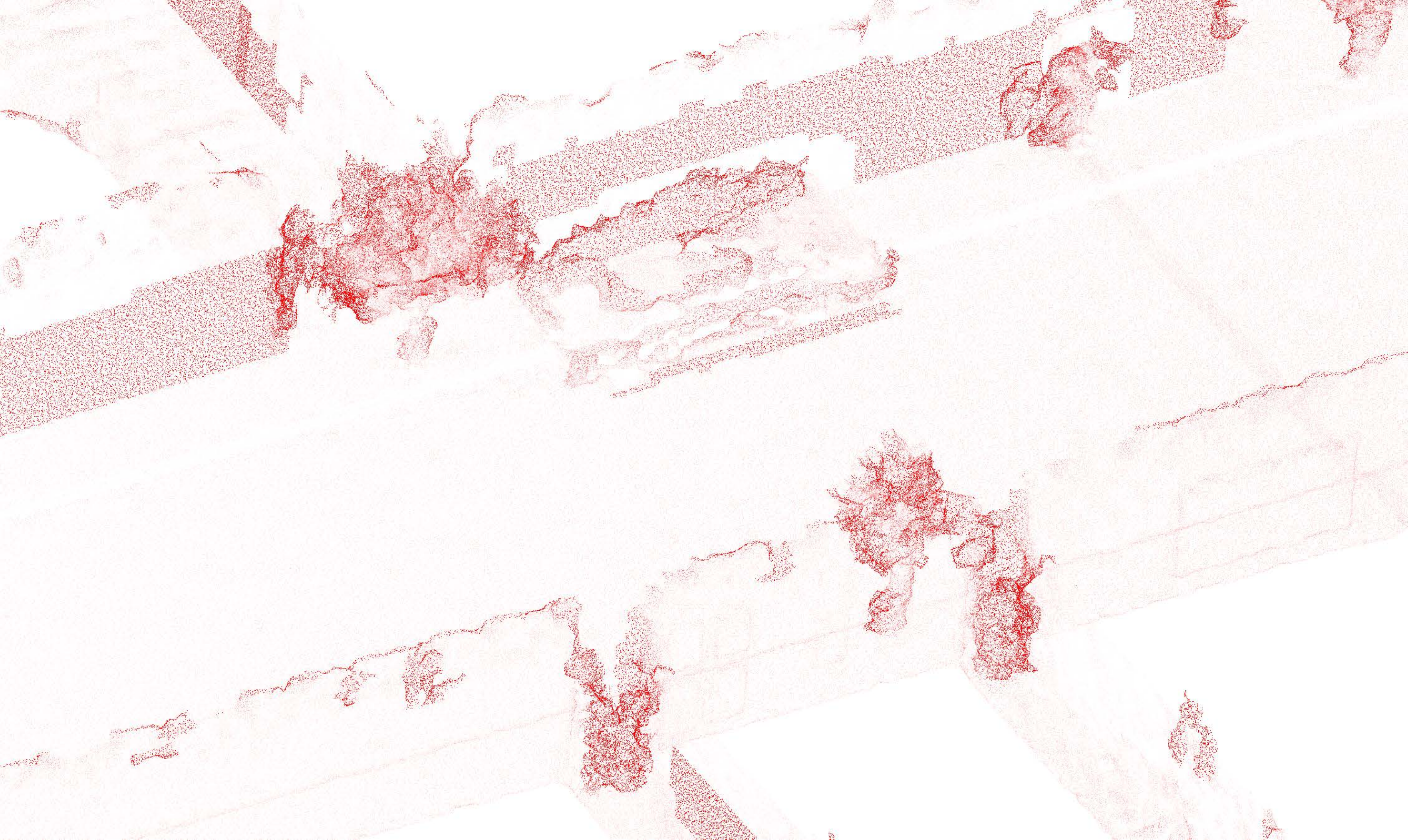}} 
\hspace{0.05em} 
\vspace{-0.3em}
\subfloat{\includegraphics[width=.24\linewidth]{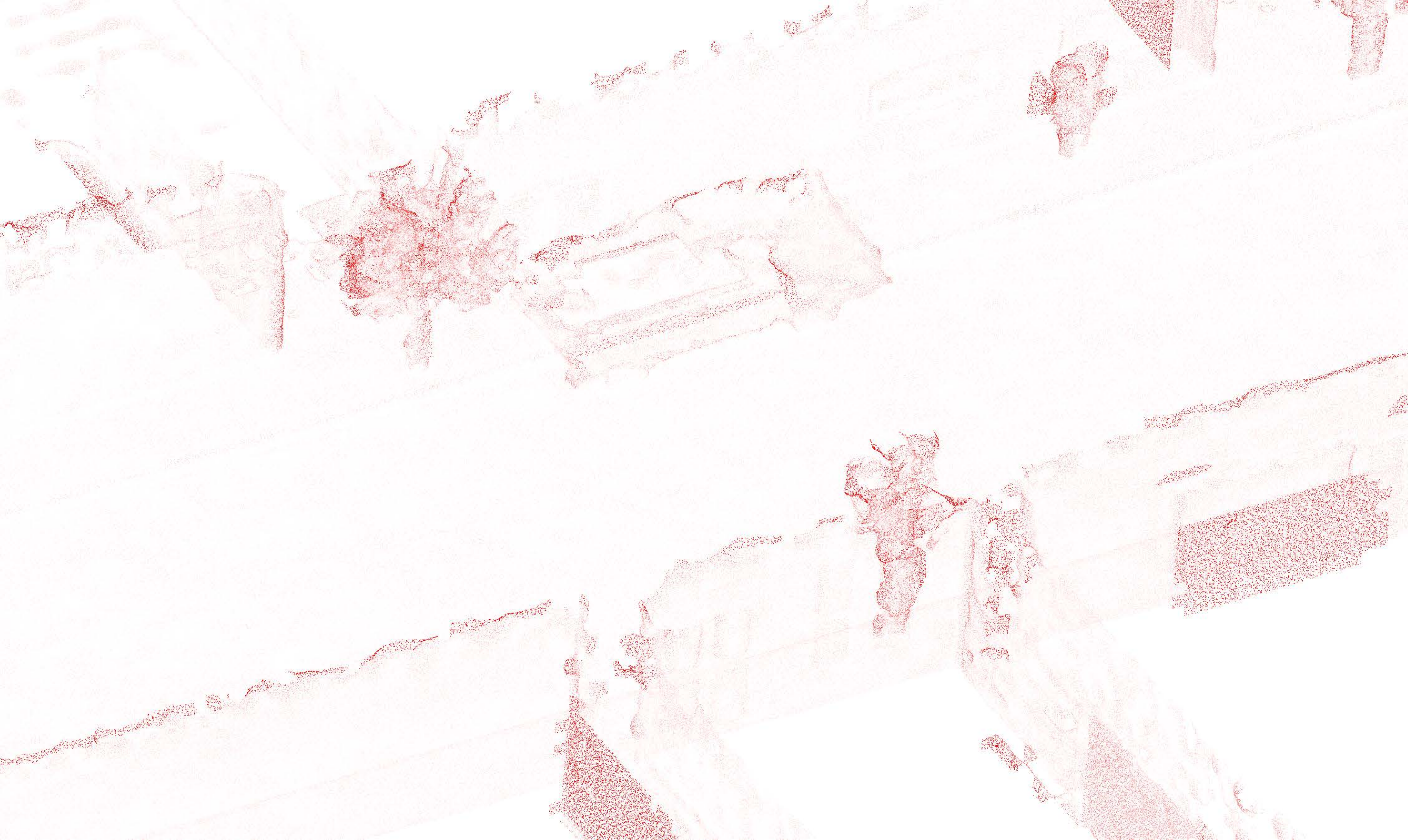}}
\caption{\textbf{The Visualization of Our 3D Reconstruction Results on \textit{Mai City} Dataset via Comparing with Other Related Methods.} The mapping results in first row are original reconstruction result, and the second row presents the error maps with ground truth mesh as a reference, where the red points stand for large error above 25cm. (\textit{Best viewed with zoom in.})}
\label{fig:rec_mai}
\end{figure*}
\begin{figure*}[b!]
    \centering
\captionsetup{position=top}
\captionsetup[subfigure]{labelformat=empty}
\subfloat[\scriptsize SLAMesh \cite{ruan2023slamesh} ]{\includegraphics[width=.24\linewidth]{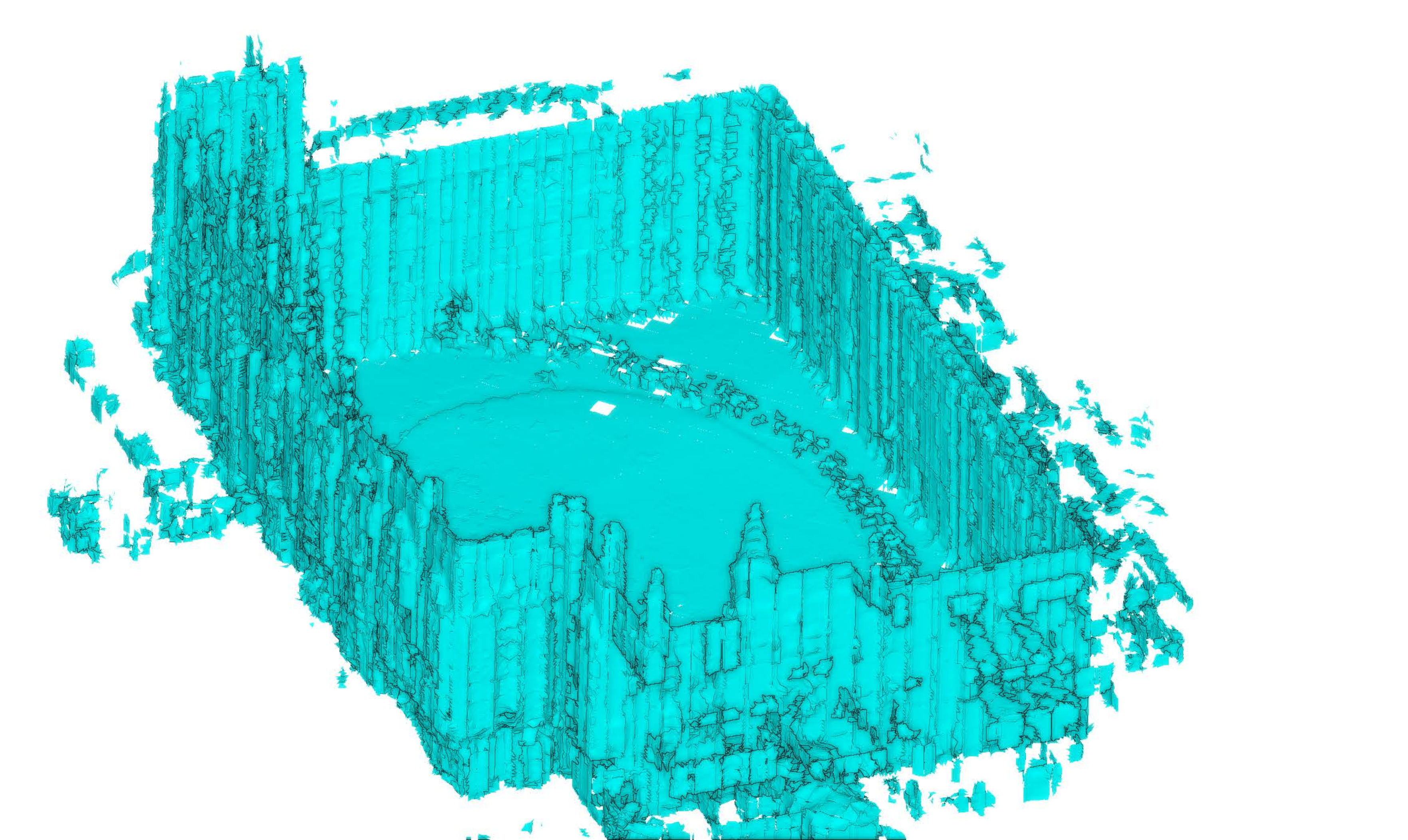}}
\hspace{0.03em} 
\vspace{-0.15em}
\subfloat[\scriptsize NeRF-LOAM \cite{deng2023nerf}  ]{\includegraphics[width=.24\linewidth]{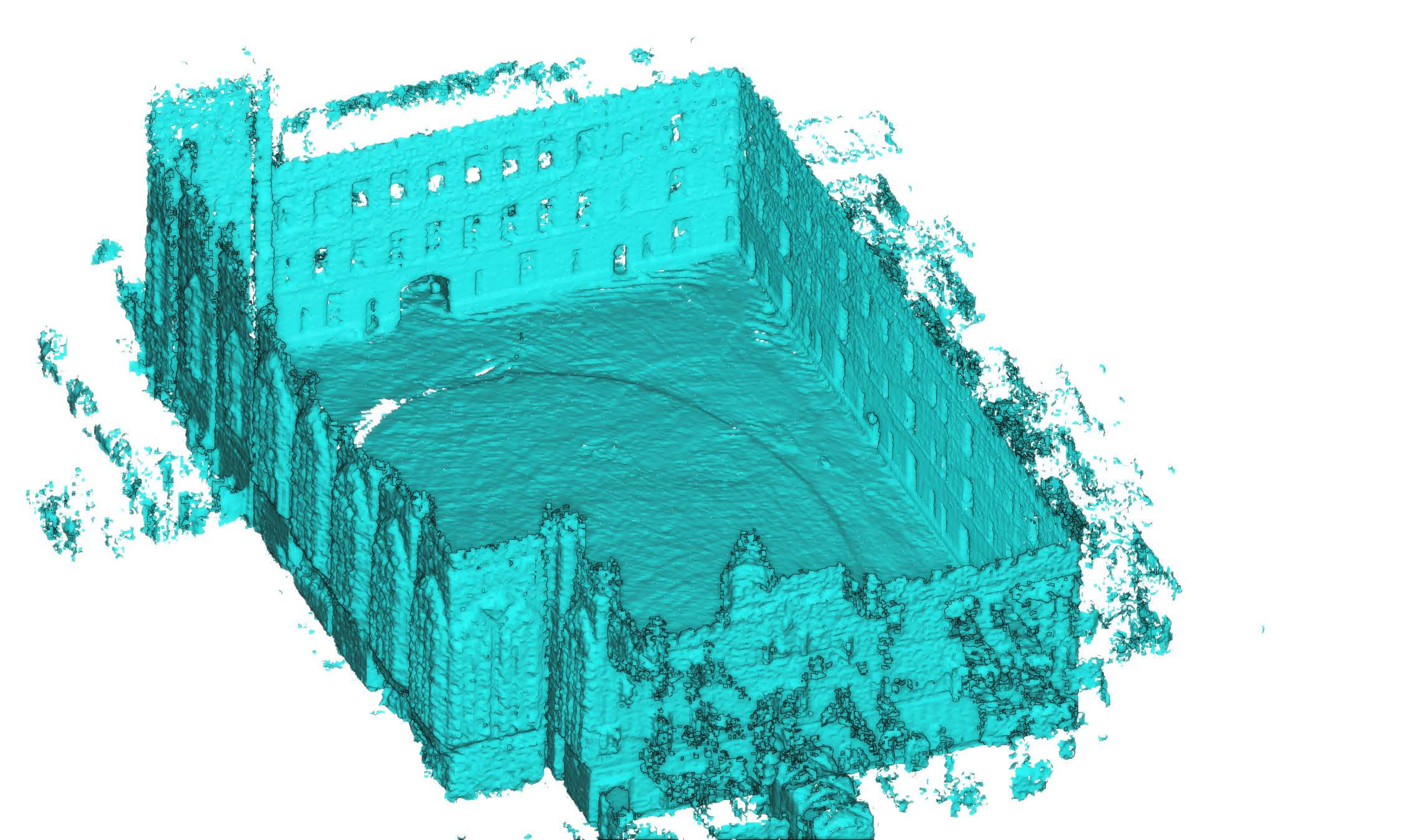}}
\hspace{0.03em} 
\vspace{-0.15em}
\subfloat[\scriptsize PIN-SLAM \cite{pan2024pin} ]{\includegraphics[width=.24\linewidth]{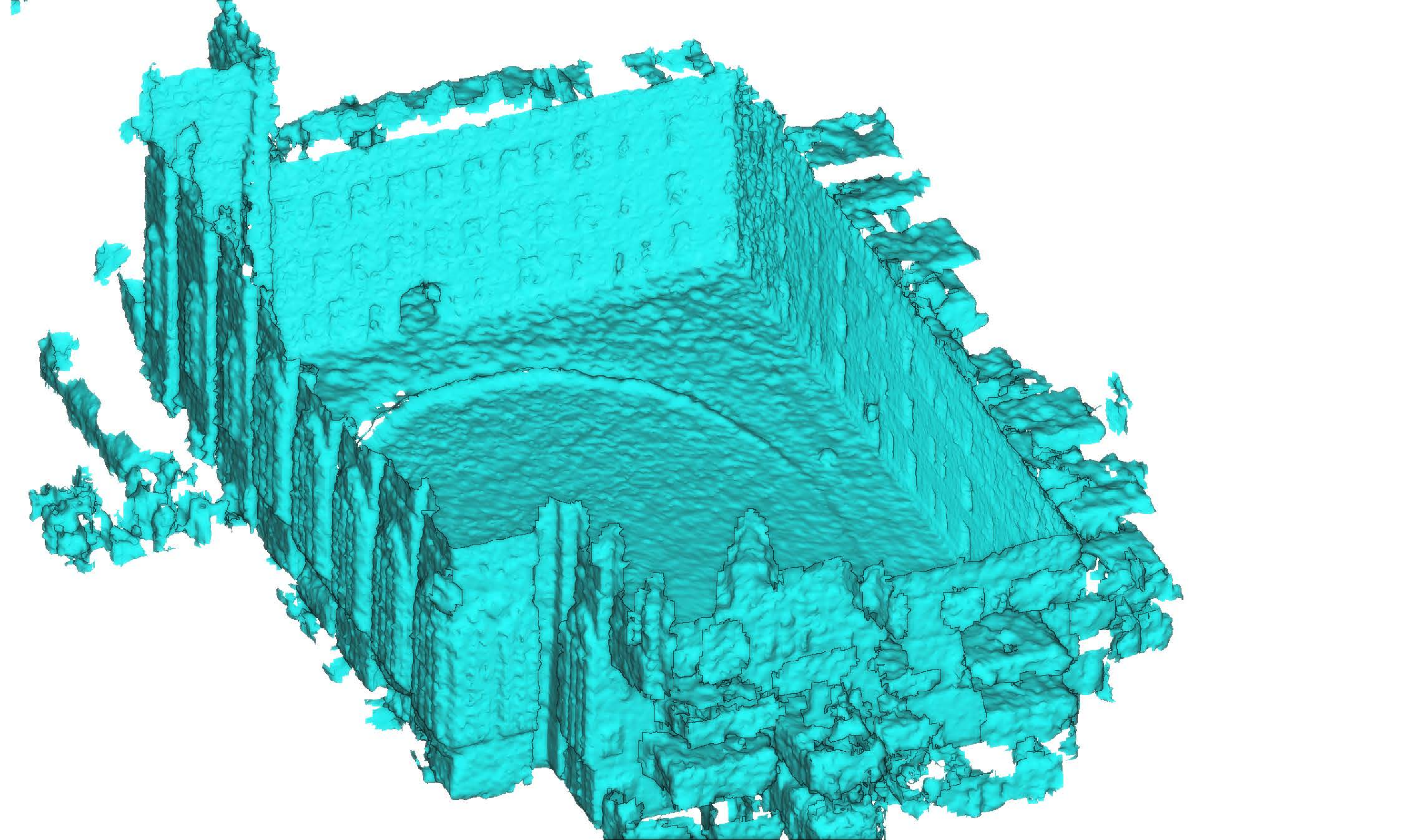}}
\hspace{0.03em} 
\vspace{-0.15em}
\subfloat[\scriptsize Hi-LOAM]{\includegraphics[width=.24\linewidth]{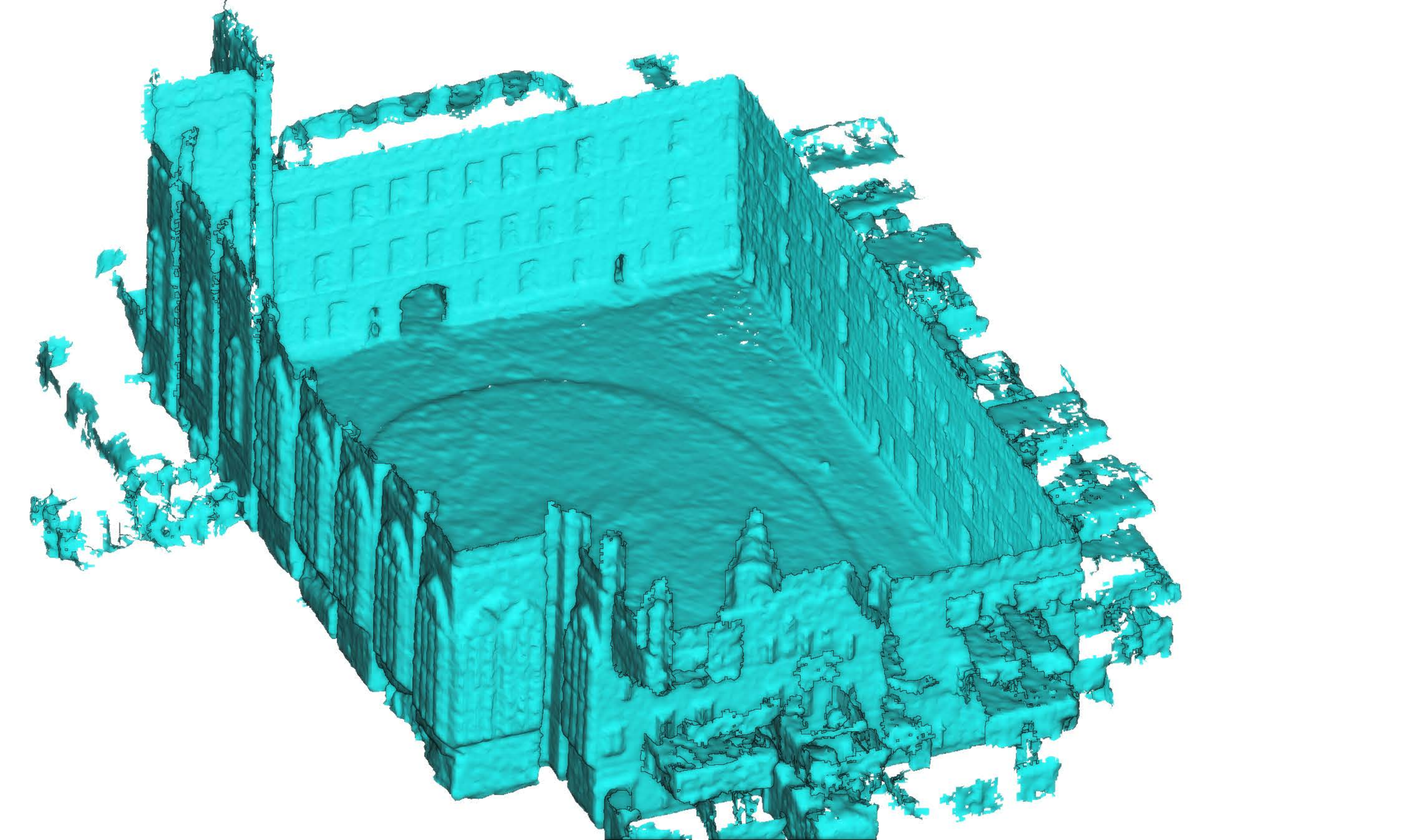}}
\vspace{-0.15em}
\subfloat{\includegraphics[width=.24\linewidth]{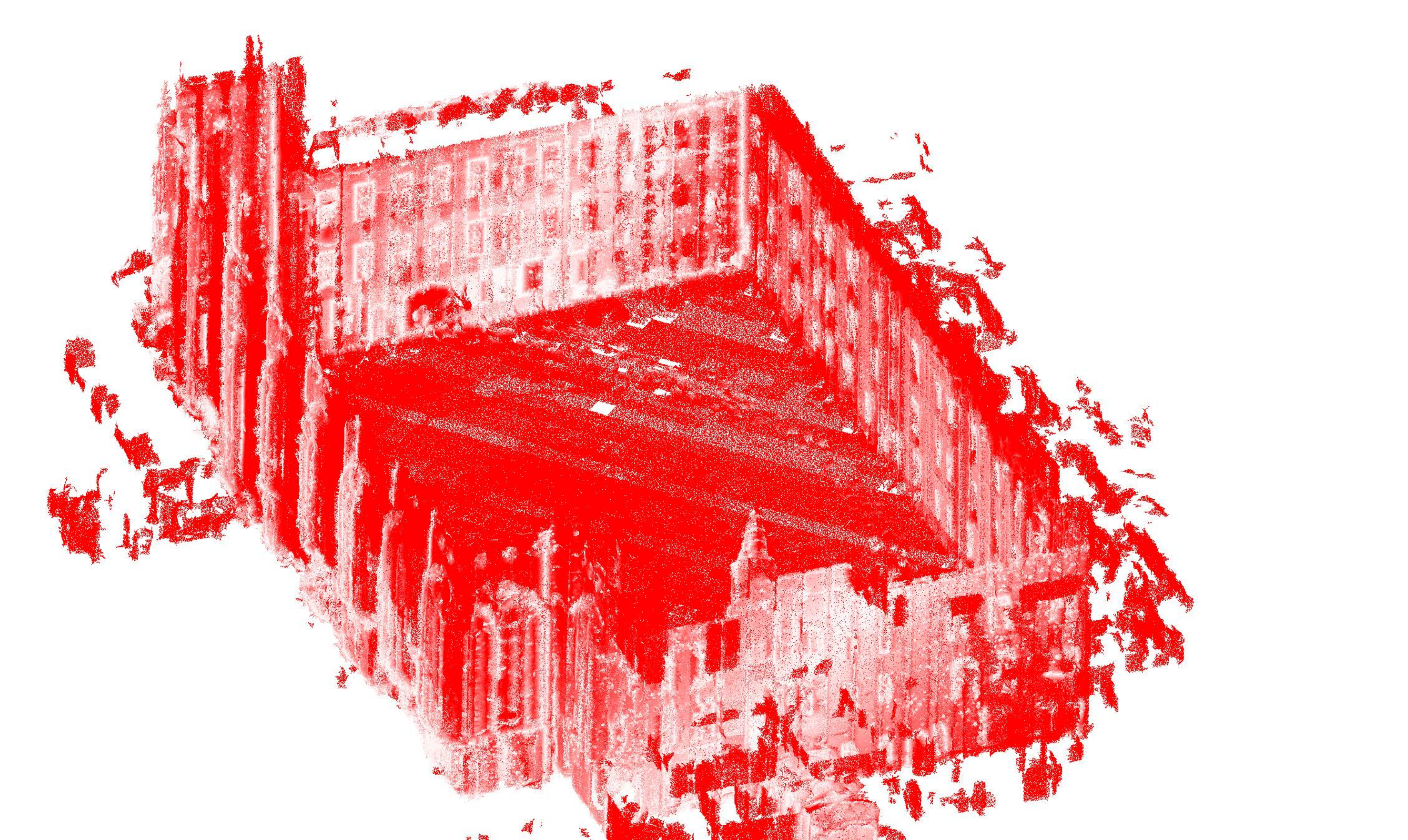}}
\hspace{0.03em} 
\vspace{-0.3em}
\subfloat{\includegraphics[width=.24\linewidth]{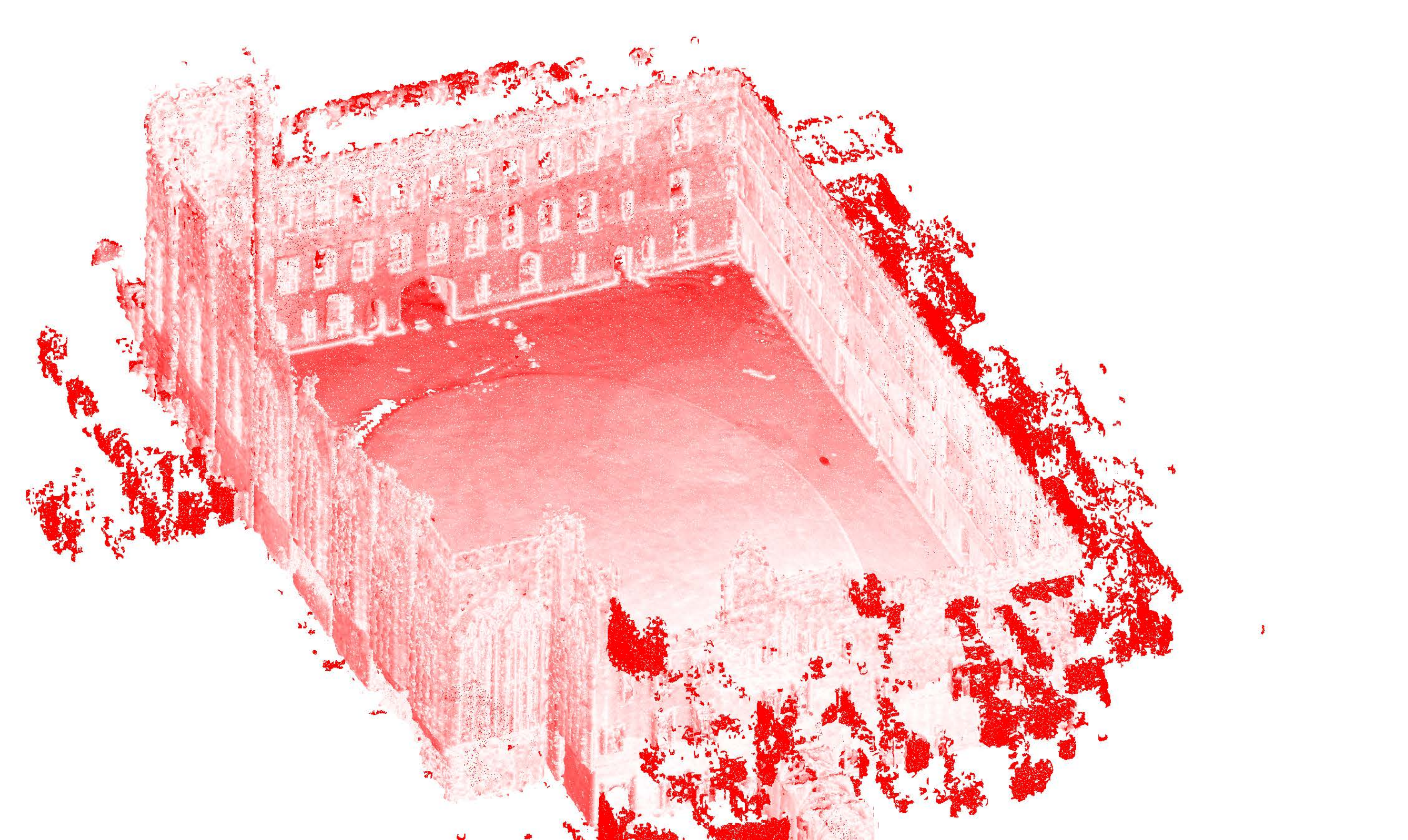}} 
\hspace{0.03em} 
\vspace{-0.3em}
\subfloat{\includegraphics[width=.24\linewidth]{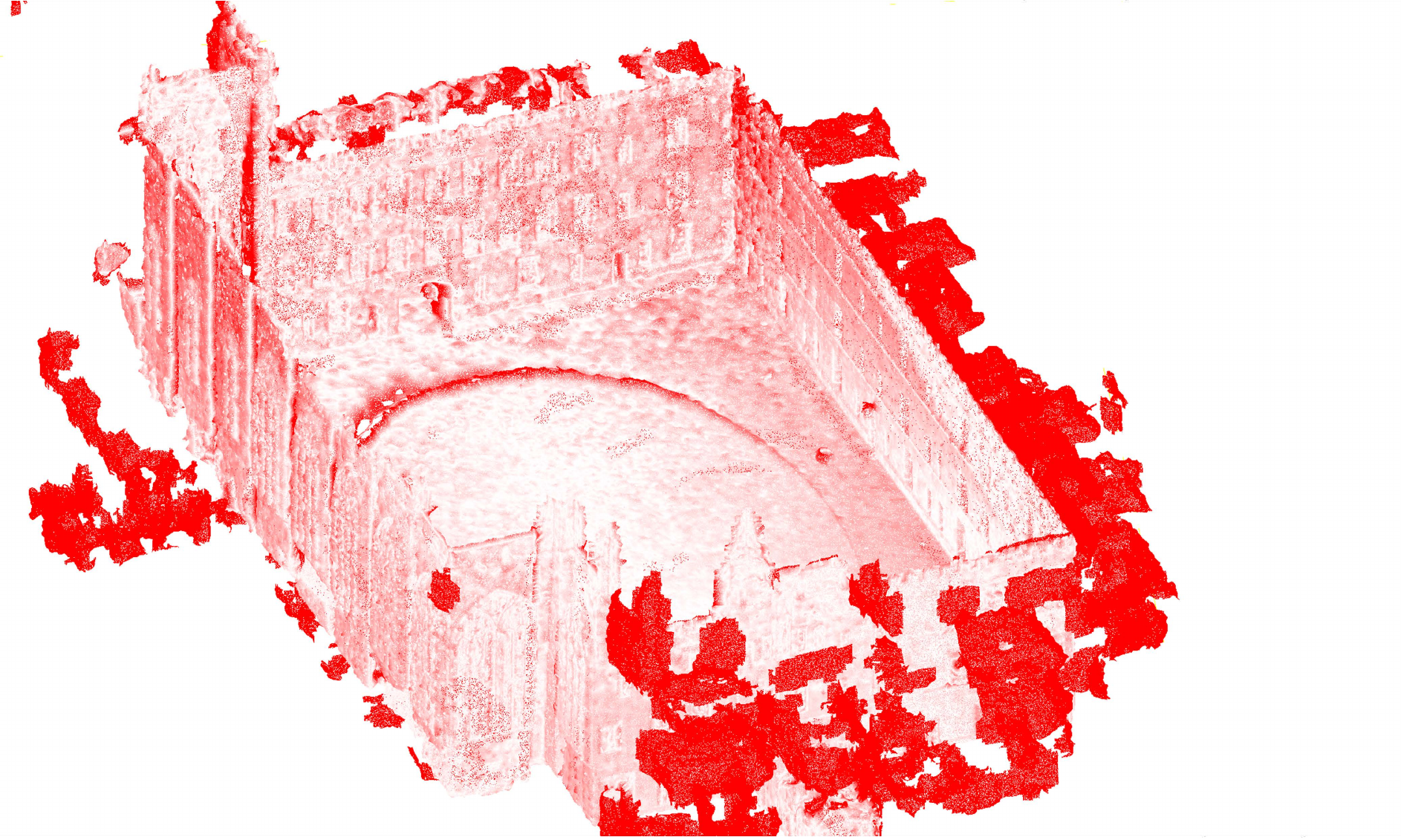}}
\hspace{0.03em} 
\vspace{-0.3em}
\subfloat{\includegraphics[width=.24\linewidth]{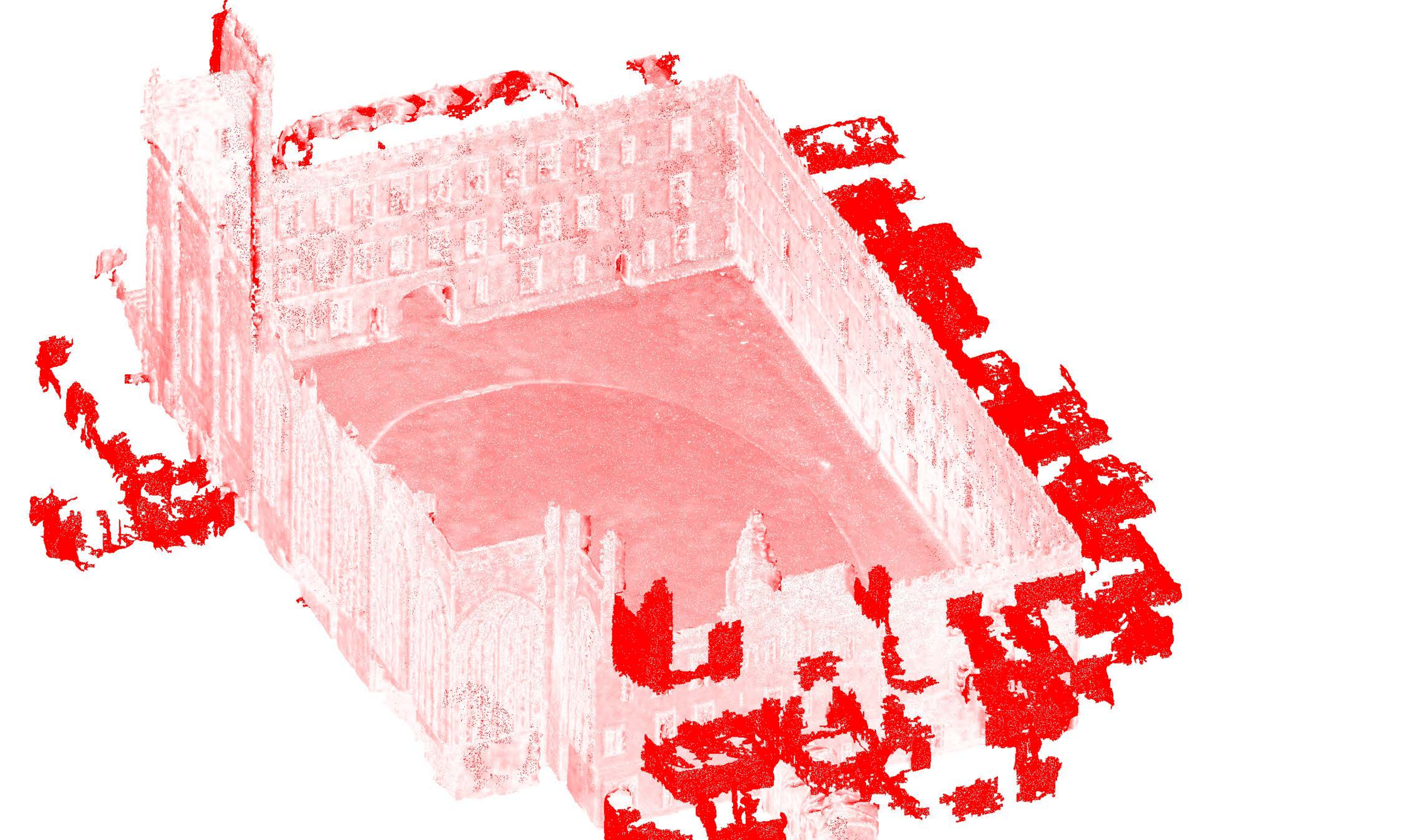}}
\caption{\textbf{The Visualization of Our 3D Reconstruction Results on \textit{Newer College} Dataset via Comparing with Other Related Methods.} The mapping results in first row are original reconstruction result, and the second row presents the error maps with using ground truth mesh as a reference, where the red points mean larger error up to 50 cm. (\textit{Best viewed with zoom in.)}}
\label{fig:rec_N}
\end{figure*}
we obtain superior localization results on the \textit{Hilti-21} dataset compared to other state-of-the-art LiDAR SLAM or odometry approaches. The trajectory and map results from two \textit{Hilti} are shown in Fig. \ref{fig:hilti_map}. Our result on \textit{Hilti-23} manifests the robustness of the hierarchical feature embedding for localization in highly-repetitive environments like construction sites, where PIN-SLAM \cite{pan2024pin} failed to boot the odometry. As shown in Table \ref{tab:ate_ncd}, we successfully conduct experiments across all of scenes from \textit{Newer College} dataset, without failures even in the particularly challenging sequences like \textit{02} and \textit{stairs}, and achieve the second-best results. To be noticed, although KISS-ICP achieves best average localization result here, while its localization is unsuccessful for sequence \textit{stairs}.

\subsubsection{\textbf{Evaluation on the Synthetic Dataset}} We test our method not only on real-world datasets, but we also evaluate its localization accuracy on synthetic dataset called \textit{Mai City} \cite{vizzo2021poisson}. For fair comparison with other methods, we conduct experiments on sequence 01 of this dataset, as shown in Table \ref{tab:ate_mai}, we achieved very favorable results compared to other state-of-the-art approaches. It demonstrates the robustness and generalizability of our method.

\begin{figure}[t!] 
    \centering 
    
    \includegraphics[width=0.48\textwidth]{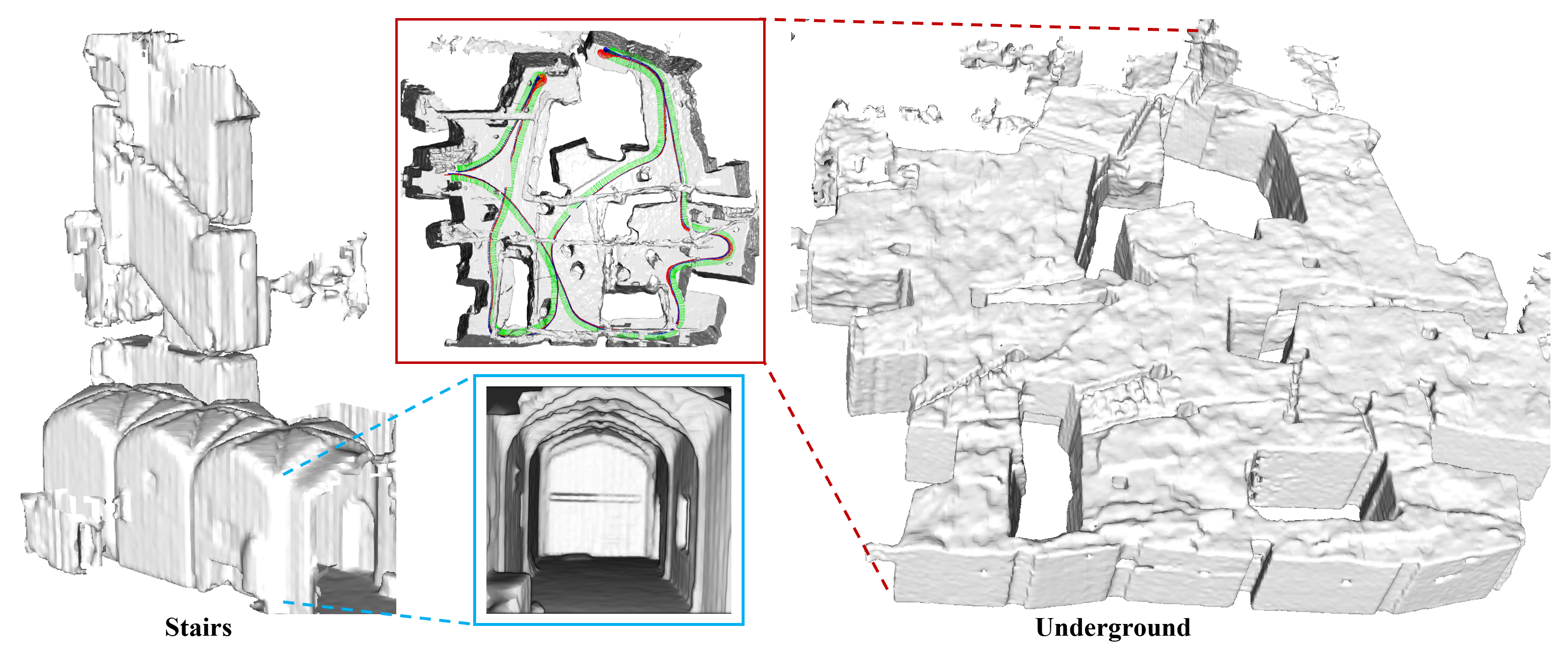} 
    \caption{\textbf{The Reconstructed Mesh of Our \FrameworkNM{} on \textit{Newer College} Dataset.} We show the estimated trajectory by overlapping it on the map for the sequence of the underground, and we show the inner structure for the sequence of stairs.  (\textit{Best viewed with zoom in.})} 
    \label{fig:college_map} 
\end{figure}
\subsection{Mapping Quality}
\subsubsection{\textbf{Quantitative Evaluation}} As shown in Table \ref{tab:recon_experiments}, we present the quantitative results of our method on the \textit{Mai City} and \textit{Newer College} datasets, compared to existing state-of-the-art (SOTA) methods. Since the mapping procedure of SHINE-mapping \cite{zhong2023shine} and VDB-Fusion \cite{vizzo2022vdbfusion} assume the pose is given, we utilize the poses estimated from KISS-ICP \cite{vizzo2023kiss} to fulfill the mapping.
\begin{figure*}[b!] 
    \centering 
    \includegraphics[width=1\textwidth]{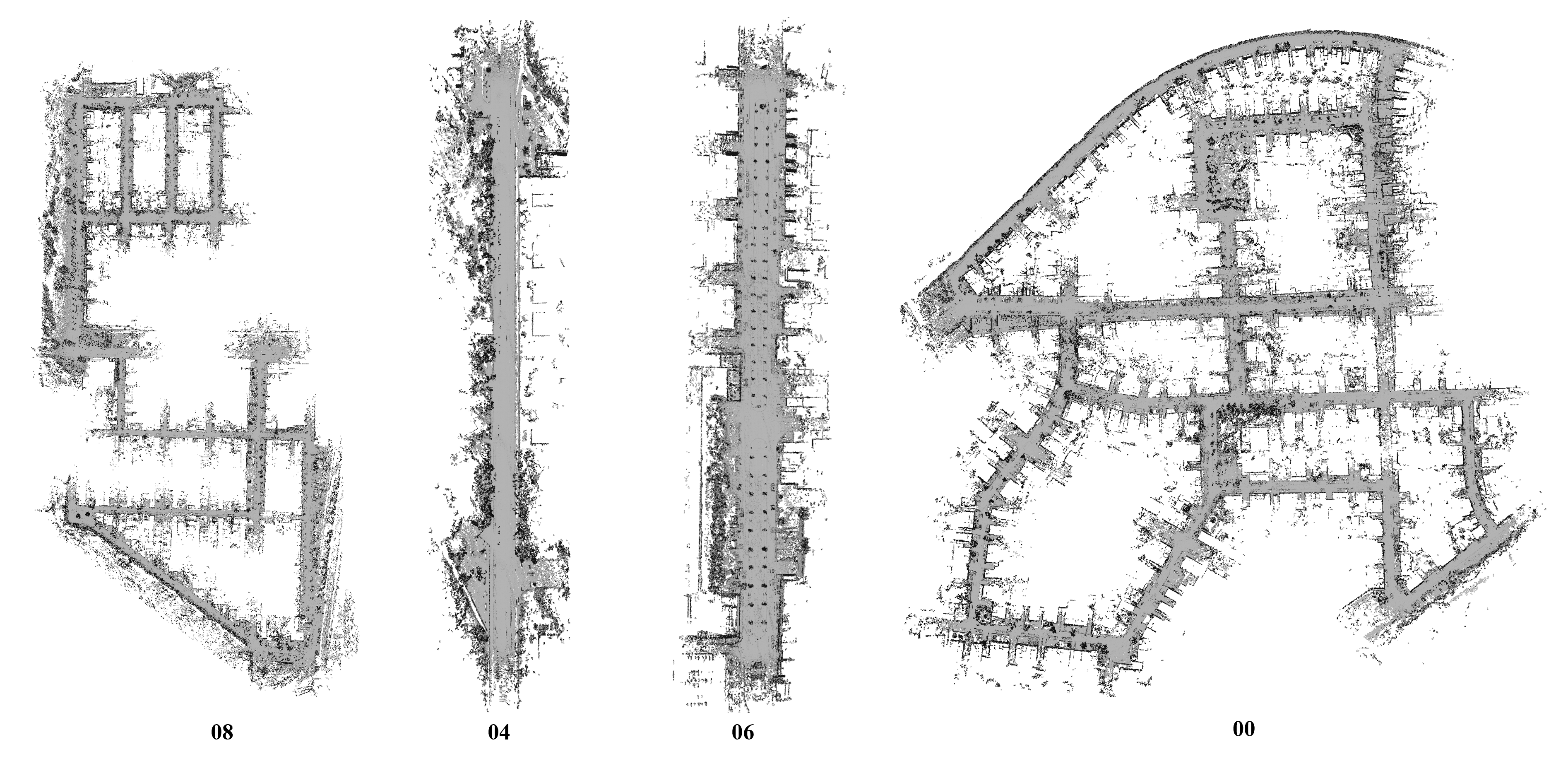} 
    \caption{\textbf{The Qualitative Results of Our Reconstructed Mesh on \textit{KITTI} Dataset.} Our reconstructed mesh on sequence 00, 04, 06, and 08 demonstrate the construction capacity of \FrameworkNM{} in large-scale environment.} 
    \label{fig:kitti_map} 
\end{figure*}
Puma \cite{vizzo2021poisson} is a Poisson-regression-based mapping approach. SLAMesh \cite{ruan2023slamesh} is an explicit mapping method, while NeRF-LOAM \cite{deng2023nerf}, PIN-SLAM \cite{pan2024pin}, and our \FrameworkNM{} are implicit mapping methods. Consistent with previous works, we use completion, accuracy, and Chamfer-L1 in cm, and F-score in \% as evaluation metrics. On \textit{Mai City} dataset, F-score is calculated
\begin{figure}[t!] 
    \centering 
    
    \includegraphics[width=0.48\textwidth]{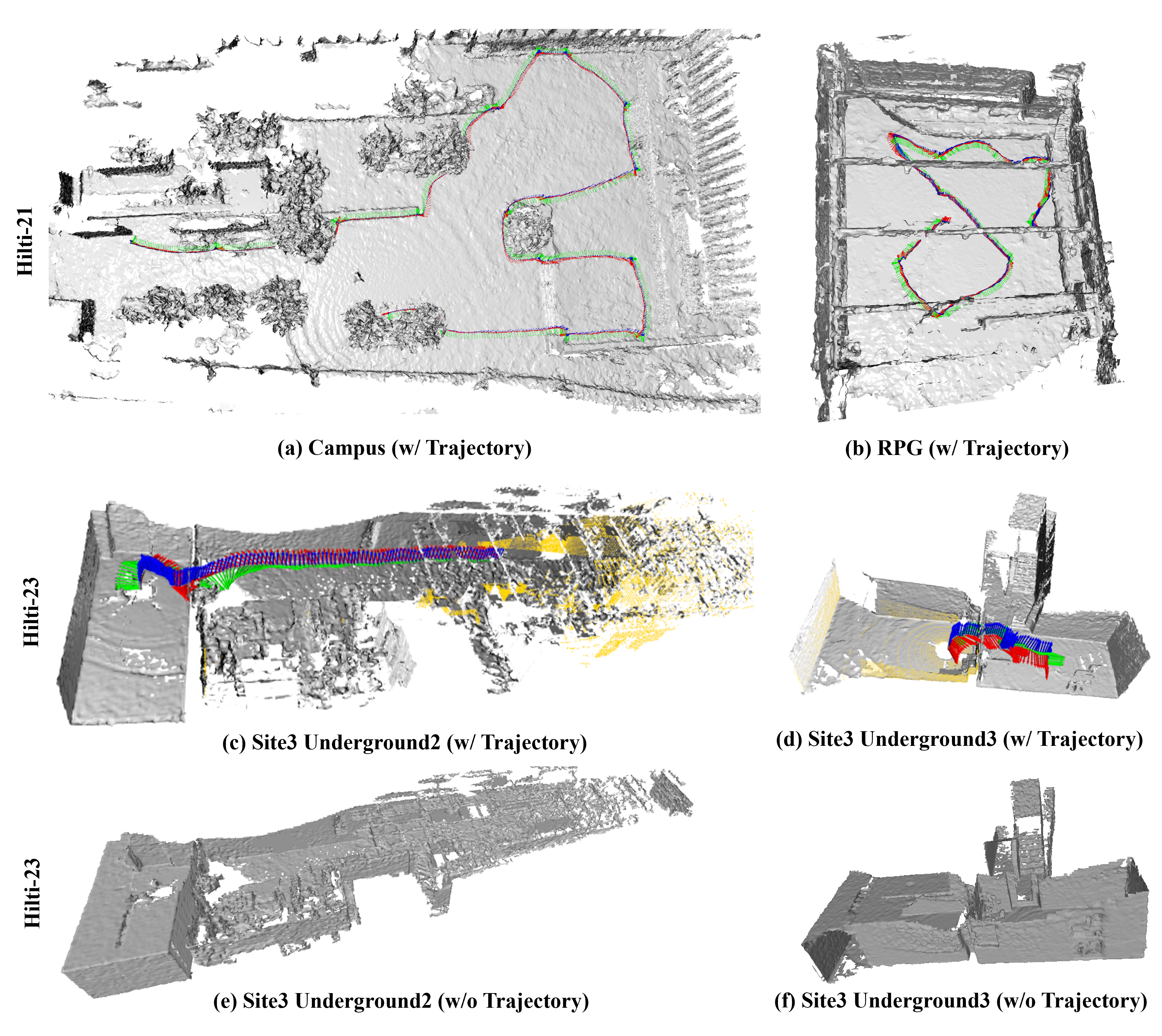} 
    \caption{\textbf{The Reconstructed Mesh of our \FrameworkNM{} on \textit{Hilti-21} and \textit{Hilti-23} Datasets.} The mapping environment of \textit{Hilti-23} is made up by construction sites, which are full with highly-repetitive or feature-less walls. The mapping ability for this extreme condition is shown in the above sub-figures. The estimated trajectory is also overlapped in the map to visualize the localization result. (\textit{Best viewed with zoomin.}) } 
    \label{fig:hilti_map} 
      \vspace{-6pt}
\end{figure}
with a 10 cm error threshold, and on \textit{Newer College} dataset, F-score is calculated with a 20 cm error threshold. From the Table \ref{tab:recon_experiments}, it can be seen that the quality of our mapping surpasses the existing state-of-the-art methods in all metrics, demonstrating the superiority of our approach.


\subsubsection{\textbf{Qualitative Evaluation}} We also conduct a qualitative comparison and analysis with the other two implicit methods, NeRF-LOAM \cite{deng2023nerf} and PIN-SLAM \cite{pan2024pin}. Corresponding results are visualized in Fig. \ref{fig:rec_mai} and Fig. \ref{fig:rec_N}, we observe that our reconstructed maps are more accurate and complete compared to other methods. The error maps also demonstrate that our method achieves higher mapping accuracy than the others. For mapping visualization of hand-held LiDAR data, reconstructed maps and corresponding trajectories for sequences from \textit{Newer College} are visualized in Fig. \ref{fig:college_map}, and the map reconstruction results for sequences from \textit{Hilti-21}, \textit{Hilti-23} are presented in Fig. \ref{fig:hilti_map}. We also visualize our map reconstruction result on \textit{KITTI} dataset in Fig. \ref{fig:kitti_map}, to show the consistent reconstruction ability in the large-scale and challenging environment.

\subsection{Ablation Study and Sensitivity Analysis}
\label{sec:exp:subsec:AblandSensit}
In this subsection, we first display the ablation study experiment to validate the effectiveness of our multi-scale hierarchical features in improving localization accuracy.
We adopt the root mean square error (RMSE) of the absolute trajectory error (ATE) as the evaluation metric, to evaluate the performance of our odometry in different feature level cases. As shown in Table \ref{tab:LOF_ATE}, we validate whether multiple
\begin{table*}[b!]
  \centering
  \caption{\textbf{The Ablation Study of the Level of Feature (LOF) on \textit{KITTI} and \textit{SemanticPOSS} LiDAR Benchmark.} Localization Performance (ATE RMSE [m]) are reported for system evaluation.}
  \resizebox{0.98\textwidth}{!}{
      \begin{tabular}{c|cccccc|c|cccccc|c}
          \toprule
         \multirow{2}{*}{\makecell{LOF}} & \multicolumn{7}{c|}{\textit{KITTI}} & \multicolumn{7}{c}{\textit{SemanticPOSS}}\\
         \cmidrule(){2-15}
         & \text{00} & \text{02} & \text{04} & \text{06} & \text{08} & \text{10} & \textbf{Avg.} & \text{00} & \text{01}      & \text{02} & \text{03} & \text{04} &  \text{05} & \textbf{Avg.} \\
          \midrule
         1 & 5.96 & 10.42 & 0.45 &  0.76 &  2.93 &  0.88 & 3.57  & 0.28 & 0.43 & 0.21 &  0.29 &  0.25 &  0.29 & 0.29 \\
         2 & 5.00 & 11.35 & 0.17 &  0.66 &  2.93 &  0.88 & 3.50 & 0.27 & 0.41 & 0.20 &  0.28 &  0.22 &  0.24 & 0.27  \\
         3 & 5.00 & 8.60 & 0.14 &  0.63 &  2.80 & 0.81 & 3.00 & 0.27 & 0.29 & 0.22 &  0.25 &  0.23 & 0.23 & 0.25  \\
          \bottomrule
      \end{tabular}
      }
    \label{tab:LOF_ATE}
\end{table*}
level of features (LOF) can improve the localization accuracy on sequences of the \textit{KITTI} dataset and \textit{SemanticPOSS} dataset. As we observe, with the number of feature levels increasing, the localization accuracy is gradually improved, it demonstrate the value of hierarchical feature design.

Next, we report sensitivity analysis experiments for three key parameters, leaf node sizes, tree level numbers, and submap numbers. Our tuning strategy is to start from the leaf node size, the system localization results when leaf node size set to 0.1, 0.2, 0.25 and 0.3 are reported in in Table \ref{tab:LeafNodesize_ATE}. As we noticed, the best leaf node size is around 0.2 based on the average value in the last column. We can slightly increase the leaf node size to 0.25 m for large-scale scenes of \textit{KITTI} sequence. Overall, we set leaf node size as 0.2 m for our \FrameworkNM{} system.

Then, we tuned the different octree level numbers from 9 to 15, as shown in Table \ref{tab:TreeLevel_ATE}. We observe that octree level numbers is related to the scale of scenes. For large-scale scenes like sequence 05 of \textit{KITTI}, which occupied an area around 10 m $\times$ 400 m, the optimal tree level is 15. For scenes like sequence 04 and sequence 06 of \textit{KITTI}, which occupied an area around 5 m $\times$ 8 m and 8 m $\times$ 15 m, the optimal tree level is 13. The exception is the tuning of \textit{KITTI 03} due to the complexity and dynamic elements in the scene, if the tree level numbers decreased too much, the performance will drop sharply. Furthermore, for small-scale indoor cases such as \text{quad\_e}, \text{ug\_e}, and \text{stairs}, the less octree level, such as 9 or 10 , is needed due to the scene size. To be noticed, for the synthetic dataset \textit{MaiCity 01}, the tree level 10 generates best result.

Lastly, the sensitivity analysis of submap size and its tangible impact on localization performance is conducted in Table \ref{tab:SubmapNumber_ATE}. Here, several typical submap sizes are examined. As can be seen, when the scene scale is larger, the submap needs larger size to cover the scene ambiguity, small submap size for large scene will decrease the localization accuracy. Same to the scale rule manifested in Table \ref{tab:TreeLevel_ATE}, for small-scale indoor cases such as \text{quad\_e}, \text{ug\_e}, and \text{stairs}, the smaller submap size, such as 50, is needed due to the scene size. It is interesting to find the submap size 50 generates the best result for the synthetic dataset \textit{MaiCity 01}, so the tuning rule of the synthetic dataset is different, considering the scale, 100 m $\times$ 5 m, of \textit{MaiCity 01} is larger than those real-world small-scale indoor cases, such as \text{quad\_e}, \text{ug\_e}, and \text{stairs}.

\begin{figure}[h!] 
    \centering 
    \includegraphics[width=0.48\textwidth]{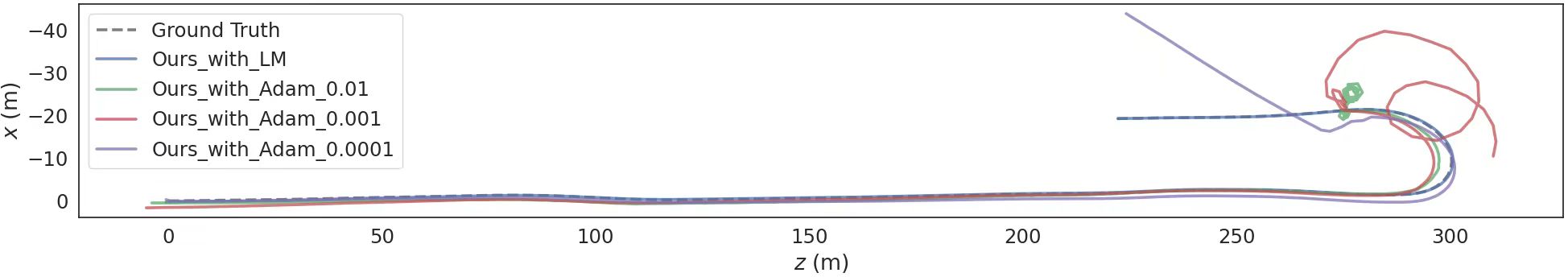} 
    \caption{Ablation Study of Different Optimizer Setting. The blue line represents the results obtained by replacing the Levenberg-Marquardt (LM) method with the Adam optimizer in our approach. The black dashed line represents the ground truth trajectory. It is evaluated by first 380 scans in \textit{KITTI 06}, which is a typical case for optimizer selection.}    \label{fig:kitti_06} 
\end{figure}

On the other hand, we also examine to utilize the Adam optimizer as the optimization method rather than the Levenberg-Marquardt (LM) to search the best rotation $\mathbf{R}$ and translation $\mathbf{t}$. We find it out the training procedure of implicit localization is sensitive to the learning rate if Adam is adopted. We need to fine-tune the learning rate to complete the pose estimation of certain sequences, the trajectory of failure cases are shown in Fig. \ref{fig:kitti_06}. And we only present the results of the first 380 frames of the sequence 06 of \textit{KITTI} dataset. It can be seen that there are obvious localization errors in the trajectories of other methods compared to the ground truth, leading to spinning, sharp drift failures in the sequence.

\begin{table*}
\centering 
\caption{\textbf{Sensitivity Analysis of Leaf Node Sizes on \textit{KITTI}, \textit{Newer College}, and \textit{Mai City} Benchmark.} Localization Performance (ATE RMSE [m]) are reported for system evaluation. \xmark \hspace{0.03cm} denotes the execution failure.}
\resizebox{0.985\textwidth}{!}{
  \begin{tabular}{c|cccccccc|c}
    \toprule
    \multirow{2}{*}{\makecell{ Leaf Node Size }}
   &  \textit{KITTI 03} & \textit{KITTI 05} & \textit{KITTI 06} &
    \textit{KITTI 04} &\text{quad\_e}  & \text{ug\_e} & \text{stairs} & \textit{MaiCity 01} & Avg.\\
        \cmidrule(){2-9}
     & 25m$\times$400m & 10m$\times$400m & 8m$\times$15m &
    5m$\times$8m & 40m$\times$30m & 30m$\times$20m & 10m$\times$2m & 100m$\times$5m &    \\   
    \midrule
    0.1    & \xmark &	2.097552 & 0.728338 & 0.155302 &	0.111781 & 4.796704 &	0.112531 & 0.016174
 & 1.145483 
\\
    0.2  & \xmark 
    &
    1.636037 & 0.529766  & \bf{0.134920 } &	\bf{0.093267} & \bf{0.052786}	&\bf{0.077359} & \bf{0.006776} & \textbf{0.360590}
  \\
    0.25 
    &\bf{0.477824}	& \bf{1.460346} &\bf{0.391477} & 0.154256 & 0.097045 &	0.076913 & 0.265139 &	0.014541
 & 0.367193
  \\
    0.3  &  0.544378 & 1.547646 & 0.521631  & 0.170558 & 0.099131 & \xmark & 0.261116 & 0.013292
 & 0.451107
   \\
    \bottomrule
  \end{tabular}
  } 
  \vspace{-2pt}
  \label{tab:LeafNodesize_ATE}
\end{table*}

\begin{table*}
\centering 
\caption{\textbf{Sensitivity Analysis of Tree Level Numbers on \textit{KITTI}, \textit{Newer College}, and \textit{Mai City} Benchmark.} Localization Performance (ATE RMSE [m]) are reported for system evaluation. \xmark \hspace{0.03cm} denotes the execution failure.}
\resizebox{0.985\textwidth}{!}{
  \begin{tabular}{c|cccccccc}
    \toprule
     \multirow{2}{*}{\makecell{Tree Levels }} 
    &  \textit{KITTI 03} & \textit{KITTI 05} & \textit{KITTI 06} &
    \textit{KITTI 04} &\text{quad\_e}  & \text{ug\_e} & \text{stairs} & \textit{MaiCity 01} \\
        \cmidrule(){2-9}
     & 25m$\times$400m & 10m$\times$400m & 8m$\times$15m &
    5m$\times$8m & 40m$\times$30m & 30m$\times$20m & 10m$\times$2m & 100m$\times$5m \\
    \midrule
    9    & \xmark & \xmark &	\xmark & \xmark &	0.100113 & \bf{0.052786}
 &	0.169223 & 0.012432 
 \\
    10 & \xmark & \xmark  & \xmark & \xmark &	\bf{0.097937} & 0.081195 &\bf{0.079966} &\bf{0.006776} 
\\
    11  & 82.758527& \xmark &	0.761584  & 44.536743 & 0.099720 &	0.230535 & 0.112515 & 0.009499
 
  \\
    12 &  \bf{0.477824} &
    1.621471 & 0.488580  & 0.142222 & 0.104466 & 0.214824 & 0.138130 & 0.010439
 
 \\
    13  & 0.572033  &	1.653113 & \bf{0.391477}  & \bf{0.134920} &	0.100589 & 0.464130 &	0.194548 & 0.016236 
    \\
    14  & 0.578985 &	1.602890 & 0.518762  & 0.171795 &	\xmark & \xmark &	0.127334 & 0.022344 
  \\
    15   & 0.593703 &	\bf{1.460346} &  0.504772  & 0.167075 &	\xmark & \xmark &	\xmark & 0.021543
  \\

    \bottomrule
  \end{tabular}
  } 
  \vspace{-2pt}
  \label{tab:TreeLevel_ATE}
\end{table*}

\begin{table*}
\centering 
\caption{\textbf{Sensitivity Analysis of Submap Numbers on \textit{KITTI}, \textit{Newer College}, and \textit{Mai City} Benchmark.} Localization Performance (ATE RMSE [m]) are reported for system evaluation. \xmark \hspace{0.03cm} denotes the execution failure.}
\resizebox{0.985\textwidth}{!}{
  \begin{tabular}{c|ccccccccc}
    \toprule
    \multirow{2}{*}{\makecell{Submap Numbers}} &  \textit{KITTI 03} & \textit{KITTI 05} & \textit{KITTI 06} &
    \textit{KITTI 04} &\text{quad\_e}  & \text{ug\_e} & \text{stairs} & \textit{MaiCity 01} \\
        \cmidrule(){2-9}
     & 25m$\times$400m & 10m$\times$400m & 8m$\times$15m &
    5m$\times$8m & 40m$\times$30m & 30m$\times$20m & 10m$\times$2m & 100m$\times$5m \\
    
    \midrule
    50  &  \xmark &  \xmark &  \xmark  & \xmark & 0.093394 & \bf{0.052786} &	\bf{0.077359} & \bf{0.006267}  
 \\
    60  & \xmark & \xmark &  \xmark  & \xmark & 0.093809 & 0.068807 &	0.089445
 &0.012224 
  \\
    100
   
    & 0.538114 
  & 1.564616 &	0.513978  & 0.150586 & \bf{0.093267} &	0.081796 & 0.079966 &	0.012838  
  
  \\
    150 &  0.582111
  & 1.717656 & 0.486967 &0.145847&  0.094171 & 0.069654
 & 0.106437 & \xmark
  

 \\
    200  & 0.581955
    &\bf{1.460346} 
    & \bf{0.391477}  & \bf{0.134920}
 &	0.099720 & 0.134966 & 0.143389
 & \xmark 
 
  \\
    250  & 0.581955  &	1.503061 & 0.489490  & 0.134658 &	0.100572 & 0.125165 & 0.198023 & \xmark 
  \\
     300   & \bf{0.477824} &	1.539094 & 0.480684 & 0.134792 &	0.100750 & 0.090225 &	0.125138
 & \xmark 
  \\
    \bottomrule
  \end{tabular}
  } 
  \vspace{-2pt}
  \label{tab:SubmapNumber_ATE}
\end{table*}

\subsection{Memory and Running time Analysis}
First, we perform a memory usage comparison analysis with other methods. We utilize the memory consumption of the raw point cloud as a reference. SuMa \cite{behley2018efficient} adopts surfels as the basic map elements, while the others utilize mesh maps. For VDB Fusion \cite{vizzo2022vdbfusion} map, we choose a voxel grid of 20 cm and select the odometry estimation of KISS-ICP as its pose. For the map of PIN-SLAM and our method, we use the same resolution of 40 cm for the neural points and the latent feature grid (leaf node). As shown in Table \ref{tab:mapmemory}, we achieve the second-best memory efficiency. Compared to PIN-SLAM, our map consumes slightly more memory, this is because we adopt hierarchical multi-scale features instead of a single-level feature representation, trading increased memory usage for improved map quality.
\begin{table}[b!]
  \centering
  \caption{Memory consumption comparison with different method in MB. NCD means \textit{Newer College} dataset.}
  \resizebox{0.48\textwidth}{!}{
  \begin{tabular}{ccccc}
  \toprule
  Method & KITTI \text{00} & KITTI \text{05} & KITTI \text{08} & NCD \text{02} \\
  \midrule
  Raw point cloud  & 13624.2 & 8284.7 & 12214.1 & 26559.0 \\

  SuMa\cite{behley2018efficient} & 887.7 & 512.6 & 835.7 & 79.0 \\
  Puma\cite{vizzo2021poisson} & 2032.9 & 1317.4 & 1894.1 & 1503.7 \\
  VDB Fusion\cite{vizzo2022vdbfusion} & 748.1 & 434.6 & 958.6 & 462.5 \\ 
  PIN-SLAM\cite{pan2024pin} & \bf{102.1 } & \bf{66.3 } & \bf{138.8 }  & \bf{76.8 } \\
  \midrule
  Hi-LOAM & \underline{242.1} & \underline{169.5} & \underline{284.3} & \underline{176.6}\\
  \bottomrule
  \end{tabular}
  }
  \label{tab:mapmemory}
\end{table}


\begin{table}[b!]
  \centering
  \caption{Comparison of the average running time and the localization error of different method on \textit{KITTI} sequence 00-10. ARTE represents the average relative translational error.}
  \resizebox{0.48\textwidth}{!}{
  \begin{tabular}{c|ccc}
  \toprule
  Method  & Time per frame [s] $\downarrow$ & FPS [Hz] $\uparrow$ & ARTE [\%] $\downarrow$  \\
  \midrule
  PIN-SLAM \cite{pan2024pin}  & \bf{0.034}  & \bf{29.4} & \underline{0.51}  \\

  NeRF-LOAM \cite{deng2023nerf} & 4.53 & 0.2  & 1.69 \\
  \midrule
  Hi-LOAM & \underline{0.6} & \underline{1.7} & \bf{0.48} \\
  \bottomrule
  \end{tabular}
  }
  \label{tab:time_ate}
\end{table}
For the second experiment, we compare the computational resource consumption of our odometry method with two other implicit mapping approaches PIN-SLAM \cite{pan2024pin} and NeRF-LOAM \cite{deng2023nerf} on the KITTI dataset, the experiment is conducted on a single NVIDIA GeForce RTX 4090 graphics processing unit (GPU). As shown in Table \ref{tab:time_ate}, we evaluate the average pose estimation time per frame. Although our method requires more time than PIN-SLAM for estimating each frame’s pose, it delivers higher localization accuracy. This performance gain is primarily due to the use of octree-based hierarchical features, which provides richer geometric and spatial information, by sacrificing more computational demand than single-layer features methods. In comparison with NeRF-LOAM, our method consistently outperforms it in terms of both efficiency and accuracy. 

As shown in Table \ref{tab:time_submodule}, we also analyze the per-frame computational cost of each sub-module in our method across different datasets. It can be observed that the pose estimation time of the odometry module is generally much lower than that of the map optimization module. This is because our method estimates poses by matching the scan to the implicit map, which relies on the optimized map for accurate localization. As a result, we perform more iterations in the map optimization process to ensure higher pose accuracy.
\begin{table}[t!]
  \centering
  \caption{\textbf{Breakdown of Average Running Time per Frame (s) for Each Module on Different Datasets.} Quad\_e from the \textit{Newer College} dataset, KAI from the \textit{MulRAN} dataset, Lab from the \textit{Hilti-21} dataset, POSS 00 from the \textit{SemanticPOSS} dataset.}
  \resizebox{0.48\textwidth}{!}{
  \begin{tabular}{c|cccccc}
  \toprule
  Submodule  & KITTI 03 & MaiCity 01 & Quad\_e & KA1 & Lab & POSS 00 \\
  \midrule
  Odometry   &  0.3 & 0.14 & 0.34 & 0.38 & 0.2 & 0.49 \\
  Map Optimization & 3.6 & 2.87 &3.0 & 5.5 & 4.7 & 4.5\\

  \bottomrule
  \end{tabular}
  }
  \label{tab:time_submodule}
\end{table}

\section{Conclusion}
This paper shows a novel implicit framework, \FrameworkNM{}, to accomplish the LiDAR-based localization and mapping task, with the core idea of utilizing multi-scale hierarchical latent features based on the octree and neural networks to represent the geometry information. The octree-based latent feature can adaptively fit 3D space structure with multi-scale levels of detail, which provides a solid foundation to efficiently approximate shapes in the physical world and further facilitate the localization task. 
Qualitative and quantitative experiments are conducted on multiple real-world and synthetic datasets, we demonstrate that \FrameworkNM{} achieves better mapping quality than state-of-the-art (SOTA) LiDAR SLAM approaches, achieves better localization accuracy than SOTA learning-based LiDAR Odometry and on par with the SOTA ICP-based LiDAR Odometry. In the future, we anticipate to enroll other modality information into the latent features, explore the loop closure detection, and generalize to more types of LiDAR sensor data for embodied AI applications.




\bibliographystyle{IEEEtran}
\bibliography{mybib}
\section{Biography Section}

\vspace{-33pt}
\begin{IEEEbiography}[{\includegraphics[width=1in,height=1.25in,clip,keepaspectratio]{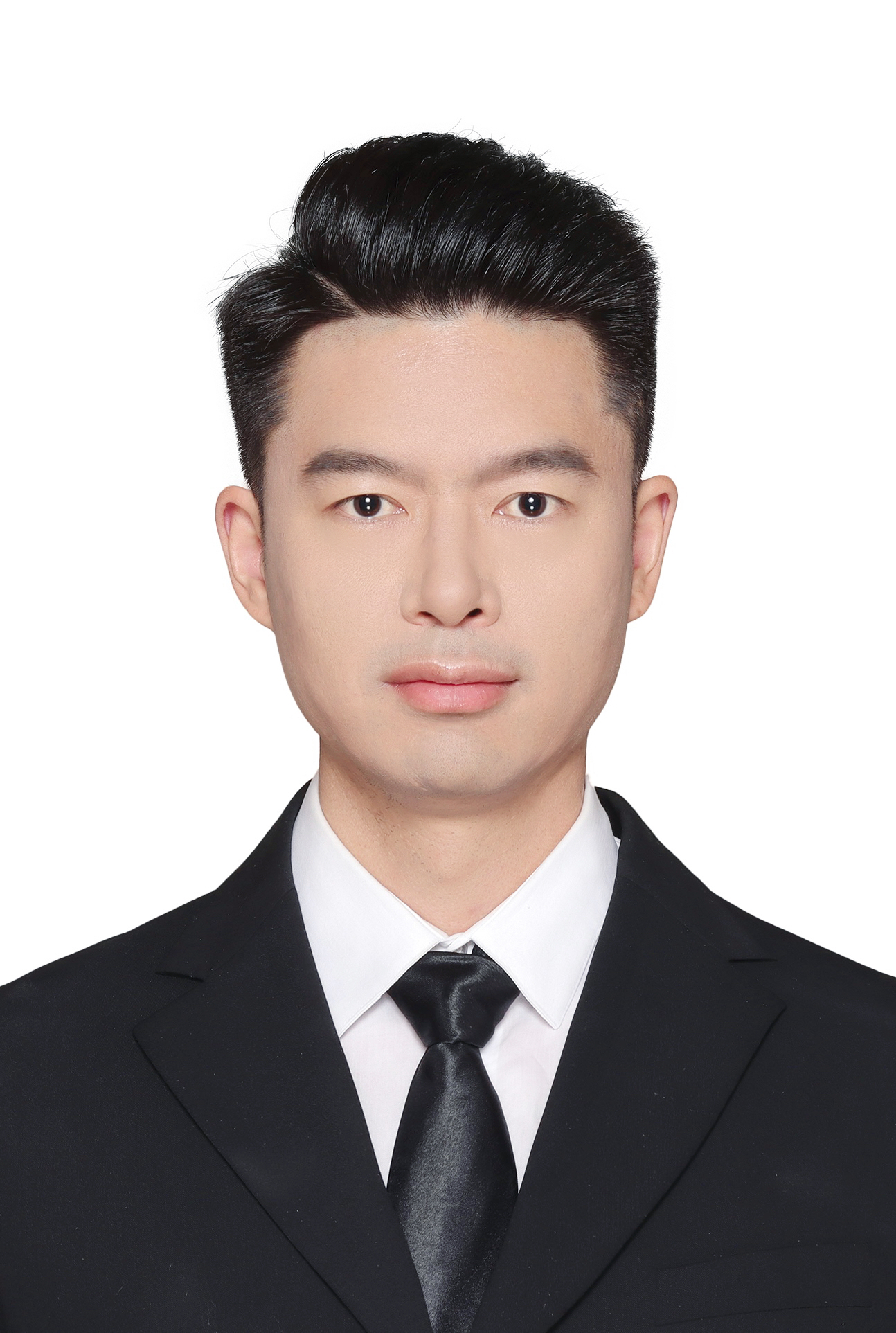}}]{Zhiliu Yang}
received his Ph.D. in Electrical and Computer Engineering from Clarkson University, Potsdam, New York, USA, in 2021. Currently he is an assistant professor in the School of Information Science and Engineering at Yunnan University, Kunming, China. He was a recipient of Young Talents Award of Xing Dian Program of Yunnan Province, China. His current research interest is in Embodied Intelligence, such as visual navigation, large-scale mapping, and agents localization.
\end{IEEEbiography}
\vspace{-20pt}

\begin{IEEEbiography}[{\includegraphics[width=1in,height=1.25in,clip,keepaspectratio]{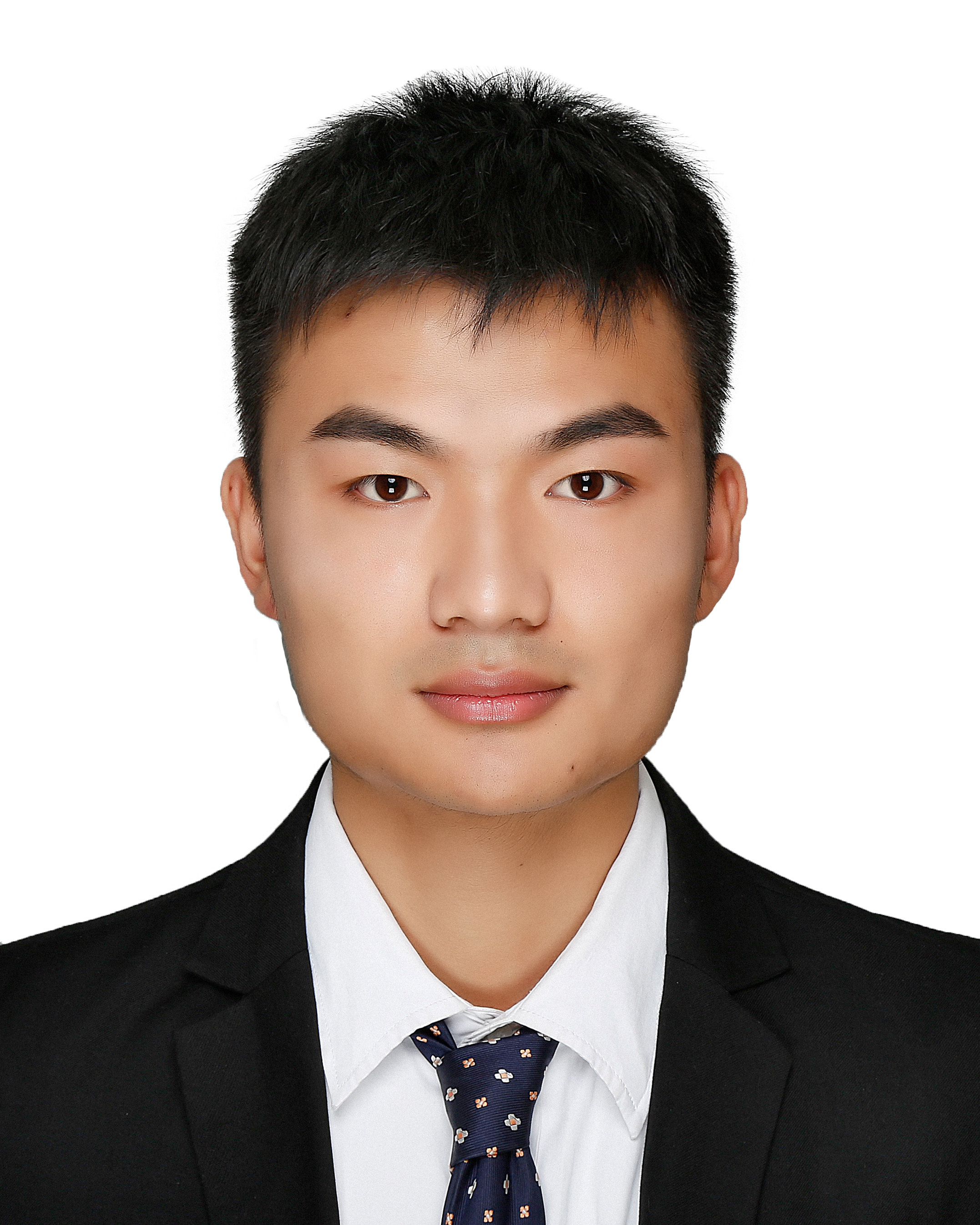}}]{Jianyuan Zhang} received his B.E. degree in Electronics Science and Technology from the Yunnan University, Kunming, China, in 2020. He is currently working toward the M.S. degree with the School of Information Science and Engineering, Yunnan University, Kunming, China. His research interests include visual perception, LiDAR odometry and implicit scene reconstruction.
\end{IEEEbiography}
\vspace{-20pt}

\begin{IEEEbiography}[{\includegraphics[width=1in,height=1.25in,clip,keepaspectratio]{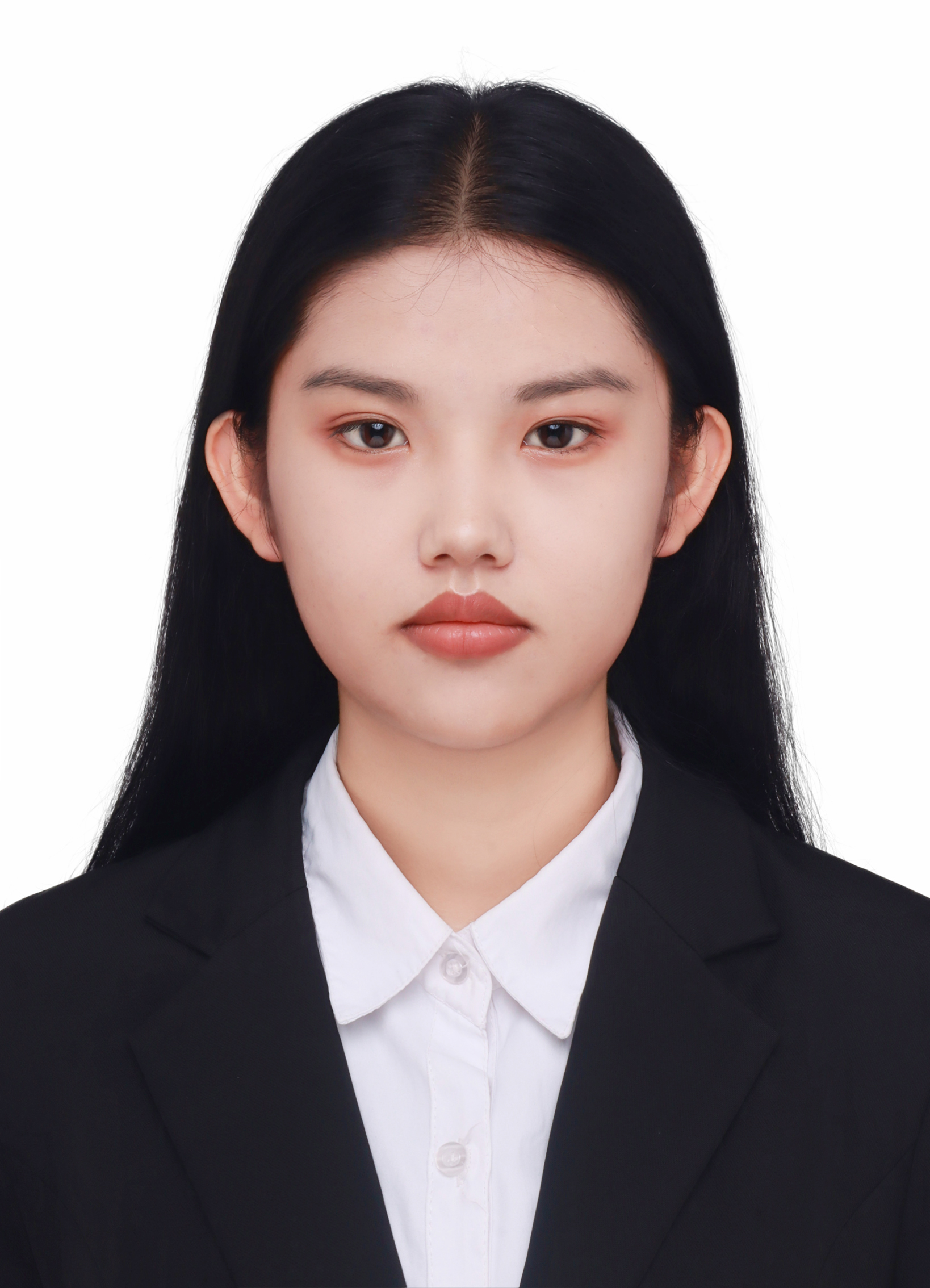}}]{Lianhui Zhao}
received her B.E. degree in Electronic Information Engineering from the Southwest Minzu University, Chengdu, China, in 2024. She is currently working toward the M.S. degree with the School of Information Science and Engineering, Yunnan University, Kunming, China. Her research interests include multi-modal spatial intelligence and 3D scene reconstruction.
\end{IEEEbiography}

\begin{IEEEbiography}[{\includegraphics[width=1in,height=1.25in,clip,keepaspectratio]{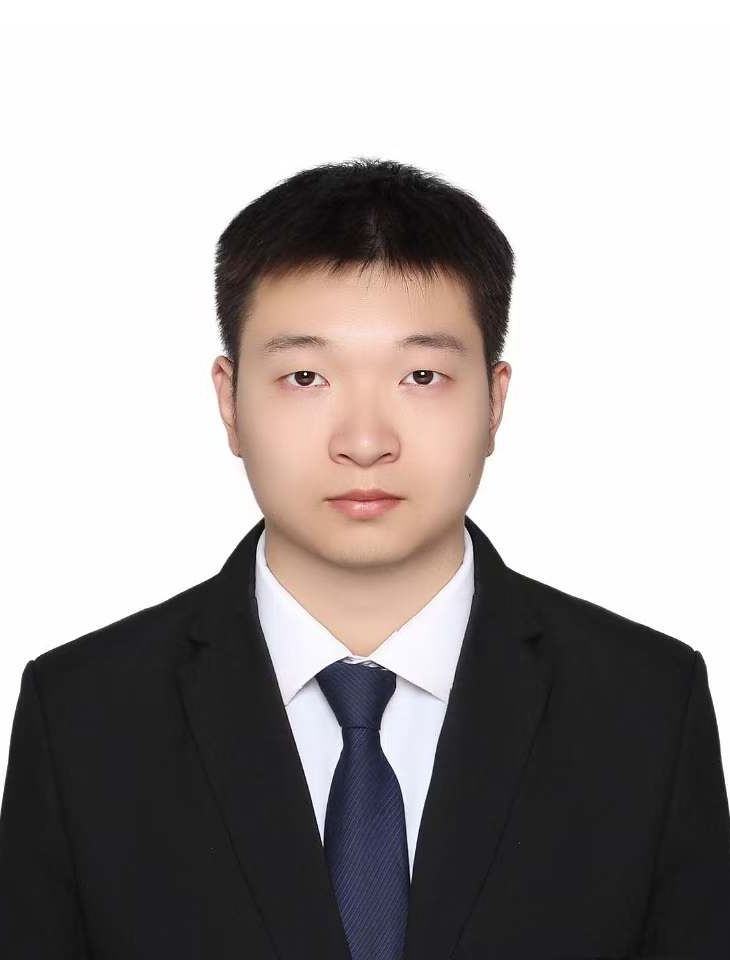}}]{Jinyu Dai} received his B.E. degree in Communication Engineering from Kunming University of Science and Technology, Kunming, China, in 2022. He is currently working toward the M.S. degree with the School of Information Science and Engineering, Yunnan University, Kunming, China. His research interests include visual perception and scene construction.
\end{IEEEbiography}
\vspace{-500pt}

\begin{IEEEbiography}[{\includegraphics[width=1in,height=1.25in,clip,keepaspectratio]{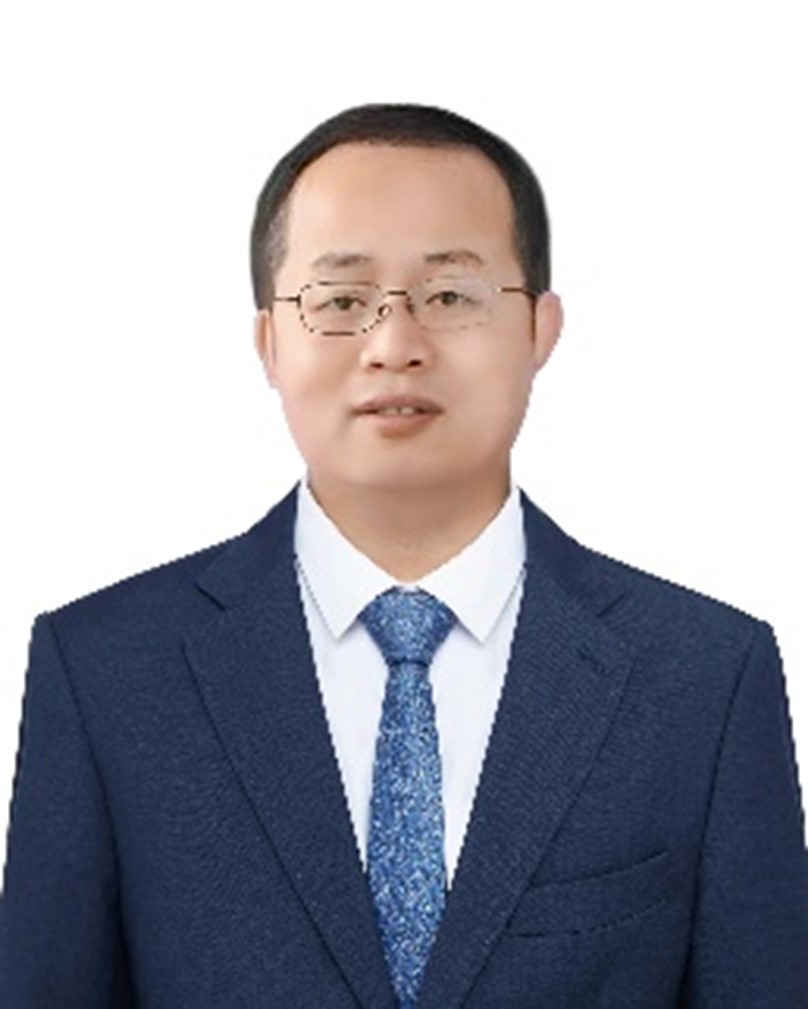}}]{Zhu Yang} received his B.E. degree in Electronic Information Engineering from University of Science and Technology of China (USTC) in 2000, and he received his Ph.D. degree from Beijing Institute of Technology in 2019. Currently, he is an assistant professor in the School of Information and Electronics, Beijing Institute of Technology. His research interests are chip design and signal processing related to intelligent remote sensing.
\end{IEEEbiography}

\end{document}